\documentclass{article}

\usepackage{microtype}
\usepackage{graphicx}
\usepackage{subfigure}
\usepackage{booktabs} % for professional tables

\usepackage{hyperref}

\usepackage[accepted]{icml2024}

\usepackage{amsmath}
\usepackage{amssymb}
\usepackage{mathtools}
\usepackage{amsthm}
\usepackage{xcolor}         % colors
\usepackage{bbm}
\usepackage{array}
\usepackage{tabularx}
\usepackage{ulem}
\usepackage{wrapfig}
\newcolumntype{P}[1]{>{\centering\arraybackslash}p{#1}}
\newcolumntype{M}[1]{>{\centering\arraybackslash}m{#1}}
\usepackage{amsmath,amsfonts,bm}
\usepackage{amsthm}
\usepackage{amssymb}
\usepackage{mathrsfs}

\usepackage{graphicx}
\usepackage{subfigure}

\def\bfa{{\boldsymbol a}}
\def\bfb{{\boldsymbol b}}
\def\bfc{{\boldsymbol c}}

\def\bfn{{\boldsymbol n}}
\def\bfo{{\boldsymbol o}}
\def\bfp{{\boldsymbol p}}
\def\bfq{{\boldsymbol q}}
\def\bfr{{\boldsymbol r}}

\def\bfv{{\boldsymbol v}}

\def\bfx{{\boldsymbol x}}
\def\bfy{{\boldsymbol y}}
\def\bfz{{\boldsymbol z}}
\def\bfmu{{\boldsymbol \mu}}
\def\bfnu{{\boldsymbol \nu}}

\def\bfO{{\boldsymbol O}}
\def\bfP{{\boldsymbol P}}

\def\bfU{{\boldsymbol U}}
\def\bfV{{\boldsymbol V}}
\def\bfW{{\boldsymbol W}}
\def\bfX{{\boldsymbol X}}

\newcommand{\R}{\mathbb{R}}

\newcommand{\be}{\begin{equation}}
\newcommand{\ee}{\end{equation}}

\def\eqref#1{equation~\ref{#1}}
\def\1{\bm{1}}

\DeclareMathAlphabet{\mathsfit}{\encodingdefault}{\sfdefault}{m}{sl}
\SetMathAlphabet{\mathsfit}{bold}{\encodingdefault}{\sfdefault}{bx}{n}

\usepackage[capitalize,noabbrev]{cleveref}

\theoremstyle{plain}
\newtheorem{theorem}{Theorem}[section]
\newtheorem{proposition}[theorem]{Proposition}
\newtheorem{lemma}[theorem]{Lemma}
\newtheorem{corollary}[theorem]{Corollary}
\newtheorem{condition}[theorem]{Condition}
\theoremstyle{definition}
\newtheorem{definition}[theorem]{Definition}

\theoremstyle{remark}
\newtheorem{remark}[theorem]{Remark}

\usepackage{color}
\newcommand{\MW}[1]{\textcolor{blue}{MW: #1}}
\newcommand{\mw}[1]{\textcolor{blue}{MW: #1}}
\newcommand{\ST}[1]{\textcolor{purple}{ST: #1}}
\newcommand{\HK}[1]{\textcolor{orange}{HK: #1}}

\usepackage[textsize=tiny]{todonotes}

\icmltitlerunning{How do Nonlinear Transformers Learn and Generalize in In-Context Learning?}

\begin{document}

\twocolumn[
%\icmltitle{Training Nonlinear Transformers for Efficient In-Context Learning: A Theoretical Learning and Generalization Analysis}
\icmltitle{How Do Nonlinear Transformers Learn and Generalize in In-Context Learning?}




\begin{icmlauthorlist}
\icmlauthor{Hongkang Li}{yyy}
\icmlauthor{Meng Wang}{yyy}
\icmlauthor{Songtao Lu}{comp}
\icmlauthor{Xiaodong Cui}{comp}
\icmlauthor{Pin-Yu Chen}{comp}
%\icmlauthor{Firstname6 Lastname6}{sch,yyy,comp}
%\icmlauthor{Firstname7 Lastname7}{comp}
%\icmlauthor{Firstname8 Lastname8}{sch}
%\icmlauthor{Firstname8 Lastname8}{yyy,comp}
\end{icmlauthorlist}

\icmlaffiliation{yyy}{Department of Electrical, Computer, and System Engineering, Rensselaer Polytechnic Institute, Troy, NY, USA}
\icmlaffiliation{comp}{IBM Thomas J. Watson Research Center, Yorktown Heights, NY, USA}
%\icmlaffiliation{sch}{School of ZZZ, Institute of WWW, Location, Country}

\icmlcorrespondingauthor{Hongkang Li}{lih35@rpi.edu}
\icmlcorrespondingauthor{Meng Wang}{wangm7@rpi.edu}
\icmlcorrespondingauthor{Songtao Lu}{songtao@ibm.com}
\icmlcorrespondingauthor{Xiaodong Cui}{cuix@us.ibm.com}
\icmlcorrespondingauthor{Pin-Yu Chen}{pin-yu.chen@ibm.com}

\icmlkeywords{Machine Learning, ICML}

\vskip 0.3in
]



\printAffiliationsAndNotice{}  % leave blank if no need to mention equal contribution

\begin{abstract}
  Transformer-based large language models have displayed impressive   in-context learning capabilities, 
  where a pre-trained model can handle new tasks without fine-tuning by simply augmenting the query with some input-output examples from that task. Despite the empirical success, the mechanics of how to train a Transformer to achieve ICL and the corresponding ICL capacity is mostly elusive due to the technical challenges of analyzing the nonconvex training problems resulting from the nonlinear self-attention and nonlinear activation in Transformers. To the best of our knowledge, this paper provides the first theoretical analysis of the training dynamics of Transformers with
nonlinear self-attention and nonlinear MLP, together with
the ICL generalization capability of the resulting model. Focusing on a group of binary classification tasks, we train Transformers using data from a subset of these tasks and quantify the impact of various factors on the ICL generalization performance on the remaining unseen tasks with and without data distribution shifts.   We also analyze how different components in the learned Transformers contribute to the ICL performance. 
Furthermore, we provide the first theoretical analysis of how model pruning affects ICL performance and prove that proper magnitude-based pruning can have a minimal impact on ICL while reducing inference costs.
These theoretical findings are justified through numerical experiments.

\end{abstract}

\vspace{-2mm}
\section{Introduction}

Transformers now serve as the backbone architecture for a wide range of modern, large-scale foundation models, including prominent language models like GPT-3 \citep{BMRS20}, PaLM \citep{CNDB22}, LLaMa \citep{TLIM23}, as well as versatile visual and multi-modal models such as CLIP \citep{RKHR21}, DALL-E \citep{RPGG21}, and GPT-4 \citep{gpt4}. One intriguing capability exhibited by certain large language models (LLMs) is known as  ``\textbf{in-context learning}'' (\textbf{ICL}) \citep{BMRS20}.  
Given a pre-trained model $F (\Psi)$,  parameterized by weights $\Psi$, the conventional approach fine-tunes  $\Psi$ separately for each downstream task using data from that task.  %  to meet different downstream tasks,
%while a recent intriguing concept called \textit{\textbf{in-context learning}} (ICL) allows $M_{\bm  \theta}$ % trained entirely in an unsupervised fashion 
In contrast,   ICL allows $F(\Psi)$ to handle multiple unseen tasks simultaneously without any fine-tuning. % with the same   weights $\bm  \theta$. 
%a recent  intriguing concept called \textit{in-context learning} allows a transformer-based foundation model trained entirely in an unsupervised fashion to  solve a large number of tasks without any fine-tuning of model parameters. S
\citet{GTLV22} is the first paper to mathematically formulate ICL. Briefly speaking, to predict $f(\bfx_{\textrm{query}})$ of a query input $\bfx_{\textrm{query}}$ for a new task represented by the label function $f$, 
 ICL augments $\bfx_{\textrm{query}}$ by $l$ example input-output pairs  $(\bfx_i, f(\bfx_i))_{i=1}^l$. %\PYB{Not clear what this function is doing; is it a task label function?} 
 The resulting so-called \textit{prompt} is sent to the model $F(\Psi)$, and, surprisingly, the model can output a prediction close to $f(x_{\textrm{query}})$. 
 %\textcolor{red}{XD: The input-output pair in the prompts and the final output of query may not be the same $f$ actually. The prompts provide a ground truth but output $f$ is a learned function of the mapping.}
  Thus, ICL is an efficient alternative to the resource-consuming fine-tuning methods. ICL has shown outstanding performance in multiple tasks in practice, including question answering \citep{LSZD22, WWYK23}, natural language inference \citep{LSSC22, WWYK23}, text generation \citep{BMRS20, LB21}, etc.   %\mw{ICL can handle both in-domain task and out-of-domain tasks. Need to explain what they are intuitively. }

%Specifically, these models can accurately predict outcomes for new tasks without fine-tuning their internal parameters. Specifically, ICL is implemented by adding $l$ pairs of example inputs and outputs $(\bfx_i, f(\bfx_i))_{i=1}^l$ as a \textit{prompt} before the \textit{query} $\bfx_{query}$ during the inference for an inference task function $f$, \HK{as shown in figure}. No model updates are needed in this process. This makes ICL an efficient and effective alternative to classical resource-consuming fine-tuning methods for LLM. \MW{What instructions? Prompt needs to be defined.}

\iffalse
 \begin{figure}[htbp]

   \centering
    \vspace*{-4mm}
     \hspace*{-3mm}
    \includegraphics[width=0.75\linewidth,height=1.4in]{figures/icl tf new.png}
    \vspace*{-3mm}
    \caption{\footnotesize{Example: ICL on English to German translation task}}
    \label{fig: icl general}
    \vspace*{-3mm}
 \end{figure}

%\mw{Too small to read the figure. Do not use wrap figure. 
%\mw{The output on the top should be only f(query). } 
\fi

In parallel, model pruning \citep{HMD15, WWWC16} can reduce the inference cost by removing some weights after training. It has been extensively evaluated in various applications.  Among various pruning techniques, such as gradient methods \citep{MTKA16} and reconstruction error minimization \citep{LWL17}, magnitude-based pruning \citep{WWWC16} is the most popular approach due to its simplicity and demonstrated promising empirical results. A few recent works \citep{FA23, MFW23, SLBK23, LWDZ23} also explore the pruning of LLMs to preserve their ICL capacity while accelerating the inference.

 \iffalse
 \citet{FA23} solves several sparse regression instances for sparsification. \citep{MFW23} adopts structural pruning to remove non-critical selectively
coupled structures based on gradient information. For magnitude-based methods, \citet{SLBK23} conduct pruning by the norm of weights and activations. \citet{LWDZ23} dynamically prune weights based on the input. 
 
 % is another technique that reduces the computation cost by removing partial model elements. 
  A few recent works also propose efficient pruning methods for LLMs, such as magnitude-based pruning methods \citep{SLBK23, LWDZ23}, so that the pruned model still has the ICL capability  \citet{MFW23, FA23}. %including those on .} \mw{If any of these methods use magnitude based pruning on any layer, need to mention}.
%\HK{\citet{SLBK23, MFW23, FA23} propose efficient pruning methods for in-context inference. } 
\fi

Despite the  empirical success of ICL, one fundamental and theoretical question   is less investigated, which is:
\begin{center}
\textit{How can a Transformer be trained to perform ICL and generalize in and out of domain successfully and efficiently?}
\end{center}
%\ST{We ask this question. How should we answer it? you need to use this question to introduce our work and then narrow down the scope of this work. This is a good opportunity to avoid some drawbacks of this work and guide the reviewer and audience to the problem that we really want to focus on.} 
%\ST{you can say either why or how. do you plan to emphasize both?}\HK{revised}
%Some recent works \citep{BCWX23, ASAM23} analyze this question by studying the expressive power of hierarchical Transformers to learn in context. 

  Some recent works attempt to answer this question for linear regression tasks \citep{LIPO23, ZFB23}. Specifically, \citet{LIPO23} investigate the generalization gap and stability of ICL. %\MW{Not sure what this means, treating the Transformer as an algorithm.} 
\citet{ZFB23} explore the training and generalization of ICL with Transformers, especially with distribution shifts during inference. \citet{WZCB23} studies the required number of pre-training tasks for a desirable ICL property. %in various conditions of ICL. 
\citet{HCL23} characterizes the training dynamics using Transformers with softmax attention and linear MLP. However, these results are either built upon simplified Transformer models by ignoring nonlinear self-attention \citep{ZFB23, WZCB23} or nonlinear activation in the multilayer perceptron (MLP) \citep{HCL23, ZFB23, WZCB23} or cannot characterize how to train a model to achieve the desirable ICL capability  with distribution-shifted data \citep{HCL23, LIPO23, WZCB23} . % to generalize to out-of-domain tasks \mw{Not clear to peoplle not the authors of these papers what data shift and task shift mean here. Any way to explain this limitation at a high level?.

\vspace{-2mm}
\subsection{Major Contributions of This Work}

%\ST{It seems that this part is a little bit redundant. We have already section 3.1. remark 3.6. I suggest shrinking this part a little bit, maybe half (give some space to the numerical part) and the full version to the appendix. It is common. }
%\MW{I assume our work is still the first if removing the setting of multi-task classification. I modified it as follows.}

To the best of our knowledge, our work is the first theoretical analysis of the training dynamics of Transformers with nonlinear self-attention and nonlinear MLP, together with the ICL generalization capability of the resulting model. Moreover, our paper provides the first theoretical analysis of the impact of model pruning on ICL performance.
% for both in-domiausing a nonlinear Transformer for classification problems.  %Moreover, different from the linear regression problem studied in xxxxxx, this paper also analyzes the ICL for multi-task classification for the first time.  
%\MW{add "nonlinear self attention" to distinguish with [23]?}
Focusing on a group of binary classification tasks, we show that training a Transformer using prompts from a subset of these tasks can return a model with the ICL capability to generalize to the rest of these tasks. %, not \HK{only} in the training data. 
We provide a quantitative analysis of the required number of training data, iterations, the length of prompts, and the resulting ICL performance. Although our analysis is centered on a simplified single-head and one-layer Transformer with softmax self-attention and ReLU MLP,   our theoretical insights shed light on practical architectures. Our major contributions include:

1. \textbf{A theoretical characterization of how to train Transformers to enhance their ICL capability.} We consider a data model where input data include both relevant patterns that determine the labels and irrelevant patterns that do not affect the labels. We quantify how the training and the resulting ICL generalization performance are affected by various factors, such as the magnitude of relevant features and the fraction of context examples that contain the same relevant pattern as the new query. In addition to proving the ICL capability of the learned Transformer to generalize to new binary tasks based on the relevant patterns that appear in the training data, we also prove the ICL capability to generalize to tasks based on patterns that are linear combinations of the relevant patterns and are unseen in the training data. 

2. \textbf{Expand the theoretical understanding of the mechanism of the ICL capability of Transformers.} We prove that when sending a prompt %including context examples and a query 
to a properly trained Transformer, the attention weights are concentrated on contexts that share the same relevant pattern as the query. Then, the ReLU MLP layer promotes %the impact of 
the label embedding of these examples, thus making the correct prediction for the query. Similar insights have appeared in \citep{HCL23}. We expand the analysis to Transformers with nonlinear MLP layers and new tasks with a data distribution shift.

3. \textbf{Theoretical justification of magnitude-based pruning in preserving ICL.} Based on the characterization of the trained Transformer, our paper also provides the first theoretical analysis of the ICL inference performance when the trained model is pruned by removing neurons in the MLP layer. We show that pruning a set of neurons with a small magnitude has little effect on the generalization while pruning the remaining neurons leads to a large generalization error growing with the pruning rate. To the best of our knowledge, no theoretical analysis exists on how model pruning affects ICL. 

\begin{table*}[htbp]
    \centering
    \begin{tabular}{M{2.8cm} M{1.6cm} M{1.6cm} M{1.8cm} M{2.0cm}  M{2.5cm}}
    \toprule
     Theoretical Works     & Nonlinear Attention & Nonlinear MLP & Training Analysis  & Distribution-Shifted Data & Tasks\\
    \hline
        \citet{LIPO23} & \checkmark & \checkmark &  & & linear regression\\
    \citet{ZFB23} & & &\checkmark &\checkmark  & linear regression\\
    \citet{HCL23} & \checkmark & &\checkmark  & & linear regression\\
    \citet{WZCB23} & & & \checkmark  & & linear regression \\
    \hline
        Ours & \checkmark & \checkmark & \checkmark  &\checkmark & classification
         \\
         \bottomrule
    \end{tabular}
    \caption{Comparison with existing works about training analysis and generalization guarantee of in-context learning}
    \label{tab:comparison}
    \vspace{-4mm}
\end{table*}

\vspace{-2mm}
\subsection{Related Work}

\textbf{Expressive power of ICL} Some existing works study the expressive power of Transformers to implement algorithms via ICL. \citet{ASAM23, VNRS23} demonstrate that Transformers conduct gradient descent during the forward pass of Transformers with prompts as inputs. %\MW{ICL with the prompt as the input?}.  \ST{Their mode is nonlinear, right? why do you put it in this paragraph?} \HK{I treat it as a work on expressive power} \ST{Use two paragraphs to discuss the expressiveness and linearity seperately} 
\citet{ACDS23, CCS23} extend the conclusion to preconditioned and functional gradient descent via ICL.  \citet{GTLV22, BCWX23, GHMW23} show the existence of Transformers that can implement a broad class of machine learning algorithms in context.

\textbf{The optimization and generalization of Transformers} Beyond in-context learning, there are several other works about the optimization and generalization analysis of fine-tuning or prompt tuning on Transformers. \citet{JSL22, LWLC23, LWMS23, luo2024enhancing} study the generalization of one-layer Transformer by assuming spatial association or the majority voting of tokens. %\citet{LWMS23} investigate the effect of the relative positional encoding in training for Graph Transformer.
\citet{LLR23} delve into how one-layer Transformers learn semantic structure. \citet{ORST23} depict the trajectory of prompt tuning of attention networks. \citet{TLZO23, TLTO23} characterize that the gradient updates of the prompt or weights converge to a max-margin SVM solution.   \citet{TWCD23, TWZC23} probe the training dynamics of Transformers for the next token prediction problem given infinitely long sequences.

 \textbf{Theoretical generalization analysis of pruning}
%\HK{does pruning on LLM an repeat to the paragraph between line 30-36 right?}
%Pruning methods for LLM \citep{} refer to removing partial of the model to improve inference efficiency.
%\citet{FA23} solves several sparse regression instances for sparsification. \citep{MFW23} adopts structural pruning to remove non-critical selectively coupled structures based on gradient information. For magnitude-based methods, \citet{SLBK23} conduct pruning by the norm of weights and activations. \citet{LWDZ23} dynamically prune weights based on the input. Except for the above experimental works, Some recent literature focuses on the theoretical analysis of generalization of the pruned model \citep{ZWLC21_sparse, ZWCL23, YLGW23}. 
A few recent works consider analyzing the generalizations performance of model pruning theoretically. For example, 
\citet{ZWLC21_sparse} study the sample complexity of training a pruned network with a given sparse ground truth weight. \citet{YW23} investigate the neural tangent kernel of the pruned model. \citet{ZWCL23, YLGW23} consider the generalization using magnitude pruning under a feature learning framework.
However, these works are built on convolutional neural networks, and no theoretical works are for LLM or Transformer-based models.  

%\mw{which are experimental? which theoretical?}\HK{revised}

\vspace{-2mm}
\section{Problem Formulation}\label{sec: problem formulation}
%\vspace{-2mm}

This work studies the optimization and generalization of binary classification problems for in-context learning. Consider a query $\bfx_{query}$ and its label $z$.  %where $\bfx_{query}$ follows the distribution $\mathcal{D}$. 
Define a set of binary classification tasks $\mathcal{T}$, consisting of multiple task functions. The label $z \in\{+1,-1\}$ is mapped from $\bfx_{query}\in\mathbb{R}^{d_\mathcal{X}}$ through a task $f$ that is randomly chosen from $\mathcal{T}$, i.e., $z=f(\bfx_{query})\in\{+1,-1\}, f\in\mathcal{T}$. %\ST{maybe we can use a figure or one example to explain/define this setting.}

\vspace{-2mm}
\subsection{Training to Enhance ICL Capability} \label{subsec: formualtion_training}
Following the framework of training for ICL in \citep{GTLV22, ASAM23, BCWX23}, we consider the problem of training such that the model has the ICL capability to generalize to new tasks using prompts. % a model that can implement ICL by %the training dataset $\{\bfP^n, z^n\}_{n=1}^N$, where $\bfP$ is
The idea is to update the model during the training process using pairs of the constructed prompt, embedded as $\bfP$ for the query $\bfx_{query}$, and its label $f(\bfx_{query})$. % in ICL under a supervised learning setup. 
We start by formulating $\bfP$ and then introduce the learning model in this section. 

%In ICL \citep{GTLV22}, we learn the data by prepending a prompt to each query to enable a correct prediction in different tasks. 
Following \citep{VNRS23, ZFB23, HCL23}, the prompt  embedding  $\bfP$ of query % we denote such formulated data for 
$\bfx_{query}$ is formulated as: %as $\bfP$:
\begin{equation}
\vspace{-2mm}
\begin{aligned}
    \bfP=&\begin{pmatrix}
    \bfx_1 & \bfx_2 &\cdots & \bfx_l & \bfx_{query}\\
    \bfy_1 & \bfy_2 &\cdots & \bfy_l & \bf{0}
    \end{pmatrix}\\
    :=&(\bfp_1, \bfp_2, \cdots, \bfp_{query})\in\mathbb{R}^{(d_\mathcal{X}+d_\mathcal{Y})\times(l+1)},\label{eqn: data}
\end{aligned}
\end{equation}
where the last column of $\bfP$, denoted by   $\bfp_{query}$, includes the query $\bfx_{query}$ with padding zeros, and the first $l$ columns are the  contexts  for $\bfx_{query}$. %\MW{Still feels that prompt includes the query.} \HK{I checked literature. Two called it the embedding matrix of the prompt. One called it input sequence. One is unclear. What about ``prompt embedding'' for this P? For the first l columns, we can call it ``contexts''. One paper did so.} 
We respectively call $\bfx_i$ and $\bfy_i$, $i\in[l]$ \textit{context} inputs and outputs, %in the prompt. 
%$\bfx_i\in\mathbb{R}^{d_\mathcal{X}}$ follows the same distribution as the corresponding $\bfx_{query}$. \MW{The distributions of xi and xquery are different. xi depends on x query and alpha. remove this sentence.} 
where $l$ is also known as the prompt length. Let $\text{Embd}(\cdot)$ be the embedding function of each context output. $\bfy_i\in\mathbb{R}^{d_\mathcal{Y}}$ in (\ref{eqn: data}) is defined as $\bfy_i=\text{Embd}(f(\bfx_i))$. %, given $f$ as the task chosen from $\mathcal{T}$ for $\bfx_{query}$ and $y$. 
Hence, $\bfP$ is a function of $f$. The first $d_\mathcal{X}$ dimensions of $\bfp_i$ are referred to as the feature embedding, while the last $d_\mathcal{Y}$ dimensions are called the label embedding. % for $i\in[l]$. 

The \textbf{learning model} is a single-head, one-layer Transformer with one self-attention layer and one two-layer perceptron. Mathematically, it can be written as 
\vspace{-2mm}
\begin{equation}
\begin{aligned}
    &F(\Psi;\bfP)=\bfa^\top\text{Relu}(\bfW_O\sum_{i=1}^l \bfW_V \bfp_i \cdot \text{attn}(\Psi; \bfP, i)),\\
    &\text{attn}(\Psi; \bfP, i)=\text{softmax}((\bfW_K\bfp_i)^\top\bfW_Q\bfp_{query}),\label{eqn: transformer}
\end{aligned}
\end{equation}
where $\bfW_Q,\bfW_K\in\mathbb{R}^{m_a\times (d_\mathcal{X}+d_\mathcal{Y})}$, $\bfW_V\in\mathbb{R}^{m_b\times (d_\mathcal{X}+d_\mathcal{Y)}}$ are the embedding matrices for queries, keys, and values, respectively, and $\bfW_O\in\mathbb{R}^{m\times m_b}$ and $\bfa\in\mathbb{R}^m$ are parameters in the MLP layer. Here, $\text{softmax}((\bfW_K\bfp_i)^\top\bfW_Q\bfp_{query})=e^{(\bfW_K\bfp_i)^\top\bfW_Q\bfp_{query}}/\sum_{j=1}^l e^{(\bfW_K\bfp_j)^\top\bfW_Q\bfp_{query}}$. $\Psi:=\{\bfW_Q$, $\bfW_K, \bfW_V, \bfW_O, \bfa\}$ denotes the set of all model weights. Typically, $\min(m_a, m_b)>d_\mathcal{X}+d_\mathcal{Y}$.

The \textbf{training problem} to enhance the ICL capability solves the empirical risk minimization problem,  % minimizes the empirical risk loss $R_N(\Psi)$, 
\vspace{-2mm}
\begin{equation}
    \min_\Psi R_N(\Psi):=\frac{1}{N}\sum_{n=1}^N \ell(\Psi;\bfP^n,z^n),\label{eqn: loss_ERM}
\end{equation}
%\MW{The training problem is not typical training, so it deserves a separate line.} %where $\Psi$ is the parameter in the Transformer.
using $N$ pairs of prompt embedding and label pairs $\{\bfP^n,z^n\}_{n=1}^N$. For the $n$-th pair,  $\bfx_{query}^n$ and the context input $\bfx_i^n$ are all sampled from an unknown distribution $\mathcal{D}$, the task $f^n$ is sampled from $\mathcal{T}$, and $\bfP^n$ is constructed following (\ref{eqn: data}).  The loss function is a Hinge loss, i.e., $\ell(\Psi;\bfP^n,z^n)=\max\{0,1-z^n \cdot F(\Psi;\bfP^n)\}$, where $F(\Psi; \bfP^n)$ is defined in (\ref{eqn: transformer}). Let $\mathcal{T}_{tr} =\bigcup_{n=1}^N f^n$ denote the set of tasks that appear in the training samples. Note that $\mathcal{T}_{tr} \subset \mathcal{T}$, and 
%We denote the distribution of training queries as $\mathcal{D}$. The task $f^n$ of the training prompt embedding $\bfP^n$ for all $n\in[N]$ composes $\mathcal{T}_{tr}$, which is a subset of $\mathcal{T}$.  Since there are multiple tasks in $\mathcal{T}_{tr}$, 
 (\ref{eqn: loss_ERM}) is a \textit{multi-task learning} problem when $|\mathcal{T}_{tr}|>1$. 

\vspace{-2mm}
\subsection{Generalization Evaluation} 
We define two quantities to evaluate the ICL generalization performance to new tasks as follows.

\textbf{In-Domain Generalization}: %During evaluation, 
If the testing queries are also drawn from $\mathcal{D}$ % as the training queries 
and all the testing tasks are drawn from $\mathcal{T}$, we call it \textit{in-domain} inference, and the in-domain generalization error is defined as\footnote{In terms of evaluating generalization on unseen tasks, (\ref{eqn: id_generalization}) is almost equivalent to replacing $f\in \mathcal{T}$ with $f\in \mathcal{T}\backslash \mathcal{T}_{tr}$ in the subscript. This is because we later prove that all of our analysis can hold when training on a small fraction of tasks (Condition \ref{cond: task}). Therefore, an $\mathcal{O}(\epsilon)$ generalization error on $f\in \mathcal{T}$ can indeed reflect an $\mathcal{O}(\epsilon)$ generalization error on $f\in \mathcal{T}\backslash\mathcal{T}_{tr}$} %(\ref{eqn: id_generalization}) is then defined based on the testing prompt embedding $\bfP$ as constructed in (\ref{eqn: data}). 
\vspace{-2mm}
\begin{equation}
   \underset{\bfx_{query}\sim\mathcal{D}, f\in\mathcal{T}}{\mathbb{E}} [\ell(\Psi; \bfP, z)],\label{eqn: id_generalization}
\end{equation}
where $\bfP$ is defined in (\ref{eqn: data}). Note that the in-domain performance includes the testing performance on \textit{unseen} tasks in $\mathcal{T}\backslash \mathcal{T}_{tr}$ that do not appear in the training samples.

\textbf{Out-of-Domain Generalization}: Suppose the testing queries $\bfx_{query}$ follow  the distribution $\mathcal{D}'$ ($\mathcal{D}'\neq \mathcal{D}$), and the binary classification tasks that map the testing queries to the labels are drawn %from a new task
a set $\mathcal{T}'$ ($\mathcal{T}'\neq \mathcal{T}$). %Let $f$ denote the corresponding task for $\bfx_{query}$. Then, by formulating $\bfP$ as in (\ref{eqn: data}) for testing, 
Then, the \textit{out-of-domain} generalization error can be defined as 
\vspace{-2mm}
\begin{equation}
    \underset{\bfx_{query}\sim\mathcal{D}', f\in\mathcal{T}'}{\mathbb{E}}[\ell(\Psi; \bfP, z)].\label{eqn: ood_generalization}
\end{equation}

\vspace{-2mm}
\subsection{Training Algorithm}\label{subsec: SGD}
The model is trained using stochastic gradient descent (SGD) with step size $\eta$ with batch size $B$, summarized in Algorithm \ref{alg: sgd} in Appendix \ref{sec: more experiment}. %The training data $\bfP$ in each batch is sampled from the distribution where $\bfx_i, \bfx_{query}\sim\mathcal{D}$. 
%\MW{ Do not know that this means, maybe could be removed . We denote the set of training tasks composed of }%Let $\bfI_d$ be an $d\times d$ identity matrix and $\delta\in(0,0.1)$. $\bfW_Q^{(0)}$, $\bfW_K^{(0)}$ are initialized from $\delta\bfM$, where $\bfM[0:d_\mathcal{X},0:d_\mathcal{X}]=\bfI_{d_{\mathcal{X}}}$ and all other entries are $0$. $\bfW_V^{(0)}$ is initialized from $\delta\bfI_{d_{\mathcal{X}}+d_{\mathcal{Y}}}$.
$\bfW_Q$, $\bfW_K$ and $\bfW_V$ are initialized such that all diagonal entries of $\bfW_V^{(0)}$, and the first $d_\mathcal{X}$ diagonal entries of $\bfW_Q^{(0)}$ and $\bfW_K^{(0)}$ are set as $\delta$ with $\delta\in(0,0.2]$, and all other entries are $0$. Each entry of $\bfW_O^{(0)}$ is generated from $\mathcal{N}(0,\xi^2), \xi=1/\sqrt{m}$ and each entry of $\bfa$ is uniformly sampled from $\{1/m, -1/m\}$. Besides, $\bfa$ does not update during training. 

\vspace{-2mm}
\subsection{Model Pruning }
We also consider the case that the learned model $\Psi$ is pruned to reduce the inference computation. Let $\mathcal{S} \subset [m]$ denote the index set of neurons in the output layer. Pruning neurons in $\mathcal{S}$ correspond to removing the corresponding rows in $\bfW_O$, resulting in the reduced matrix size of $(m-|\mathcal{S}|)\cdot m_b$. 

\vspace{-2mm}
\section{Theoretical Results}

\begin{figure*}[ht]
%\vspace*{-4mm}
\centering
\centerline{
\begin{tabular}{cc}
\includegraphics[width=.35\textwidth,height=1.05in]{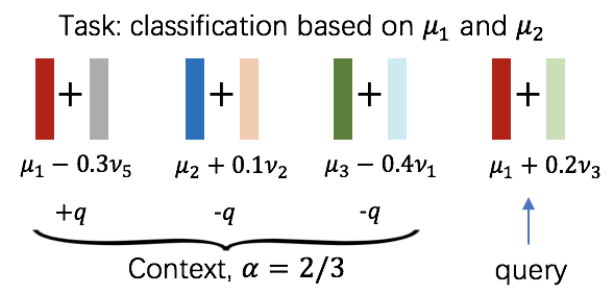}  
&
\hspace*{-1mm}
\includegraphics[width=.35\textwidth,height=1.05in]{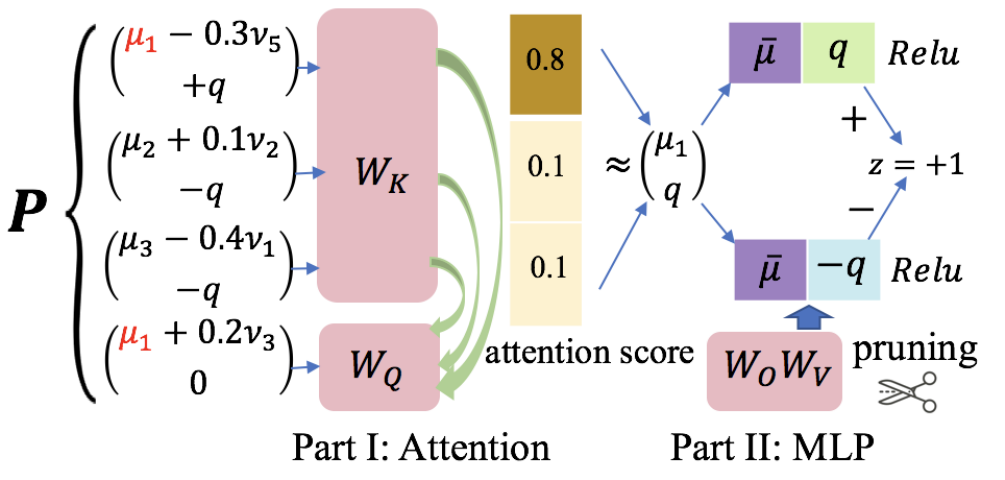}\vspace*{-1mm}
\\
(A) 
& \hspace*{-2mm} (B)
% & \hspace*{-3mm} (C)
\end{tabular}}
\vspace*{-3mm}
\caption{\footnotesize{(A) Example of prompt embedding. $l=3$, $\alpha=2/3$.   (B) The mechanism of a trained Transformer (\ref{eqn: transformer}) to implement ICL. Part I: The attention layer assigns the largest attention score (0.8) on $\bfmu_1-0.3\bfnu_5$, which has the same IDR pattern as the query. Then the weighted sum of input tokens is close to $(\bfmu_1^\top,\bfq^\top)^\top$ by the trained attention layer. Part II: The  neurons in $\bfW_O\bfW_V$ with a large magnitude % trained MLP layer makes large neurons in $\bfW_O\bfW_V$ close to be in directions of 
are aligned with $\bar{\bfmu}$ and $\pm\bfq$ in the first $d_\mathcal{X}$ and the rest  $d_\mathcal{Y}$ dimensions, respectively. Then the prediction is based on the part of $\pm\bfq$ that varies for different queries rather than the part of $\bar{\bfmu}$ that is universal for all IDR patterns.}}
%This enables the MLP layer to make predictions based on the label embedding rather than the feature embedding of the input.}}}
\vspace*{-4mm}
\label{fig: overview}
\end{figure*}
%\vspace{-2mm}
We first summarize the main insights in Section \ref{subsec: insight}. Section \ref{sec: data task} formally presents our analysis model. Section \ref{subsec: in-domain-gen} presents the formal theoretical results on the learning performance and the resulting ICL generalization. Section \ref{subsec: prune} provides the theoretical result that magnitude-based pruning on the out layer does not hurt ICL performance. 

\vspace{-2mm}
\subsection{Main Theoretical Insights}\label{subsec: insight}
%\vspace{-2mm}
%\HK{Before formally presenting the theoretical results, we summarize the main insights as follows. We specifically formulate a data model where in/out-of-domain-relevant patterns in the prompts are decisive patterns for predictions. Denote $\beta$ as the magnitude of in/out-of-domain-relevant patterns, and $\alpha$ and $\alpha'$ as the probability of contexts with in- and out-domain-relevant patterns defining the task for each prompt.}

We consider a class of binary classification tasks where the binary labels in each task are determined by two out of   $M_1$ \textit{in-domain-relevant patterns}. The training data include pairs of prompt embedding and labels from a small subset of these tasks. In-domain generalization evaluates the ICL capability of the learned model on tasks using all possible combinations of these $M_1$   patterns. Out-of-domain generalization further evaluates the binary classification tasks that are determined by pairs of  \textit{out-of-domain-relevant patterns}, which are some linear combinations of these $M_1$ patterns. 

%\textbf{Quantivel Learnig Analysis with Guaranetted In- and Out-of-Domain Generalization}. We quantivetly prove that the leared mdoo achiees desriable genaraltion in both In-domain and Out-of-domain tasks. THe required number of trainig data and training iterations both are shown to be polynomail in $\beta^{-1}$ and $\alpha^{-1}$, where $\beta$ is the norm of the relavant features, and $\alpha$ is the fraction of context examples from the same task as the query. A larger $\alpha$ meansa the context exampls provide more context example about the query, which in term reduces  the reduces the sample requiremetns and  speeds up leanring/ 

\textbf{P1. Quantitative Learning Analysis With Guaranteed In- and Out-of-Domain Generalization}. 
We quantitatively prove the learned model achieves desirable generalization in both in-domain and out-of-domain tasks. The required number of training data and iterations are polynomial in $\beta^{-1}$ and $\alpha^{-1}$, where $\beta$ represents the norm of relevant patterns, and $\alpha$ denotes the fraction of context inputs with the same in-domain-relevant pattern as the query. A higher $\alpha$ implies that the context examples offer more information about the query, consequently reducing the sample requirements and expediting the learning process.

\textbf{P2. Mechanism of Transformers in Implementing ICL}. %We show that if training tasks can cover all task-relevant features balancedly, testing on any unseen task based on two task-relevant features can return a zero generalization error. The required length of the testing prompt is inversely associated with the fraction of IDR patterns in the testing prompt.
 We elucidate the mechanism where the learned Transformers make predictions in- and out-of-domain in context. We quantitatively show that the self-attention layer attends to context examples with relevant patterns of the query task and promotes learning of these relevant patterns. Then, the two-layer perceptron promotes the label embeddings that correspond to these examples so as to predict the label of the query accurately.  

\textbf{P3. Magnitude-Based Pruning Preserves ICL}. We quantify the ICL generalization if neurons with the smallest magnitude after training in the MLP layer are removed and prove that the generalization is almost unaffected even when a constant fraction of neurons are removed. In contrast, the generalization error is proved to be at least $\Omega(R)$ when $R$ fraction of neurons with large magnitude are removed.

%\HK{We first theoretically show that the trained neurons can be divided into two groups by magnitude. Then, we quantitatively conclude that pruning neurons with a small magnitude can generalize almost as well as no pruning, but pruning large neurons leads to an increasing generalization error as pruning more. Meanwhile, the generalization error is linear in ${\alpha'}^{-1}$.} 

%\textbf{P4. Mechanism of Transformers in Implementing ICL.} %We show that if training tasks can cover all task-relevant features balancedly, testing on any unseen task based on two task-relevant features can return a zero generalization error. The required length of the testing prompt is inversely associated with the fraction of IDR patterns in the testing prompt.
%\HK{We elucidate the mechanism where Transformers make predictions in- and out-of-domain in context. Transformers first promote the magnitude of in/out-of-domain-relevant features through self-attention to select intimate contexts with the same in/out-of-domain-relevant pattern as the query. Subsequently, the trained model projects the resulting embeddings, using the MLP layer mainly based on the label embedding part, onto positive or negative scalars to make predictions.} %Such a mechanism differs from existing works \citep{LWLC23, ORST23, TWCD23} on single tasks. %\MW{Isn't this the same for typical learning? What's special about in-context learning? What is special about the mechanism so that it can get a model that has P3 property?}

\vspace{-2mm}
\subsection{The Modeling of Training Data and Tasks}\label{sec: data task}
%\vspace{-2mm}

\iffalse 
\MW{Break into multiple parts.
1. Data model.
2. Task. Here, only define one task on two features, not all features.
3. Training Prompt/ Testing prompt. Here mention for features not defined in a task, random labels}
Before formally presenting the main results, we first introduce the models for data and tasks.
\fi

\textbf{In-Domain Data and Tasks.} Consider $M_1$ \textit{in-domain-relevant} (\textit{IDR}) patterns $\{\bfmu_j\}_{j=1}^{M_1}$ and $M_2\ (= \mathcal{O}(M_1))$ \textit{in-domain-irrelevant} (\textit{IDI}) patterns $\{\bfnu_k\}_{k=1}^{M_2}$ ($M_1+M_2=d_\mathcal{X}$)   in $\mathbb{R}^{d_\mathcal{X}}$, where these $M_1+M_2$ patterns are pairwise orthogonal, and $\|\bfmu_j\|=\|\bfnu_k\|=\beta\geq 1$ ($\beta$ is a constant) for $j\in[M_1]$,$k\in[M_2]$.  Each in-domain data $\bfx$ drawn from $\mathcal{D}$ is generated by %Each query $\bfx_{query}$ is generated by 
\iffalse 
To be more specific, let $M_1\ (2\leq M_1\leq m_a, m_b)$ denote the number of \textit{IDR} (short for \textit{in-domain-relevant}) %\MW{not sure if it is a standard terminology. If not, revise.}
patterns represented by $\{\bfmu_1,\bfmu_2,\cdots,\bfmu_{M_1}\}$ and $M_2\ (2\leq M_2\leq m_a, m_b)$ as the number of \textit{IDI} (short for \textit{in-domain-irrelevant}) patterns represented by $\{\bfnu_1,\bfnu_2,\cdots,\bfnu_{M_2}\}$ in $\mathbb{R}^{d_\mathcal{X}}$. Here, $\{\bfmu_1, \cdots,\bfmu_{M_1},\bfnu_1,\cdots, \bfnu_{M_2}\}$ forms an orthogonal set. %$\bfmu_i\perp\bfmu_j$ for $1\leq i\neq j\leq M_1$, $\bfnu_i\perp\bfnu_j$ for $1\leq i\neq j\leq M_2$, and $\bfmu_i\perp\bfnu_j$ for $1\leq i\leq M_1$ and $1\leq j\leq M_2$.
Also, $\|\bfmu_i\|=\|\bfnu_j\|=\beta\geq 1$ for $i\in[M_1],\ j\in[M_2]$. Then, each in-domain data $\bfx$ drawn from $\mathcal{D}$ $\bfx_{query}^n$ satisfies that for a certain $j\in[M_1]$ and $k\in[M_2]$,
%\MW{ideally, use  index i for x, j for mu and k for v. You could say mu j1 and mu j2 are orthogonal.}
%%\vspace{-1mm}
\fi  
\begin{equation}\bfx =\bfmu_{j}+\kappa \bfnu_{k},%\bfx_{query}=\bfmu_{j}+\kappa_{query}\bfnu_{k}
\label{eqn: x_mu_nu_tr}
\end{equation}

where $j\in[M_1]$ and $k\in[M_2]$ are arbitrarily selected.   
%\MW{j(i) is not rigorously, as j is a function of $x_i$, not i, but $j(x_i)$ looks messier. }
  %$\kappa_{query}$
  $\kappa$ follows a uniform distribution $\textit{U}(-K,K)$, $K\leq 1/2$.  Denote IDR($\bfx$)$:=\bfmu_j$ as the IDR pattern in data $\bfx$.  Our data assumption originates from recent feature learning works on deep learning \citep{AL23, LWLC23, ORST23} for language and vision data. To the best of our knowledge, only \citep{HCL23} theoretically analyzes the performance of ICL with softmax attention, assuming all $\bfx$ are orthogonal to each other. Our assumption in (\ref{eqn: x_mu_nu_tr}) is more general than that in \citep{HCL23}. %\HK{During training, each batch is sampled such that the number of queries of each IDR pattern is the same.}% does not require different data $\bfx$ to be orthogonal to each other.   
   %, and $\bfn_{query}^n$ is a random noise that is bounded with $\|\bfn_{query}^n\|\leq \tau$.
%\MW{Do you need unform distribution on mu and uniform distribution on nv? Or do you mean any distribution on mu and nv? We need to specify this. Otherwise, D is not clearly defined.}\HK{only kappa}

%The in-domain data distribution $\mathcal{D}$ is then defined as the distribution where the data can be represented as in (\ref{eqn: x_mu_nu_tr}). %All tasks are binary classification problems on two certain different IDR patterns. Hence, there are $M_1(M_1-1)$ tasks in total. 
%\ST{Need to comment the relation between this data model and the existing ones.}
\iffalse
Let
%%\vspace{-1mm}
\begin{equation}
    \lambda_*=\min\{\lambda_i^n, n\in[N], i\in[l+1]\}>0
\end{equation}
\fi
% \MW{The third definition of yin, make them consistent.} %where $\bfq$, $-\bfq$ represent the label embeddings for labels $+1$ and $-1$, respectively. $\|\bfq\|=1$.

%\vspace{-1mm}
%\textbf{In-domain Tasks.} 
Each in-domain task is defined as a binary classification function that decides the label based on two IDR patterns in the query. Specifically, 
\begin{definition}\label{def: task} (Definition of in-domain tasks) The in-domain task set $\mathcal{T}$ includes 
$M_1(M_1-1)$ tasks such that each task  $f\in\mathcal{T}$  is defined as
\vspace{-2mm}
\begin{equation}
    f(\bfx)=\begin{cases}+1, &\text{IDR}(\bfx)=\bfmu_a,\\
    -1, &\text{IDR}(\bfx)=\bfmu_b, \\
    \text{random from} \{+1,-1\}, &\text{otherwise},
    \end{cases}\label{eqn: def_task}
\end{equation}
where $\bfmu_a, \bfmu_b$ are two different patterns in  $\{\bfmu_j\}_{j=1}^{M_1}$ and are the decisive patterns for task $f$. 
\end{definition}
From (\ref{eqn: def_task}), the task $f$ outputs label +1 (or -1) if the IDR pattern is $\bfmu_a$ (or $\bfmu_b$). If the data contains neither of these two patterns, the label is random. 

\iffalse
$f\in\mathcal{T}$ is defined in the following two steps: (1) Randomly select $\bfmu_a, \bfmu_b$ to be IDR patterns that $f$ is based on, where $\bfmu_a, \bfmu_b$ correspond to $+1,-1$, respectively, $1\leq a,b\leq M_1$. (2) Let $\bfmu_j$ be the IDR pattern of the input $\bfx$ for $f(\cdot)$, we have
\begin{equation}
    f(\bfx)=\begin{cases}+1, &\text{If }\bfmu_j=\bfmu_a\\
    -1, &\text{If }\bfmu_j=\bfmu_b
    \end{cases}\label{eqn: def_task}
\end{equation}
If $j\notin\{a,b\}$, $f(\bfx)$ is randomly chosen from $\{+1,-1\}$ with half probability. We can deduce that the total number of in-domain tasks is $|\mathcal{T}|=\binom{M_1}{2}\cdot 2!=M_1(M_1-1)$. 
\fi

%that respectively maps inputs with $\bfmu_a$ and $\bfmu_b$, $1\leq a\neq b\leq M_1$, to $+1$ and $-1$, we have $f^{(n)}(\tilde{\bfx}_i^n)$ (including $z^n$) is $+1$ (or $-1$) if $j=a$ (or $j=b$) in (\ref{eqn: x_mu_nu_tr}). If the IDR pattern in $\bfx_i^n$ is neither $\bfmu_{a}$ nor $\bfmu_{b}$, the label of $\bfx_i^n$ is randomly chosen from $\{+1,-1\}$ with equal probability. 

%\vspace{-1mm}

\textbf{Out-of-Domain Data and Tasks}. Assume there are $M_1'$ \textit{out-of-domain-relevant} (\textit{ODR}) patterns $\{\bfmu_j'\}_{j=1}^{M_1'}$ and $M_2'$  \textit{out-of-domain-irrelevant}  (\textit{ODI}) patterns  $\{\bfnu_k'\}_{k=1}^{M_2'}$. %Each pattern has norm $\beta$ and is orthogonal to other patterns. 
Any data $\bfx$ dawn from $\mathcal{D}'$  can be generated by 
\vspace{-2mm}
\begin{equation}
    \bfx = \bfmu_{j}'+\kappa' \bfnu_{k}'\label{eqn: x_mu_nu_tt}
\end{equation}
%\MW{This seems to imply j and k are the same in training and testing data. Back to my earlier comment, j and k are not functions of i.}
where $j\in[M_1']$ and $k\in[M_2']$ are arbitrarily selected, and $\kappa' \sim\textit{U}(K',K')$ for $K'= \mathcal{O}(1)$. We use ODR($x$)$:=\bfmu'_j$ to denote the ODR pattern of $\bfx$. %\HK{reivsed} 
%is a random noise with $\|\bfo_{query}^n\|\leq \tau$. 

 The set of out-of-domain tasks  $\mathcal{T}'$ contains $M_1'(M_1'-1)$ binary classification problems that are defined in the same fashion as Definition \ref{def: task}, with the only difference of using  $\{\bfmu_j'\}_{j=1}^{M_1'}$ rather than $\{\bfmu_j\}_{j=1}^{M_1}$ to determine labels.

\textbf{Prompt Construction for Training and Testing}. Let $l_{tr}$ and $l_{ts}$ denote the length of training and testing contexts, respectively.

\textit{Training prompt embedding}: Given an input-label pair $\bfx_{query}$ and $f(\bfx_{query})$ for training,  the context inputs $\bfx_i$ in  $\bfP$ in (\ref{eqn: data}) are constructed as follows. The IDR pattern is selected from $\{\bfmu_j\}_{j=1}^{M_1}$ following a categorical distribution parameterized by $\alpha$, where where $\alpha=\Theta(1)\in(0,1]$. Specifically, each of  $\bfmu_{a}$ and $\bfmu_{b}$ (the decisive patterns of task $f$) is selected with probability $\alpha/2$, and each of these other $M_1-2$ patterns elected with probability $(1-\alpha)/(M_1-2)$. The context labels are determined by task $f$. 

\textit{Testing prompt embedding}: The context inputs for the testing query can be selected following a wide range of prompt selection methods \citep{LSZD22, RHB22, WWYK23}. %\mw{Maybe add a few references here?}  
Given an in-domain  (or out-of-domain) task $f$ that has decisive patterns 
  $\bfmu_{a}$ and $\bfmu_{b}$ (or  $\bfmu_{a}'$ and $\bfmu_{b}'$), %the IDR ( or ODR) pattern of the context inputs for the testing query is selected following a rule that 
  we only assume at least $\alpha'/2$ ($\alpha'\in(0,1]$) fraction of context inputs contain the same IDR (or ODR) pattern as the query. %\HK{revised}
For the label embedding $\bfy_i$ for both training and testing, %we consider the embedding function $\text{Embd}(\cdot)$ defined in the way that 
$\text{Embd}(+1)=\bfq$, $\text{Embd}(-1)=-\bfq$, where $\bfq\in\mathbb{R}^{d_\mathcal{Y}}$. %$\bfq\perp\{\bfmu_1,\cdots,\bfmu_{M_1},\bfnu_1,\cdots,\bfnu_{M_2}\}$. \MW{They are different dimension, right?} 
Hence, $\bfy_i\in\{\bfq, -\bfq\}$ for $i\in[l_{tr}]$ or $i\in[l_{ts}]$.

\iffalse
\begin{equation}
    \bfc_i^\top\bfc_{l+1}=\kappa(\frac{2(i-1)}{l}-1)
\end{equation}
which uniformly ranges from $-\kappa$ to $\kappa$ for $i\in[l+1]$ where $\kappa\in(0,1)$.

We first define \textit{neighbor contexts} as $\bfx_i^n$ that satisfies $\|\bfx_i^n-\bfx_{l+1}^n\|\leq r<\tau$ where $i\in[l]$ and $n\in[N]$. Denote the set of neighbor contexts of the training data $\bfP^n$ as $\mathcal{N}^n$. The indices of neighbor contexts are uniformly distributed in $[l]$. For each training data $\bfP^n$, we intuitively define that the labels of all tasks are determined by either the neighbor contexts or the position of contexts. Specifically, for the $\lambda$ ratio of training data, we assume the existence of at least one neighbor context in each data.   We require that the number of neighbor contexts with labels equal to $z^n$ is larger than those with labels opposite to $z^n$, i.e., 
\begin{equation}
    |\{i|\|\bfx_i^n-\bfx_{l+1}^n\|\leq r, y_i^n=y_{l+1}^n, i\in[l]\}|>|\{i|\|\bfx_i^n-\bfx_{l+1}^n\|\leq r, y_i^n=-y_{l+1}^n, i\in[l]\}|
\end{equation}
This provides the correct label by the majority voting of the neighbor labels. For the remaining $1-\lambda$ ratio of training data, the label is the same as that of the $s_*$-th context in the prompt for all tasks, where $s_*\in[l]$.
\fi

\vspace{-2mm}
\subsection{In-Domain and Out-of-Domain Generalization With Sample Complexity Analysis}\label{subsec: in-domain-gen}
%\vspace{-2mm}
%In-domain generalization means the testing data follows the same distribution as the training data. 
%\MW{The distribution should be defined. It is over x rather than P. The generalization definition should appear before section 3, like my earlier comment.}

%Define the set of training tasks $\mathcal{T}_{tr}=\{f^1,f^2,\cdots,f^N\}$. Since that $f_i\in\mathcal{T}$ for all $i\in[N]$, we have $\mathcal{T}_{tr}\subset\mathcal{T}$. 
In order for the learned model $F(\Psi)$ to generalize all tasks in $\mathcal{T}$ through ICL, the training tasks in $\mathcal{T}_{tr}$ should uniformly cover all the possibilities of IDR patterns and labels, as stated by the following condition, 
%To avoid bias in the number of different tasks, we need the training tasks to cover all IDR patterns and labels uniformly. Therefore, we propose the following condition to hold.
\begin{condition}\label{cond: task}
  For any given $j\in[M_1]$ and either label $+1$ or $-1$, the number of tasks in $\mathcal{T}_{tr}$ that map $\bfmu_j$ to that label is $|\mathcal{T}_{tr}|/M_1(\geq 1)$. 
\end{condition}
Note that Condition \ref{cond: task} is easy to meet, and $|\mathcal{T}_{tr}|$ does not have to be large. In fact,  $|\mathcal{T}_{tr}|$ can be as small as $M_1$. For example, let
%An example that satisfies Condition \ref{cond: task} is as follows. 
the $i$-th task function ($i\in[M_1-1]$) in $\mathcal{T}_{tr}$ map  the queries with $\bfmu_i$ and $\bfmu_{i+1}$ as IDR patterns to $+1$ and $-1$, respectively. The $M_1$-th task function maps $\bfmu_{M_1}$ and $\bfmu_1$ to $+1$ and $-1$, respectively. We can easily verify   $\mathcal{T}_{tr}$ satisfies Condition \ref{cond: task} in this case.

%where we denote $\bfmu_{M_1+1}:=\bfmu_1$. Hence, the numbers of tasks that map $\bfmu_j$ to $+1$ and $-1$ are both $1$ for any $j\in[M_1]$. In this example, $|\mathcal{T}_{tr}|=M_1$. In fact, we have the following lemma on the required number of training tasks to be lower bounded by exactly $M_1$. 
%\MW{ $\mathcal{T}$ is not defined. It is the set of functions. It should be defined at the beginning of section 2 when $f_i^n$ is defined. }

\iffalse 
\begin{lemma}\label{lemma: mini_task}
    With in-domain tasks defined in Definition \ref{def: task} and Condition \ref{cond: task}, the number of training tasks should satisfy $|\mathcal{T}_{tr}|\geq M_1$ to make Condition \ref{cond: task} hold. 
\end{lemma}
\ST{The statement of the lemma should be complete. Say under xxx assumptions. Assume the training data modeling is true or assume the data is generated by xx. When xxx, then xxx}
\fi

%\MW{Where is $|\mathcal{T}|$ defined? }
Following \citep{SWL21, KWLS21, LWLC23}, we assume the training labels are balanced, i.e., $\big||\{n: z^n=+1\}|-|\{n: z^n=-1\}|\big|= \mathcal{O}(\sqrt{N})$. The next theorem states the training and in-domain generalization. %\MW{Your result only applies to M1 task, no more than than? Combined with previous and the next comments, I suggest formally defining T. }

\begin{theorem}\label{thm: training} (In-Domain Generalization)  Suppose Condition \ref{cond: task} holds. For any $\epsilon>0$, when %as long as with $|\mathcal{T}_{tr}|\geq M_1$ by Lemma \ref{lemma: mini_task}, 
(i) the number of neurons in $\bfW_O$ satisfies  $m\geq \Omega(M_1^2\log M_1)$, (ii) batch size $B>\Omega(\max\{\epsilon^{-2},M_1\}\cdot\log M_1)$, (iii) the lengths of training and testing contexts are
%\MW{New notation ltr, only l was defined. use either ltr and lte or l and l' for training and testing lengths consistently. }
\vspace{-2mm}
\begin{equation}
\hspace{-2mm}l_{tr}\geq \max\{\Omega(\log M_1/\alpha),\Omega(1/(\beta^2\alpha))\},\ l_{ts}\geq {\alpha'}^{-1},\label{eqn: l_tr}
\end{equation} 
%for some large $q>1$, \MW{Is q already covered in Omega? If not, then q depends on M_1?}
(iv) and the number of iterations satisfies
%\vspace{-1mm}
\begin{equation}
\begin{aligned}
    T= \Theta(\eta^{-1}M_1\alpha^{-\frac{2}{3}}\beta^{-2/3}\sqrt{\log M_1}),\label{eqn: T}
\end{aligned}
\end{equation}
%for some $\tau\leq O(1/M_1)$
with step size $\eta\leq 1$ and $N=BT$ samples, then with a high probability, the returned model satisfies that
\vspace{-2mm}
\begin{equation}
    \underset{\bfx_{query}\sim\mathcal{D}, f\in\mathcal{T}}{\mathbb{E}}[\ell(\Psi; \bfP, z)]\leq \mathcal{O}(\epsilon).
\end{equation}
%\ST{change all $O$ to $\mathcal{O}$}
 %\MW{Shall we explain what zero generalization means, i.e., for which task?}
\end{theorem}
Theorem \ref{thm: training} characterizes the sufficient condition on the model size, the required number of iterations, and the number of prompt embedding and label pairs, %, which refers to sample complexity, 
such that the trained model achieves an in-domain generalization error of $\mathcal{O}(\epsilon)$. Theorem \ref{thm: training} includes three major insights: 

1.  \textit{In-domain generalization capability using a diminishing fraction of training tasks}. Because $\mathcal{T}_{tr}$ can satisfy Condition \ref{cond: task} even when $|\mathcal{T}_{tr}|= M_1$, then the number of training tasks is only a fraction  $(M_1-1)^{-1/2}$ of the total number of in-domain tasks in $\mathcal{T}$. %, which is the minimal number of training tasks $|\mathcal{T}_{tr}|= M_1$ out of the total number $M_1(M_1-1)$, of tasks satisfying Condition \ref{cond: task} is proven to guarantee a desired in-domain generalization.

%2. (\textit{Pattern selection}) Despite a non-trivial magnitude of $|\kappa_i^n|\leq 1$, the impact of IDI patterns on the generalization error is significantly reduced to a tiny value $\lambda_*^{-1}m^{-1}$ after training, which decreases with more $\bfW_O$ neurons. 

2. (\textit{Context length}) The required length of training and testing contexts increase in the order of $\alpha^{-1}$ and ${\alpha'}^{-1}$, respectively, which implies that a longer context is needed when the fraction of IDR patterns in the context is small. 

3. (\textit{Convergence and sample complexity}) The required number of iterations and the training samples is proportional to %$(\lambda_*^2+1)^{-2/3}$, $(1-\tau)^{-2/3}$
$\alpha^{-2/3}$. This indicates that a larger fraction of the IDR pattern in the context leads to more efficient convergence and generalization.  %\ST{what do you mean? smaller generalization error and faster convergence rate?}. 

Based on the in-domain result, we can also investigate the properties of out-of-domain generalization.

\begin{theorem}\label{thm: prompts&queries}(Out-of-Domain Generalization)

%\MW{Maybe it is better to number all conditions in Theorem 3.3, and then say suppporse conditon 3.2 holds and (xx)-(xx) all hold, xxx}
    Suppose Condition \ref{cond: task} and conditions (i)-(iv) in Theorem \ref{thm: training} hold. For any $\bfmu_1',\cdots,\bfmu_{M_1}',\bfnu_1',\bfnu_{M_2}'$ that are pairwise orthogonal %form an orthogonal set, 
    and $\|\bfmu_j'\|=\|\bfnu_k'\|=\beta$, if   
    \vspace{-2mm}
    \begin{equation}
    \bfmu_j'\in\left\{\sum_{i=1}^{M_1}k_{j,i}\bfmu_i\Big|S_j:=\sum_{i=1}^{M_1}k_{j,i}\geq 1, k_{j,i}\in\mathbb{R}\right\}, \label{eqn: mu'}
    \end{equation}
%\MW{ki needs to be the same for different uj?}\HK{No} \MW{maybe use kij}
    
    and $\bfnu_k'\in\text{span}\{\bfnu_1,\bfnu_2,\cdots, \bfnu_{M_2'}\}$, $j\in[M_1']$, $k\in[M_2']$, then with high probability, the learned model can achieve an out-of-domain generalization error of
    \vspace{-2mm}
    \begin{equation}
    \underset{\bfx_{query}\sim\mathcal{D}', f\in\mathcal{T}'}{\mathbb{E}}[\ell(\Psi; \bfP, z)]\leq \mathcal{O}(\epsilon).
\end{equation}
    %\MW{Use math euqation of generalziation like (11)}
\end{theorem}

\iffalse
\begin{corollary}\label{cor: attention}
For any testing data $\bfP^n$,
\begin{equation}
    \sum_{s\in\mathcal{N}_*^n}\text{softmax}({\bfp_s^n}^\top{\bfW_K^{(T)}}^\top\bfW_Q^{(T)}\bfp_{l+1}^n)\geq 1-\Theta(1/M)
\end{equation}
\end{corollary}
\fi

\begin{remark}
Theorem \ref{thm: prompts&queries} indicates that a one-layer Transformer can generalize well in context, even in the presence of distribution shifts between the training and testing data. The conditions for a favorable generalization encompass the following: (1) the ODR patterns are linear combinations of IDR patterns with a summation of coefficients $\geq 1$,  and each ODI pattern is in the subspace spanned by IDI patterns;  
%\MW{Isn't q and -q defined by ourself. ,Hence, we can define the label embeddings as q and -q for any binary classification problem. Why is this a condition?} \HK{I deleted it}
%(2) The label embeddings of contexts in testing and training prompts are the same, i.e., either $\bfq$ or $-\bfq$; 
(2) the testing prompt is long enough, which is linear in ${\alpha'}^{-1}$, to include context inputs involving ODR patterns. %With these conditions, Corollary \ref{cor: attention} indicates that, despite distribution shift, the attention weights of testing data also concentrate on tokens of ODR patterns as training data does in Proposition \ref{prop: self-attention}
\end{remark}

\begin{remark} (Comparison with existing ICL analysis) \citep{HCL23} analyzes the generalization performance of ICL on unseen tasks under a similar data model that includes decisive and indecisive patterns. However, \citep{HCL23} only analyzes in-domain unseen tasks, while our results also apply to one type of out-of-domain tasks through data shift.
%    (1) Under a similar data assumption characterized by some decisive/indecisive patterns, our work also depicts the out-of-domain generalization by characterizing the sufficient condition of data shift, while \citep{HCL23} only studies in-domain generalization on unseen tasks. (2) In the training dynamics analysis of ICL, 
To the best of our knowledge, only \citep{ZFB23} studies out-of-domain generalization under the setup of linear regression problems with Gaussian inputs.
They conclude that, under this setup, the covariate shift, i.e., the difference between the training and testing data distributions $\mathcal{D}$ and $\mathcal{D}'$, does not guarantee generalization. We consider classification problems under a data model different from \citep{ZFB23}. We provide the out-of-domain generalization guarantee for one type of distribution between $\mathcal{D}$ and $\mathcal{D}'$.
\end{remark}

\vspace{-2mm}
\subsection{ICL With Magnitude-Based Model Pruning}\label{subsec: prune}
%We study the generalization of ICL with the model pruned. 
\begin{theorem}\label{thm: prune}
     Let $\bfr_i$ be the $i$-row of $\bfW_O\bfW_V$, $i\in[m]$. Suppose Condition \ref{cond: task} and conditions (i)-(iv) in Theorem \ref{thm: training} hold, then there exists $\mathcal{L}\subset[m]$ with $|\mathcal{L}|=\Omega(m)$ s.t., 
     \vspace{-2mm}
    \begin{equation}
        \begin{aligned}
            \|\bfr_i^{(T)}\|\geq\Omega(1),\ &i\in\mathcal{L},\\
            \|\bfr_i^{(T)}\|\leq\mathcal(1/\sqrt{M_2}),\ &i \in\mathcal{L}^c,
        \end{aligned}\label{eqn: o_norm}
    \end{equation}
    where $\mathcal{L}^c$ is the complementary set of $\mathcal{L}$. 
    Then, for any $\epsilon>0$ and any in- or out-of-domain $\bfx_{query}\sim\mathcal{D}$ (or $\mathcal{D}'$) and corresponding $f\in\mathcal{T}$ (or $\mathcal{T}'$), pruning all neurons $i\in\mathcal{L}^c$ leads to a generalization error  
    \vspace{-2mm}
   % \mw{Pruning needs to be mathematically defined, prune wo, what changes in eq (2), notation for the resulting model. Probably as a new section 2.4.}\HK{revised}
\begin{equation}\underset{\bfx_{query}, f}{\mathbb{E}}[\ell(\Psi_{\mathcal{L}^c}; \bfP, z)]\leq \mathcal{O}(\epsilon+M_1^{-1/2}),
    \end{equation}
    where $\Psi_{\mathcal{L}^c}$ represents the model weights after removing neurons in $\mathcal{L}^c$ in $\bfW_O$.
   In contrast, pruning $\mathcal{S} \subset \mathcal{L}$ with size $|\mathcal{S}|=Rm$, where $R\in(0,1)$ and is a constant, and $\alpha'\geq \Omega(M_1^{-0.5})$ %fraction of neurons $i\in\mathcal{L}$ 
   results in 
   a %out-of-domain 
   generalization error of 
        \begin{equation}\underset{\bfx_{query}, f}{\mathbb{E}}[\ell(\Psi_{\mathcal{S}}; \bfP, z)]\geq \Omega( R+(\alpha' M_1)^{-1}).
    \end{equation}
   % \mw{Need two different notations for the pruned psi. }\HK{I use $\Psi_p$}
\end{theorem}
\begin{remark}\label{rmk: prune}
     Theorem \ref{thm: prune} proves that a constant fraction of neurons in  $\mathcal{L}$ in the trained MLP layer has large weights, while the remaining ones in  $\mathcal{L}^c$ have small weights.  Pruning neurons with a smaller magnitude leads to almost the same generalization result as that of the unpruned $\Psi$. %can generate a generalization error close to using the dense model, and (3) 
    However, pruning neurons with a larger magnitude cause an increasing generalization error as the pruning ratio $R$ increases. %Meanwhile, a larger $\alpha'$ can improve the generalization. 
Theorem \ref{thm: prune} indicates that in our setup, magnitude-based pruning on $\bfW_O$ does not hurt the model's ICL capability.  %works in out-of-domain ICL inference, and increasing $\alpha'$ by designing better prompt selection methods can mitigate the performance gap due to pruning.}
\end{remark}

%Due to different data assumptions, our out-of-domain discussion corresponds to a combination of covariate shift and task shift in \citep{ZFB23}, and our data model cannot formulate their query shift. They conclude that the covariate shift, i.e., the difference between the training and testing data distributions $\mathcal{D}$ and $\mathcal{D}'$, is not tolerable in generalization. However, our result is different in that we can characterize the sufficient conditions on the data distribution shift between $\mathcal{D}$ and $\mathcal{D}'$ to ensure an out-of-domain generalization error smaller than $\mathcal{\epsilon}$.

%\MW{Is it possible to add some high-level intuition of why the Transformer handles unseen features? }

\begin{figure*}[ht]
%\vspace*{-4mm}
\centering
\centerline{
\begin{tabular}{cccc}
\includegraphics[width=.19\textwidth,height=1.2in]{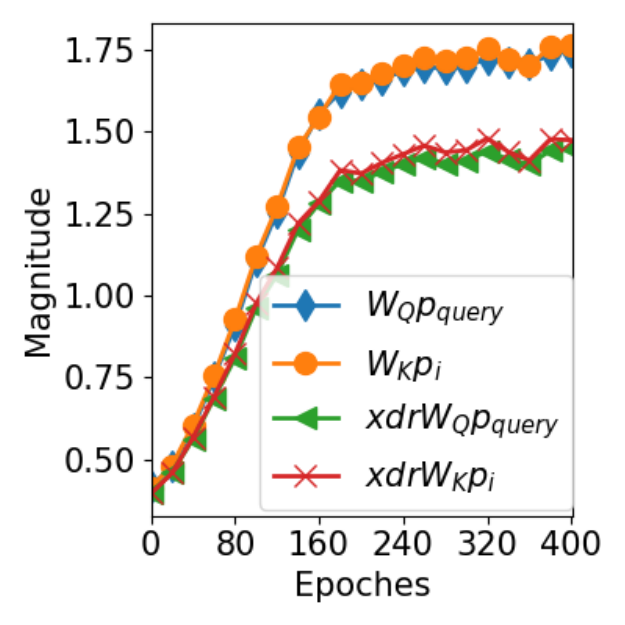}  
% &
% \hspace*{-3mm}
% \includegraphics[width=.3\linewidth,height=!]{figures/attn_dist/timm_oxfordpets_feat_dist.pdf} 
&
\hspace*{-1mm}
\includegraphics[width=.18\textwidth,height=1.2in]{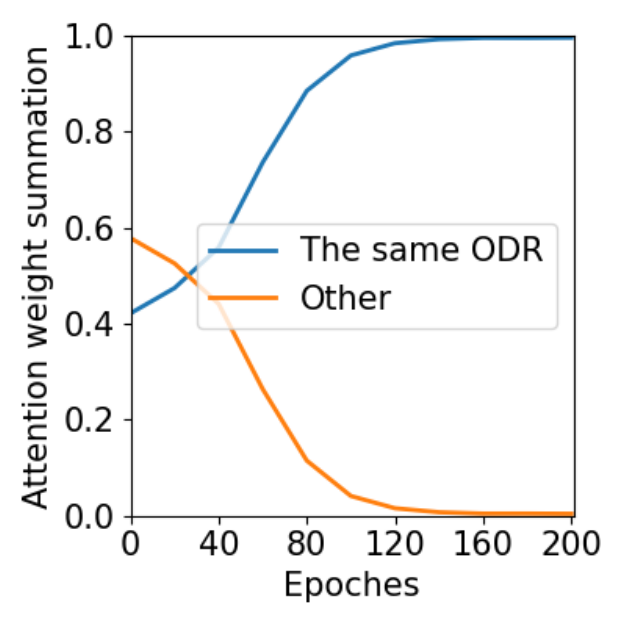}\vspace*{-1mm}
&
\hspace*{-1mm}
\includegraphics[width=.18\textwidth,height=1.2in]{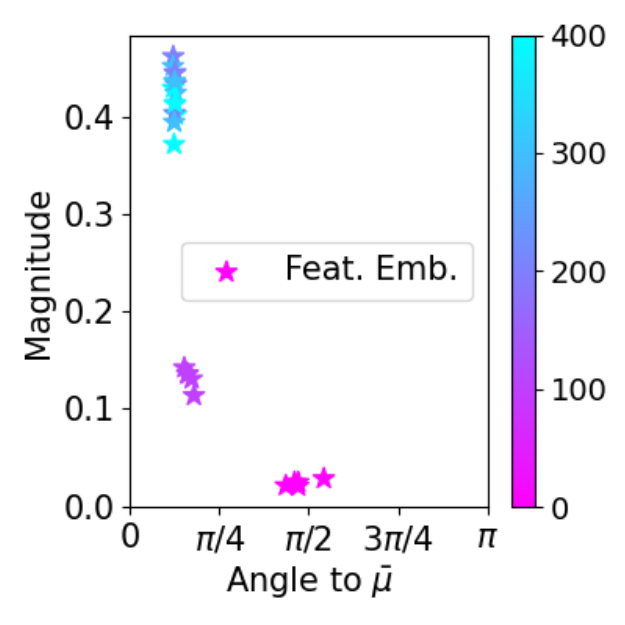}\vspace*{-1mm}
&
\hspace*{-1mm}
\includegraphics[width=.23\textwidth,height=1.2in]{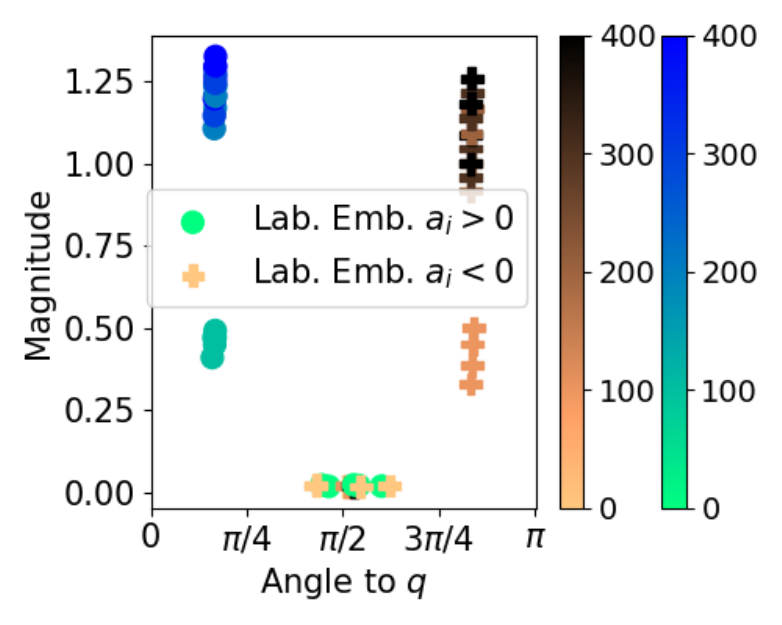}\vspace*{-1mm}
\\
(A) 
& \hspace*{-2mm} (B)
& \hspace*{-2mm} (C)
& \hspace*{-2mm} (D)
% & \hspace*{-3mm} (C)
\end{tabular}}
\vspace*{-3mm}
\caption{\footnotesize{The properties of the trained model. (A) The average norm of $\bfW_Q\bfp_{query}$, $\bfW_K\bfp_i$, $[XDR(\bfp_{query})^\top/\beta,\boldsymbol{0}^\top]\cdot\bfW_Q\bfp_{query}$, and $[XDR(\bfp_i)^\top/\beta,\boldsymbol{0}^\top]\bfW_K\bfp_i$. (B) The attention weight summation on contexts with the same ODR pattern as the query and other contexts. (C) The magnitude of the first $d_\mathcal{X}$ dimensions of $5$ neurons in $\bfW_O\bfW_V$ and their angles to $\bar{\bfmu}$ in $400$ epochs. (D) The magnitude of the rest $d_\mathcal{Y}$ dimensions of $10$ neurons in $\bfW_O\bfW_V$ and their angles to $\bfq$ in $400$ epochs.  We choose $5$ neurons for $a_i>0$ and $5$ for $a_i<0$.}} %and the angle to $\bfq$ of the label embedding of $10$ rows of $\bfW_O\bfW_V$ in $400$ epoches. }}
%\footnotesize{The properties of the trained model when $\beta=3$. (A) The average magnitude of ODR patterns mapped by $\bfW_Q^{(t)}$ and $\bfW_K^{(t)}$ increases along the training, while the magnitude of ODI patterns mapped by $\bfW_Q^{(t)}$ and $\bfW_K^{(t)}$ remains close to that at initialization. $\bfmu_j'$ and $\bfnu_k'$ are an ODR and ODI pattern, respectively. (B) The attention weight summation on context with the same ODR pattern as the query increases to $1$ during the training. (C) The angle between $\bfW_Q^{(T)}\bfp_{query}$ (or $\bfW_K^{(T)}\bfp_i$) and $(\bfb^\top,\boldsymbol{0}^\top)^\top$ is small for the out-of-domain query $\bfp_{query}$ (or out-of-domain context $\bfp_i$), where $\bfb$ is the IDR pattern of $\bfp_{query}$ (or $\bfp_i$). (D) The magnitude of feature embedding of $W_{O_{i,\cdot}}^{(T)}\bfW_V^{(T)}$ (blue ``$\times$'') and its angle to $\bar{\bfmu}$. The magnitude of feature embedding of $W_{O_{i,\cdot}}^{(T)}\bfW_V^{(T)}$ (red ``$+$'') and its angle to $+\bfq$. Feature embeddings with large magnitude have a small angle to $\bar{\bfmu}$. Label embeddings of $W_{O_{i,\cdot}}^{(T)}\bfW_V^{(T)}$ with large magnitude have a small angle with either $+\bfq$ or $-\bfq$. 

\vspace*{-4mm}
\label{fig: mechanism_4}
\end{figure*}

\vspace{-2mm}
\section{The Mechanism of ICL by the Trained Transformer}\label{sec: mechanism}
%\vspace{-2mm}
%The success of out-of-domain generalization can be understood at a high level by considering the properties of the trained model, which is elaborated more in this Section. %The trained self-attention layer can perform context selection based on ODR patterns to generate the label embedding consistent with the testing query. %Hence, the learned parameters enable similarity measurement between the out-of-domain testing query and contexts, given that IDR patterns can represent test-relevant patterns. 
%Then, the model can make accurate predictions accordingly.

Here, we provide a detailed discussion about how the generalization performance in Theorems \ref{thm: training} and \ref{thm: prompts&queries}  are achieved. 
%Based on our formulation in Section \ref{sec: problem formulation}, 
We first introduce novel properties of the self-attention layer and the MLP layer of the learned Transformer to implement ICL in Sections \ref{subsec: self-attention} and \ref{subsec: MLP}. The high-level proof idea of Theorems \ref{thm: training} and \ref{thm: prompts&queries}  %organization of the proof and disucssion are 
is  presented in Appendix \ref{subsec: proof overview}.
%\ST{you need to a figure to illustrate this mechanism}
\vspace{-2mm}
\subsection{Self-Attention Selects Contexts With the Same IDR/ODR Pattern as the Query}\label{subsec: self-attention}
%\vspace{-2mm}
%We state the following proposition about the learned self-attention layer in (\ref{eqn: transformer}).

\iffalse
\begin{proposition}\label{prop: self-attention} 
     
     Let $\Omega=[0:d_\mathcal{X}]$. With high probability, the trained model satisfies that $\bfW_Q^{(T)}[:,\Omega]\bfmu_j$ and $\bfW_K^{(T)}[:,\Omega]\bfmu_j$ are close to $\Theta(\sqrt{\log M_1})\cdot \bfmu_j$ by an error at most $\sqrt{(\log M_1)/M_1}$ for all $j\in[M_1]$. $\bfW_Q^{(T)}[:,\Omega]\bfnu_j$, $\bfW_K^{(T)}[:,\Omega]\bfnu_j$ are close to $\delta\bfnu_j$ by an error at most $\delta(\tau+1/(M_1\log M_1))$ for $j\in[M_2]$. 

\end{proposition}
\fi

We first show the learned self-attention layer promotes context examples that share the same IDR/ODR pattern as the query. Specifically, for any vector $\bfp  \in \R^{d_{\mathcal{X}}+d_{\mathcal{Y}}}$ that includes input $\bfx$ and the corresponding output  embedding $\bfy$. 
We use $\text{XDR}(\bfp)$ to represent the relevant pattern, which is the $\text{IDR}(\bfx)$ for in-domain data and $\text{ODR}(\bfx)$ for out-of-domain data. %  (or ODR) pattern in the feature embedding $\bfx$ of in-domain (or out-of-domain) $\bfp$, a column of the prompt embedding $\bfP$. 
Then
\begin{proposition}\label{prop: self-attention} %Let $\bfW_Q=[\bfW_{Q_\mathcal{X}}, \bfW_{Q_\mathcal{Y}}]$, $\bfW_K=[\bfW_{K_\mathcal{X}}, \bfW_{K_\mathcal{Y}}]$, where $\bfW_{Q_\mathcal{X}}, \bfW_{K_\mathcal{X}}\in\mathbb{R}^{m_a\times d_{\mathcal{X}}}$, $\bfW_{Q_\mathcal{Y}}, \bfW_{K_\mathcal{Y}}\in\mathbb{R}^{m_a\times d_{\mathcal{Y}}}$. 
%\ST{What is the definition of angle operator? especially for eqs. 18, 19. They are not consistent with 13 14.}
The trained model after being updated by $T$ (characterized in (\ref{eqn: T})) iterations satisfies that, for any $(\bfp, \bfW)\in\{(\bfp_{query},\bfW_Q^{(T)}), \{(\bfp_i, \bfW_K^{(T)})\}_{i=1}^l\}$, %for the query $\bfp_{query}$ and context $\bfp_i$, %i\in[l]$ defined in (\ref{eqn: data}), we have
\begin{equation}
    \|[\text{XDR}(\bfp)^\top,\boldsymbol{0}^\top]\bfW\bfp\|  \geq \Omega(\sqrt{\log M_1}),\label{eqn: W_qk xdr}
\end{equation}
\begin{equation}
    \|[\bfa^\top,\boldsymbol{0}^\top]\bfW\bfp\| \leq \mathcal{O}(\sqrt{\log M_1}(1/M_1+1/M_2)),\label{eqn: W_qk a}
\end{equation}
\begin{equation}
    \|[\bfb^\top,\boldsymbol{0}^\top]\bfW\bfp\| \leq \mathcal{O}(\sqrt{\log M_1}(1/M_1+1/M_2)),\label{eqn: W_qk b}
\end{equation}
where $\bfa$ is any IDR (or ODR) pattern that is different from $\text{XDR}(\bfp)$ for in-domain (or out-of-domain) tasks, $\bfb$ is any IDI (or ODI) pattern, and $\boldsymbol{0}$ is an all-zero vector in $\mathbb{R}^{m_a-d_\mathcal{X}}$. %\HK{I first revise it into this. I keep $p_{query}$, but I don't need to introduce $\bfp_i$ into $(XDR(x),\pm q)$}
\end{proposition}

%\MW{Remove n}
\begin{remark}
    Proposition \ref{prop: self-attention} indicates that the self-attention layer parameters $\bfW_Q^{(T)}$ and $\bfW_K^{(T)}$ in the returned model projects  $\bfp_{query}$ or context embeddings $\bfp_{i}$ mainly to the directions of the corresponding IDR pattern for in-domain data or ODR pattern for out-of-domain data. This can be deduced by combining (\ref{eqn: W_qk xdr}), (\ref{eqn: W_qk a}), and (\ref{eqn: W_qk b}), since components of $\bfW\bfp$ in other directions rather than $[XDR(\bfp)^\top,\boldsymbol{0}^\top]$ are relatively smaller. %Recall that by (\ref{eqn: x_mu_nu_tr}) and (\ref{eqn: x_mu_nu_tt}), the directions of queries and context inputs are also determined by IDI/ODI pattern. 
    Hence, Proposition \ref{prop: self-attention} implies that the learned $\bfW_Q^{(T)}$ and $\bfW_K^{(T)}$ remove  the effect of IDI/ODI patterns. % and only remains IDR/ODR patterns. 
    Meanwhile, (\ref{eqn: W_qk xdr}) states that the   $\bfW_Q^{(T)}$ and $\bfW_K^{(T)}$ enlarge the magnitude of the IDR or ODR patterns from $\Theta(1)$ to $\Theta(\sqrt{\log M_1})$, given that the $\bfW_Q^{(0)}$ and $\bfW_K^{(0)}$ are initialized with a scalar $\delta=\Theta(1)$. %Since that $\bfy_i\in\{+\bfq,-\bfq\}$, this result also implies that the inner product contribution in (\ref{eqn: IDR_enlarge}), (\ref{eqn: IDI_maintain}), (\ref{eqn: ODR_enlarge}), and (\ref{eqn: ODI_maintain}) from the first $d_{\mathcal{X}}$ columns of $\bfW_Q^{(T)}$ and $\bfW_K^{(T)}$ is much larger than that from the last $d_{\mathcal{Y}}$ columns. Therefore, the   Proposition \ref{prop: self-attention} leads to the following conclusion. 
\end{remark}
Proposition \ref{prop: self-attention} enables us to compute the attention map of the trained model. Therefore, we have the following. 

\begin{corollary}\label{cor: attention map}
 For any testing query embedding $\bfp_{query}=[\bfx_{query}^\top, 
\bf0^\top]^\top$, let $\mathcal{N}_*\in[l]$ be the set of indices of context inputs that share the same IDR (or ODR) pattern as the in-domain (or out-of-domain) $\bfx_{query}$. Then, for any constant $C>1$, by definition in (\ref{eqn: transformer}), it holds that
\vspace{-2mm}
\begin{equation}
    \sum_{s\in\mathcal{N}_*}\text{attn}(\Psi; \bfP, i)\geq 1-\Theta(1/M_1^C).\label{eqn: attn}
\end{equation}
\iffalse
\begin{equation}
    \|{\bfW_Q^{(T)}}_{1:m_a, d_{\mathcal{X}}+1:d_\mathcal{X}+d_{\mathcal{Y}}}\boldsymbol{0}\|=0,\ \|{\bfW_K^{(T)}}_{1:m_a, d_{\mathcal{X}}+1:d_\mathcal{X}+d_{\mathcal{Y}}}(\pm\bfq)\|\leq \epsilon.
\end{equation}
\fi

\end{corollary}
\vspace{-2mm}
\begin{remark}
Corollary \ref{cor: attention map} shows that after training, the attention weights become concentrated on contexts in $\mathcal{N}_*$. This means that the learned self-attention layer only selects some crucial contexts that share the same IDR/ODR pattern as the query rather than all samples uniformly or randomly.
\end{remark}

%\MW{Can we get rid of all zero vectors in Prop 1? The result may be much cleaner. On a related note, it seems the columns in all these W's that correspond to zero vectors can be arbitrary?}

\vspace{-2mm}
\subsection{MLP Neurons Distinguish Label Embeddings Rather Than Feature Embeddings.}\label{subsec: MLP}
%This part characterizes property of the trained MLP layer. %Given Corollary \ref{cor: attention map} and the fact that, with high probability, no two same training/ODI appear in a short prompt with $l_{tr}=l_{ts}=o(M_2)$ by Theorems \ref{thm: training} and \ref{thm: prompts&queries}, we can deduce that the weighted summation of $\bfp_i$ (including $\bfp_{query}$) by attention is close to $\bfmu_j$ or $\bfmu_j'$ in the feature embedding part, i.e., the first $d_\mathcal{X}$ dimensions, and close to either $\bfq$ or $-\bfq$ in the label embedding space, i.e., the last $d_{\mathcal{Y}}$ dimensions. 
%which is the training/ODR pattern of $\bfx_{query}$. Similarly, the weighted summation of $\bfy_i$ by attention is close to the embedding of the label $y$ of $\bfx_{query}$. Let $\bfW_V=(\bfW_{V_\mathcal{X}},\bfW_{V_\mathcal{Y}})$, where $\bfW_{V_\mathcal{X}}\in\mathbb{R}^{m_b\times d_\mathcal{X}}$, $\bfW_{V_\mathcal{Y}}\in\mathbb{R}^{m_b\times d_\mathcal{Y}}$. Then, one can expect that the trained $\bfW_{O_{i,\cdot}}\bfW_{V_\mathcal{X}}, i\in[m]$ shall be close to the average of all training/ODR patterns $\bfmu_j$ or $\bfmu_j'$, since that the data with different $\bfmu_j$ or $\bfmu_j'$ is uniformly distributed among all the batches. Likewise, the trained $\bfW_{O_{i,\cdot}}\bfW_{V_\mathcal{X}}$ shall be close to either $+\bfq$ or $-\bfq$, which depends on the index $i\in[m]$.
%Then, we have the following proposition for the trainable parameters $\bfW_O$ and $\bfW_V$. 
We next show that the trained MLP layer can distinguish the label embeddings for data from different classes. 

\begin{proposition}\label{prop: mlp}
%Let $\bfW_V=(\bfW_{V_\mathcal{X}},\bfW_{V_\mathcal{Y}})$, where $\bfW_{V_\mathcal{X}}\in\mathbb{R}^{m_b\times d_\mathcal{X}}$, $\bfW_{V_\mathcal{Y}}\in\mathbb{R}^{m_b\times d_\mathcal{Y}}$. 
%\ST{What is $\bfW_{O_{(i,\cdot)}}$?}

%Let $\bfW_{O_{(i,\cdot)}}$ be the $i$-th row of $\bfW_O$. 
%Let $\bfr_i=\bfW_{O_{(i,\cdot)}}\bfW_V:=(\bfr_{i_{d_\mathcal{X}}},\bfr_{i_{d_\mathcal{Y}}})$, 
Let $\bfr_i$ introduced in Theorem \ref{thm: prune} be $ (\bfr_{i_{d_\mathcal{X}}},\bfr_{i_{d_\mathcal{Y}}})$ %\HK{rigorously speaking, there should be zero here }
%\mw{First, why is that zero? If WoWv indeed have all zero columns for last few columns, the model is reduncant} \HK{my bad, it is not zero. i want to say there is some extra entries. may not be zero}
%denote the $i$-th row of $\bfW_O\bfW_V$, 
where $\bfr_{i_{d_\mathcal{X}}}\in\mathbb{R}^{1\times d_\mathcal{X}}$,  $\bfr_{i_{d_\mathcal{Y}}}\in\mathbb{R}^{1\times d_\mathcal{Y}}$.%, and $\bfr_i'\in\mathbb{R}^{m_b-d_\mathcal{X}-d_\mathcal{Y}}$. There exists a set $\mathcal{L} \subset [m]$ with size $|\mathcal{L}|=\Omega(m)$ %For a constant fraction of $i\in[m]$ and given $\lambda_*\geq 1$, we have
Then, for any $i \in \mathcal{L}$,
\vspace{-2mm}
\begin{equation}
    \bfr_{i_{d_\mathcal{X}}}^{(T)}\bar{\bfmu}/(\|\bfr_{i_{d_\mathcal{X}}}^{(T)}\|\cdot \|\bar{\bfmu}\|)\geq 1-\Theta(1)/M_2,\label{eqn: o_angle}
\end{equation}
\vspace{-2mm}
\begin{equation}
    \bfr_{i_{d_\mathcal{Y}}}^{(T)} \bfq_e/(\|\bfr_{i_{d_\mathcal{Y}}}^{(T)} \|\cdot \|\bfq_e\|)\geq 1-\Theta(1)/M_1,\label{eqn: o_angle 2}
\end{equation}
where $\bar{\bfmu}=\sum_{k=1}^{M_1}\bfmu_k^\top/M_1$, $\bfq_e=\bfq$ if $a_i>0$ and $\bfq_e=-\bfq$ if $a_i<0$, where $a_i$ is the $i$-th entry of $\bfa$ in (\ref{eqn: data}). 
\iffalse
    For a constant fraction of $i\in[m]$, we have \begin{multline}
{\bfW_{O}^{(T)}}_{i,d_\mathcal{X}+1:d_\mathcal{X}+d_\mathcal{Y}}{\mathbf{sa}(\Psi^{(T)},\bfP^n)}_{d_\mathcal{X}+1:d_\mathcal{X}+d_\mathcal{Y}}
\\
>{\bfW_{O}^{(T)}}_{i,1:d_\mathcal{X}}{\mathbf{sa}(\Psi^{(T)},\bfP^n)}_{1:d_\mathcal{X}}.
        \end{multline}
    For other $i$, $\|{\bfW_O^{(T)}}_{i,1:m_b}\mathbf{sa}(\Psi^{(T)},\bfP^n)\|\leq \mathcal{O}(\xi)$.
\fi
\end{proposition}
%￥\MW{what is ai}\HK{the i-th entry in a in eqn 2}
%\mw{Remove SC as it only appears here. Option 1: expand the definition, shall be able to fit in one line. Option 2, other formats of results like (14)-(16). Please also check the revised Prop 4.5 statement.}
\begin{remark}
Proposition \ref{prop: mlp} demonstrates that neurons with indices in $\mathcal{L}$ have the following two properties. (P1) The first $d_\mathcal{X}$ entries  of all the corresponding row vectors in $\bfW_O^{(T)} \bfW_V^{(T)}$  approximate the average of all IDR patterns $\bfmu_j$, $j\in[M_1]$. (P2) The  next  $d_\mathcal{Y}$ entries of the $i$th row of $\bfW_O^{(T)} \bfW_V^{(T)}$ approximates the label embedding $\bfq$ when $a_i>0$ and approximates $-\bfq$ when $a_i<0$. (P1) indicates that the output layer focuses on all IDR patterns equally rather than any IDI pattern. (P2) indicates that the MLP layer can distinguish label embeddings for different classes.

%Rthat except for some MLP neurons that lead to a small magnitude on $\|\bfW_{O_{(i,\cdot)}}\bfW_V^{(T)}\|$, the remaining MLP neurons $\bfW_{O_{(i,\cdot)}}^{(T)}$ with $\bfW_V^{(T)}$ are close to a direction, where (1) the feature embedding, i.e., the first $d_\mathcal{X}$ dimensions, is the average or sum of all IDR patterns $\bfmu_j$, $j\in[M_1]$; %(This is because the data with different $\bfmu_j$ is uniformly distributed among all the batches during the training);
%(2) the label embedding, i.e., the last $d_\mathcal{Y}$ dimensions, is $\bfq$ for positive neurons with $a_i>0$ or $-\bfq$ for negative neurons with $a_i<0$. From (1), we know that the projections of the feature embedding onto different $\bfmu_j, j\in[M_1]$ are similar, while the projections of the label embedding onto $\bfq$ and $-\bfq$ are different. This means the MLP neurons mainly distinguish different label embeddings for binary classification instead of feature embeddings for different IDR patterns. 
\end{remark}

\vspace{-2mm}
\section{Numerical Experiments}

\textbf{Data Generation} We verify our theoretical findings %using data generated by Section \ref{sec: problem formulation} 
using data generated as described in Section \ref{sec: problem formulation}. Let $d_\mathcal{X}=d_\mathcal{Y}=30$, $\beta=3$, $K'=5$, $K=0.5$. The in-context binary classification error is evaluated by $\mathbb{E}_{(\bfx,y)}[\Pr(y\cdot F(\Psi; \bfP)<0)]$ for $\bfx$ following either $\mathcal{D}$ or $\mathcal{D}'$ and $\bfP$ constructed in (\ref{eqn: data}). If not otherwise specified, we set $M_1=6$, $M_2=24$. For out-of-domain generalization, $M_1'=3$, $\bfnu_i'=\bfnu_i$ for $i\in[M_2']$. $\bfmu_1'=0.3\cdot(\bfmu_1-\bfmu_2)+a\bfmu_5+b\bfmu_6$. %$\bfmu_2'=0.3\cdot(\bfmu_3-\bfmu_4)+a\bfmu_7+b\bfmu_8$. 
$\bfmu_2'=\sqrt{2}/{2}\cdot(\bfmu_1+\bfmu_2)$. $\bfmu_3'=\sqrt{2}/{2}\cdot(\bfmu_3+\bfmu_4)$. %\footnote{Note that this setting allows the same IDR patterns used in different $\bfmu_j'$, $j\in[M_1']$}. 
For testing, we select contexts with the two decisive patterns with $\alpha'/2$ probability each and others with $(1-\alpha')/(M_1'-2)$ probability each to keep the context outputs balanced. %For $\bfmu_1'$ defined in (\ref{eqn: mu'}) with $S_1$ as the summation of coefficients of $\bfmu_i$, we set $k_{j1}=0.8$, $k_{ji}=0$ for $i=2,3,4$.

\textbf{Model and Training Setup}: The models we use include both the one-layer Transformer defined in (\ref{eqn: transformer}) and the 3-layer 2-head real-world model GPT-2 \citep{RWCL19} following \citep{BCWX23, WZCB23}.  If not otherwise specified, we set $\alpha=0.8$, $l_{tr}=20$ for training.  The training tasks are formulated as follows to satisfy Condition \ref{cond: task}. %\mw{bfa is firstly defined here or introduced before?}\HK{only for here} 
Define $\bfa_i=\bfa_{i+M_1}=\bfmu_i$ for $i\in [M_1]$, and then the $((k-1)\cdot M_1+j)$-th task function maps the queries with $\bfa_j$ and $\bfa_{j+k}$ as IDR patterns to $+1$ and $-1$, respectively, for $j\in[M_1]$ and $k\in[U]$. For the one-layer Transformer, we use $U=1$ and $m_a=m_b=60$. Hence, $|\mathcal{T}_{tr}|=6$, and there are $|\mathcal{T}\backslash\mathcal{T}_{tr}|=24$ in-domain unseen tasks. For GPT-2, $U=4$. Then, $|\mathcal{T}_{tr}|=24$, $|\mathcal{T}\backslash\mathcal{T}_{tr}|=6$. Note that we evaluate in-domain generalization error only on unseen tasks $\mathcal{T}\backslash\mathcal{T}_{tr}$, which is generally an upper bound of that defined in (\ref{eqn: id_generalization}) after sufficient training. 

%\ST{some details can be put in the appendix.}
\vspace{-2mm}
\subsection{Experiments on the Generalization of ICL}
\begin{figure}[!h]
\vspace*{-4mm}
\centering
\centerline{
\begin{tabular}{cc}
\includegraphics[width=.2\textwidth,height=1.3in]{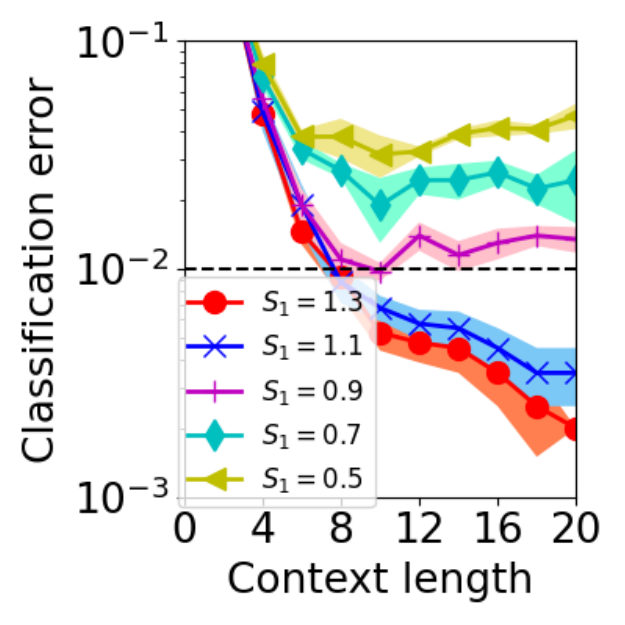}  
% &
% \hspace*{-3mm}
% \includegraphics[width=.3\linewidth,height=!]{figures/attn_dist/timm_oxfordpets_feat_dist.pdf} 
&
\hspace*{-1mm}
\includegraphics[width=.2\textwidth,height=1.3in]{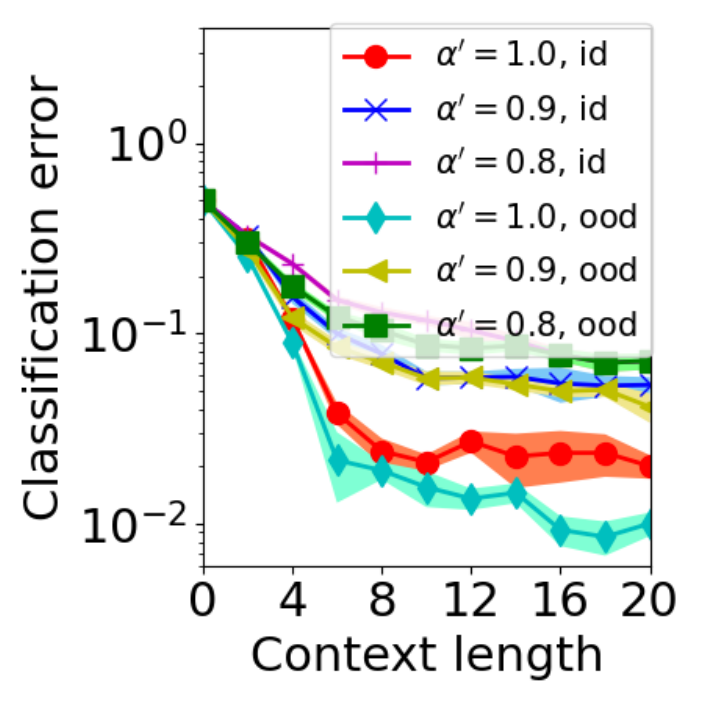}\vspace*{-1mm}
\\
(A) 
& \hspace*{-2mm} (B)
% & \hspace*{-3mm} (C)
\end{tabular}}
\vspace*{-3mm}
\caption{\footnotesize{Out-of-domain ICL classification error on GPT-2 with (a) different $S_1$ on GPT-2 (b) different $\alpha'$ for in-domain (id) and out-of-domain (ood) generalization. }}\label{fig: OOD-gpt2}
\vspace{-3mm}
\end{figure}

%We set $\alpha=0.8$, $l_{tr}=20$. For the training on GPT-2, we formulate the training tasks as follows: Let $\bfa_i=\bfa_{i+M_1}=\bfmu_i$ for $i\leq M_1$, and then the $(k-1)\cdot M_1+j$-th task function map the queries with $\bfmu_j$ and $\bfmu_{j+k}$ as IDR patterns to $+1$ and $-1$, respectively, for $j\in[M_1]$ and $k\in[4]$. Hence, $|\mathcal{T}_{tr}|=24$. For in-domain generalization, there are $|\mathcal{T}|-|\mathcal{T}_{tr}|=6$ unseen tasks. 
We first verify the sufficient condition (\ref{eqn: mu'}) for out-of-domain generalization. From the selection of $\bfmu'$'s, we know that  $S_1=a+b$, $S_2=S_3=\sqrt{2}$. %by (\ref{eqn: mu'}). 
We vary $a$ and $b$ while satisfying $a^2+b^2+2\cdot0.3^2=1$. 
%The result is averaged over all cases that satisfy (i) $a+b=S$, (ii) $a^2+b^2+2\cdot0.3^2=1$. 
Figure \ref{fig: OOD-gpt2} (A) shows that the out-of-domain classification error archives $<0.01$ when $S_1\geq 1$ and deviates from $0$ when $S_1<1$, which justifies the necessity of condition (\ref{eqn: mu'}). We then investigate how the context length is affected by $\alpha'$, i.e., the fraction of contexts with the same IDR/ODR pattern as the query. Figure \ref{fig: OOD-gpt2} (B) indicates that a longer testing context length is needed when $\alpha'$   is smaller for in- or out-of-domain, which is consistent with the lower bound of $l_{ts}$ in (\ref{eqn: l_tr}) and Theorem \ref{thm: prompts&queries}. 

\begin{figure}[!h]
%\vspace*{-4mm}
\centering
\centerline{
\begin{tabular}{cc}
\includegraphics[width=.2\textwidth,height=1.3in]{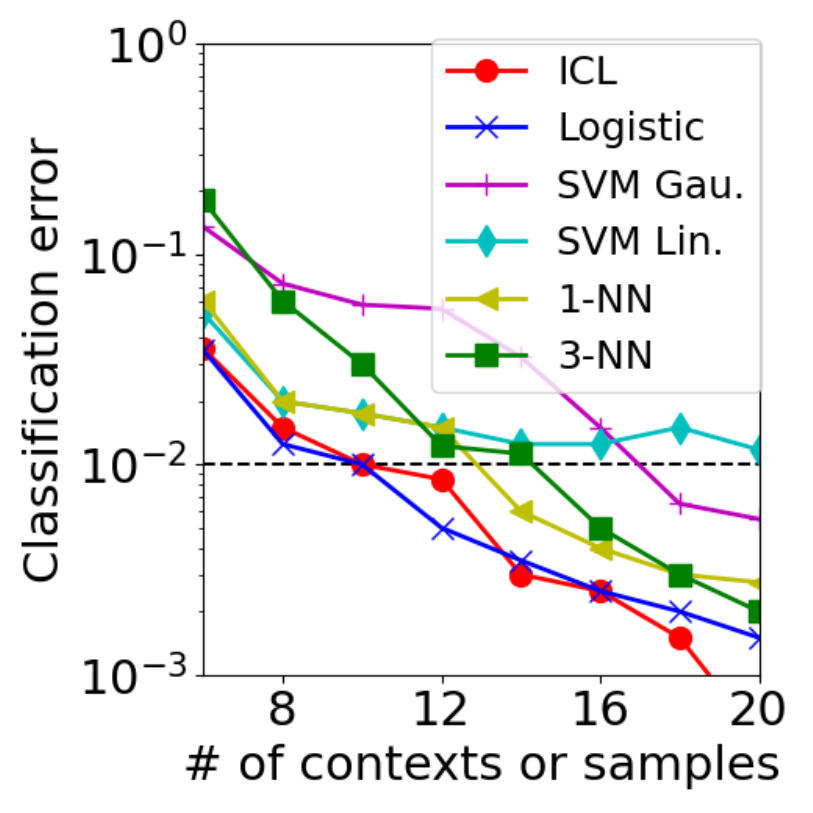}  
% &
% \hspace*{-3mm}
% \includegraphics[width=.3\linewidth,height=!]{figures/attn_dist/timm_oxfordpets_feat_dist.pdf} 
&
\hspace*{-1mm}
\includegraphics[width=.2\textwidth,height=1.3in]{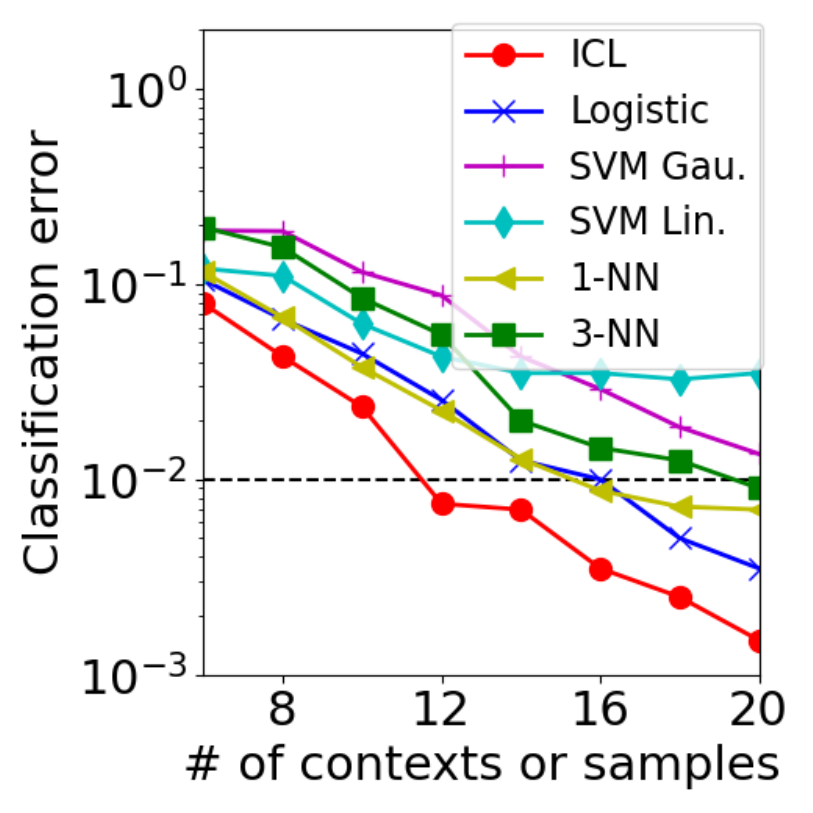}\vspace*{-1mm}
\\
(A) 
& \hspace*{-2mm} (B)
% & \hspace*{-3mm} (C)
\end{tabular}}
\vspace*{-3mm}
\caption{\footnotesize{Binary classification performance of using ICL, logistic regression (Logistic), SVM with Gaussian kernel (SVM Gau.), SVM with linear kernel (SVM Lin.), 1-nearest neighbor (1-NN), and 3-nearest neighbor (3-NN) with one-layer Transformer when (A) $\alpha'=0.8$ (B) $\alpha'=0.6$.}}\label{fig: one-layer-compare}
\vspace{-3mm}
\end{figure}

 We then compare ICL with other machine learning algorithms for classification, where contexts are used as training samples for these methods. Figure \ref{fig: one-layer-compare} (A) and (B) show that when $\alpha'=0.8$, the advance of ICL   over other algorithms is not significant, while when $\alpha'=0.6$, ICL is the most sample-efficient for a small generalization error. %This means that, if properly pre-trained,
 Thus, ICL can remove irrelevant data and is more robust to random noise in labels than other learning algorithms.

\begin{figure}[!h]
%\vspace*{-4mm}
\centering
\centerline{
\begin{tabular}{cc}
\includegraphics[width=.2\textwidth,height=1.3in]{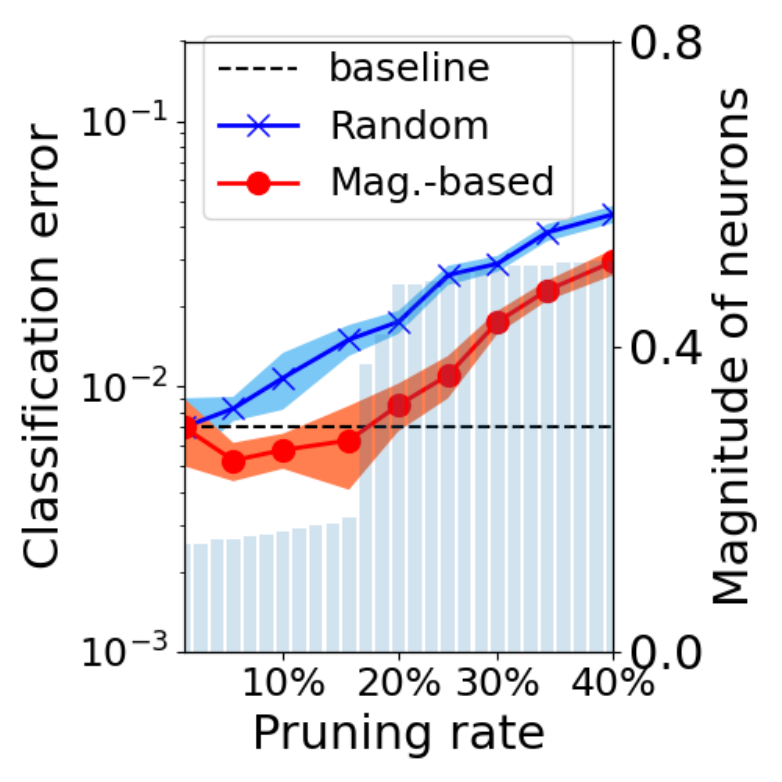}  
% &
% \hspace*{-3mm}
% \includegraphics[width=.3\linewidth,height=!]{figures/attn_dist/timm_oxfordpets_feat_dist.pdf} 
&
\hspace*{-1mm}
\includegraphics[width=.2\textwidth,height=1.3in]{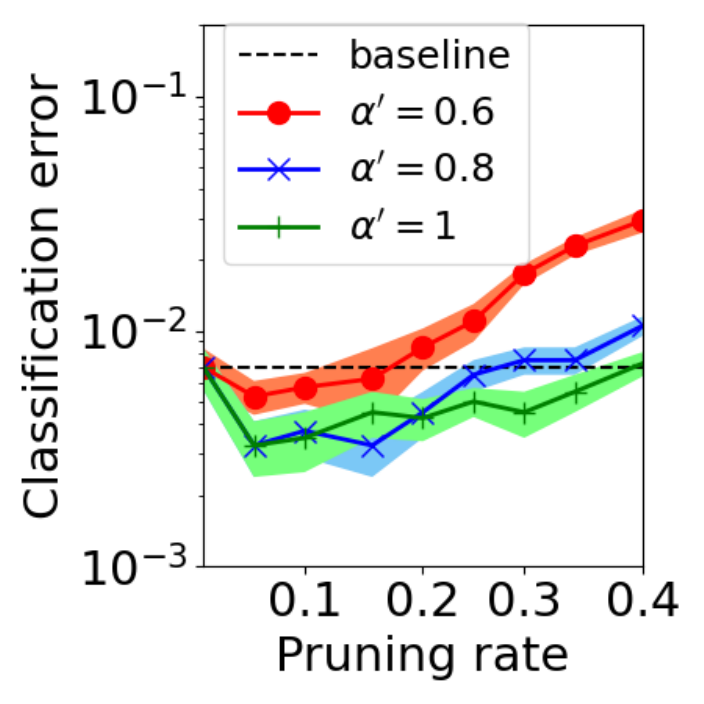}\vspace*{-1mm}
\\
(A) 
& \hspace*{-2mm} (B)
% & \hspace*{-3mm} (C)
\end{tabular}}
\vspace*{-3mm}
\caption{\footnotesize{(A) Out-of-domain classification error (left y-axis for curves) with model pruning of the trained $\bfW_O$ using baseline (no pruning), random pruning, and magnitude-based pruning (Mag.-based), and the magnitude of each neuron of $\bfW_O$ (right y-axis for light blue bars) (B) Out-of-domain classification error when varying $\alpha'$. These two are implemented on a one-layer Transformer.} }\label{fig: pruning}
\end{figure}

 We also investigate the effect of pruning techniques on ICL. Let $\alpha=0.6$. Figure \ref{fig: pruning} (A) shows that magnitude-based pruning does not hurt out-of-domain generalization if the pruning rate is lower than around $15\%$, which is the ratio of $\bfW_O$ neurons with a small magnitude. The generalization error increases as the pruning rate increases when pruning neurons with large weights. This is consistent with Theorem \ref{thm: prune} and Remark \ref{rmk: prune}. Figure \ref{fig: pruning} (B) justifies the impact of $\alpha'$  in Theorem \ref{thm: prune} that larger $\alpha'$ can improve the performance of the pruned model.

%\ST{Can we have the results on T5 or chatgpt 2?}

\vspace{-2mm}
\subsection{Experiments on the Mechanism of ICL}
%\PYB{I think we should bring Fig. 2 to this page}
We examine our findings regarding the mechanism of ICL in Section \ref{sec: mechanism} using a one-layer Transformer formulated in (\ref{eqn: transformer}). %We set $m_a=m_b=60$, $\alpha=0.6$. The training tasks are formulated via a similar scheme as for GPT-2, but with $k\in[1]$. Hence, $|\mathcal{T}_{tr}|=6$.
In Figure \ref{fig: mechanism_4} (A) and (B), we consider out-of-domain data with $a=b=0.64$. Figure \ref{fig: mechanism_4} (A) shows that for any query $\bfp_{query}$ (or context example $\bfp_i$ for $i\in l_{ts}$), the norm of $[XDR(\bfp)^\top, \boldsymbol{0}^\top]\bfW_Q\bfp_{query}$ (or $[XDR(\bfp)^\top, \boldsymbol{0}^\top]\bfW_K\bfp_i$) is close to the norm of $\bfW_Q\bfp_{query}$ (or $\bfW_K\bfp_i$). %while the small difference between the two pairs of curves
This implies that the components of $\bfW_Q\bfp_{query}$ (or $\bfW_K\bfp_i$) in directions other than $[XDR(\bfp)^\top, \boldsymbol{0}^\top]$ are small, which is consistent with (\ref{eqn: W_qk a}) and (\ref{eqn: W_qk b}) in Proposition \ref{prop: self-attention}.  Moreover, these norms increase from initialization during  training, which % The growing magnitude 
justifies (\ref{eqn: W_qk xdr}).  
Figure \ref{fig: mechanism_4} (B) depicts the concentration of attention on contexts in $\mathcal{N}_*$ after training. This verifies Corollary \ref{cor: attention map}. % of concentrated attention  
Figure \ref{fig: mechanism_4} (C) and (D) jointly verify Proposition \ref{prop: mlp}. The color bars represent the epochs of training. We can observe that except for some neurons,  $\bfr_{i_{d_\mathcal{X}}}$ grows to be close to the direction of $\bar{\bfmu}$ with a larger magnitude in Figure \ref{fig: mechanism_4} (C). Moreover,  Figure \ref{fig: mechanism_4} (D) shows for $a_i>0$ (or $a_i<0$), $\bfr_{i_{d_\mathcal{Y}}}$ becomes close to $\bfq$ (or $-\bfq$) with a large magnitude.

\vspace{-2mm}
\section{Conclusion} %\ST{there is no need to have this section if there is no enough space. numerical section is much more important than this section.}
This paper provides theoretical analyses of the training dynamics of Transformers with nonlinear attention and nonlinear MLP, and the resulting ICL capability for new tasks with possible data shift. This paper also provides a theoretical justification for magnitude-based pruning to reduce inference costs while maintaining the ICL capability. Future directions include designing practical prompt selection algorithms and model pruning methods based on the obtained insights, as well as investigating ICL on generation tasks.

%\HK{This work studies the training and in/out-of-domain generalization of ICL using a single-head, one-layer nonlinear Transformer for classification. Focusing on data and task formulation based on in/out-of-domain-relevant patterns, we explicitly quantify the required number of samples and iterations of supervised pre-training for in-domain generalization and out-of-domain generalization if out-of-domain patterns satisfy a certain relationship with in-domain patterns. We theoretically show a mechanism of trained Transformers implementing ICL and prove how magnitude-based pruning on Transformers preserves ICL.}

%\HK{This paper considers a simplified but representative nonlinear Transformer architecture to theoretically study ICL on classification tasks. Future directions include (1) designing practical prompt selection algorithms and model pruning methods based on the insights, (2) investigating ICL on generation tasks, etc.}


\section*{Acknowledgements}
This work was supported by IBM through the IBM-Rensselaer Future of Computing Research Collaboration. We thank Dr. Hui Wan (\textit{hui.wan@gmail.com}) at Google Cloud AI, USA, for the valuable discussion.
We thank all anonymous reviewers for their constructive comments.

\section*{Impact Statement}
This paper aims to explore the mechanisms of transformer-based neural networks in the context of in-context learning. The primary focus is on the mathematical analysis of generalization errors. To the best of our knowledge, no potential societal consequences are associated with our work. %Thus, there is no need to specifically highlight any societal impacts in this regard.


%\clearpage
%a\nocite{langley00}
%\newpage
\normalem
%\bibliography{ref_Hongkang.bib, ref.bib}
%\bibliographystyle{icml2024}

%%%%%%%%%%%%%%%%%%%%%%%%%%%%%%%%%%%%%%%%%%%%%%%%%%%%%%%%%%%%%%%%%%%%%%%%%%%%%%%
%%%%%%%%%%%%%%%%%%%%%%%%%%%%%%%%%%%%%%%%%%%%%%%%%%%%%%%%%%%%%%%%%%%%%%%%%%%%%%%
% APPENDIX
%%%%%%%%%%%%%%%%%%%%%%%%%%%%%%%%%%%%%%%%%%%%%%%%%%%%%%%%%%%%%%%%%%%%%%%%%%%%%%%
%%%%%%%%%%%%%%%%%%%%%%%%%%%%%%%%%%%%%%%%%%%%%%%%%%%%%%%%%%%%%%%%%%%%%%%%%%%%%%%
\appendix
\newpage
\onecolumn

\section{Proof Sketch}
We partially include the proof backbone in Section \ref{sec: mechanism} when introducing the mechanism of the trained Transformer. We elaborate more about our proof intuition of Theorem \ref{thm: training} in the following.

We first briefly introduce our proof intuition of Theorem \ref{thm: training}. We first respectively build Lemmas \ref{lemma: QK}, \ref{lemma: V}, and \ref{lemma: O} to characterize gradient updates for $\bfW_Q$ and $\bfW_K$, $\bfW_V$ and $\bfW_O$. These Lemmas are based on an observation that a constant fraction of neurons in $\bfW_O$ can always be activated (Lemma \ref{lemma: initial_WU}, \ref{lemma: update_WU}) to avoid the non-smoothness of Relu activation. The orthogonality of patterns and Definition \ref{def: task} enable the self-attention layer to learn in-domain-relevant (IDR) patterns rather than in-domain-irrelevant (IDI) patterns and select contexts with the same IDR pattern as the query. Then, to develop Theorem \ref{thm: training}, we use Lemma \ref{lemma: QK} to show that the attention weights converge to be close to $1$ when $\eta T \geq \Omega(M_1\sqrt{\log M_1})$. Next, we compute the network output according to the label embedding using Lemma \ref{lemma: V} and \ref{lemma: O}. Finally, we derive the required number of iterations to make the generalization error $\mathcal{O}(\epsilon)$ by concentration inequalities.

We then would like to specify how we handle the Relu activation in the gradient of $\bfW_K$, $\bfW_Q$. We also want to clarify that how we can handle the training dynamics related to the softmax is different from \citep{HCL23} although we both derive a sparse attention distribution. Inspired by the intuition of feature-learning analyses for two-layer Relu networks \citep{BG21, SWL21, ZWCL23, AL23}, we initialize the model such that at least a constant fraction of the neurons of $\bfW_O$ are activated (Lemma \ref{lemma: initial_WU}), which are called lucky neurons as in \citep{ZWCL23, LWLC23}. We prove that these lucky neurons are always activated (Lemma \ref{lemma: update_WU}) and grow with an increasing magnitude and two fixed directions of the label embedding along the training (Lemma \ref{lemma: O}). Then, we can show that the gradient growths of $\bfW_Q$ and $\bfW_K$ can be lower bounded by contributions from these lucky neurons. Therefore, we are able to characterize the gradient updates of $\bfW_K$ and $\bfW_Q$ given a dynamic $\bfW_O$. This process is different from \citep{HCL23} since \citep{HCL23} does not include Relu MLP, so there is no need to study lucky neurons. Besides, we use Hinge loss, while their training loss is logistic loss, which leads to more training phases, as a difference in the training dynamics between us.

\section{Addition Discussions and Extensions}\label{sec: additional discussion}
\subsection{The Motivation to Study NONLINEAR Transformers}
The reasons we study nonlinear Transformers in this work are as follows. First, nonlinear Transformers for ICL, which are different from linear Transformers, are common in practice but less explored in theory. Nonlinear attention and nonlinear MLP are default components of standard Transformers \citep{VSPU17} and are widely applied in large language models for implementing ICL in practice. Existing works show that nonlinear Transformers exhibit their empirical advantages when learning nonlinear functions \citep{CCS23} or conducting dynamic programming tasks \citep{YAHF24}. However, state-of-the-art theoretical works \citep{ZFB23, HCL23, WZCB23} ignore the nonlinearities (partially) to simplify the analysis or the presentation. Second, the analysis of nonlinear Transformers is quite different from that of Transformers without nonlinearities. For example, softmax attention has a different derivative from linear attention, which includes nonlinear exponential terms and needs a more complicated computation of the gradient updates. Relu MLP provides several non-differential points, which makes the loss landscape more challenging to analyze.

\subsection{The Discussion on Single/Multi-Head Attention}
There are several reasons why we only study single-head attention in the main body of the paper. First, all the previous theoretical works studying the optimization and generalization of Transformers on ICL \citep{ZFB23, HCL23, WZCB23} only consider single-head attention in the network. Some concurrent works consider multi-head attention, but they either do not study ICL \citep{DGTT23, CL24} or do not involve convergence/generalization guarantee \citep{CRHT24}. Hence, the question of how the ICL ability on unseen tasks and out-of-domain data is obtained by training is still unexplored. Our theoretical analysis studies the convergence and generalization of ICL using Transformers with softmax attention and Relu MLP, involving generalization on unseen tasks and OOD data. Second, our empirical experiments on GPT-2 in Figure \ref{fig: OOD-gpt2} are conducted with two heads to verify our theoretical findings, which means some theoretical conclusions hold in Transformers with multiple heads.

However, our analysis for single-head attention can be extended to multi-head attention to some degree. Consider a multi-head attention layer where the layer output is a concatenation of the output of each head, i.e., 
\begin{equation}
{\bigg\Vert}_{h=1}^H\sum_{i=1}^l \bfW_{V_h}\bfp_i\cdot \text{softmax}({\bfp_i}^\top \bfW_{K_h}^\top \bfW_{Q_h} \bfp_{query}).
\end{equation}
The overall conclusion will remain the same, given the same data formulation and initialization on each head because of the orthogonality of patterns. Specifically, we can still show that each attention head selects contexts with the same in-domain-relevant (IDR) pattern as the query. The MLP layer will still make predictions based on the label embedding, as suggested in Section \ref{subsec: MLP}. We will leave the analysis on multi-head attention with more general settings as future directions.

\subsection{Extension to Multiple Patterns for One Class}
We can extend our analysis to the case that several orthogonal IDR patterns correspond to the label $+1$, while some other orthogonal IDR patterns correspond to the label $-1$. Then, as long as there is always a context input that shares the same IDR/ODR pattern as the query, we can still prove that the self-attention layer selects contexts with the same IDR/ODR pattern as the query. Furthermore, we can show the MLP layer makes predictions based on the label embedding. Therefore, the mechanism remains the same as the current setting in the manuscript, where one pattern corresponds to one pattern. We will leave other cases where the data formulation is different in future works.

The reason why we use our current setting in the main body of the paper is to simplify the presentation while emphasizing our major contributions of analyzing optimization and generalization of nonlinear Transformers both in-domain and out-of-domain. As the first work on this problem, as far as we know, we believe our data formulation keeps the necessary complexity.

\subsection{Additional Related Works}
We introduce other existing theoretical works on learning and generalization of neural networks in this section. Some works \citep{ZSJB17, FCL20, LZW22, ZLWL23, LZZW24} study the generalization performance following the model recovery framework by probing the local convexity around a ground truth parameter. %Another line of works, which is usually for overparameterized neural networks, is 
The neural-tangent-kernel (NTK) analysis \citep{JGH18, ALL19, ALS19,  CG19, ZG19,  CCGZ20, LWLC22, SLW24} considers strongly overparameterized networks  %NTK can
to linearize the neural network around the initialization. The generalization performance is independent of the feature distribution. %, which may oversimplify the network architecture and 
\cite{DM20, SWL21, KWLS21, BG21, ZWCL23, LWLC23, ZLYC24, CZWL23, CWMW24} investigate the generalization of neural networks assuming  %that %each data can be divided into 
a data model consisting of discriminative patterns and background patterns. Our analysis belongs to the last line of research.

\section{Additional Experiments and the Algorithm}\label{sec: more experiment}
We first present the training algorithm introduced in Section \ref{subsec: SGD}.
\begin{algorithm}[ht]
\begin{algorithmic}[1]
\caption{Training with Stochastic Gradient Descent (SGD)}\label{alg: sgd}
\STATE{\textbf{Hyperparameters:}} 
The step size $\eta$, the number of iterations $T$, batch size $B$.
\STATE{\textbf{Initialization:}} Each entry of $\bfW_O^{(0)}$ and $\bfa^{(0)}$ from $\mathcal{N}(0,\xi^2)$ and $\text{Uniform}(\{+1/\sqrt{m},-1/\sqrt{m}\})$, respectively. 
$\bfW_Q$, $\bfW_K$ and $\bfW_V$ are initialized such that all diagonal entries of $\bfW_V^{(0)}$, and the first $d_\mathcal{X}$ diagonal entries of $\bfW_Q^{(0)}$ and $\bfW_K^{(0)}$ are set as $\delta$ with $\delta\in(0,0.2]$.

\iffalse
\STATE{\textbf{Stage-1 training by SGD:}} Let $\bfU^{(0)}=\bfW_O^{(0)}$. For $t=0,1,\cdots,T_0-1$, 
\begin{equation}
\begin{aligned}
\bfU^{(t+1)}&=\bfU^{(t)}-\eta \cdot \frac{1}{B} \sum_{n\in\mathcal{B}_t} \nabla_{\bfU^{(t)}}\ell(\bfx_n,y_n;\bfa^{(0)}, \bfU^{(t)},\bfW_V^{(0)},\bfW_K^{(0)}, \bfW_Q^{(0)},\bfb^{(0)}),
\end{aligned}
\end{equation}
\begin{equation}
    \bfW_O^{(0)}=\bfU^{(T_0-1)}.
\end{equation}
\fi
\STATE{\textbf{Training by SGD:}} For each iteration, we independently sample $\bfx_{query}\sim\mathcal{D}$, $f\in\mathcal{T}_{tr}$ to form a batch of training prompt and labels $\{\bfP^n, z^n\}_{n\in\mathcal{B}_t}$ as introduced in Section \ref{sec: data task}. Each IDR pattern is sampled equally likely in each batch. For each $t=0,1,\cdots,T-1$ and $\bfW^{(t)}\in\Psi^{(t)}$
\vspace{-0.1in}
%\textcolor{red}{$$\bfW_{t+1}=\bfW_t-\eta_0\widetilde{\nabla f_n(\bfW_t)}=\bfW_t-\eta_0\Big(\nabla f_n(\bfW^*)+\frac{1}{n}\sum_{i=1}^n \nu_i\Big)$$}
\begin{equation}\label{eqn:gradient}
\begin{aligned}
\bfW^{(t+1)}&=\bfW^{(t)}-\eta \cdot \frac{1}{B} \sum_{n\in\mathcal{B}_t} \nabla_{\bfW^{(t)}}\ell(\Psi^{(t)}; \bfP^n, z^n).
\end{aligned}
\end{equation}
\vspace{-0.1in}
\STATE{\textbf{Output: }} $\bfW_O^{(T)}$, $\bfW_V^{(T)}$, $\bfW_K^{(T)}$,  $\bfW_Q^{(T)}$.
\end{algorithmic}
\end{algorithm}

Then, we introduce additional experiments to verify our theory. 

\subsection{The impact of $\alpha$}
\begin{figure}[htb]
\centering
\vspace*{-0mm}
\centerline{
\begin{tabular}{cc}
\hspace*{0mm}\includegraphics[width=.26\textwidth,height=1.6in]{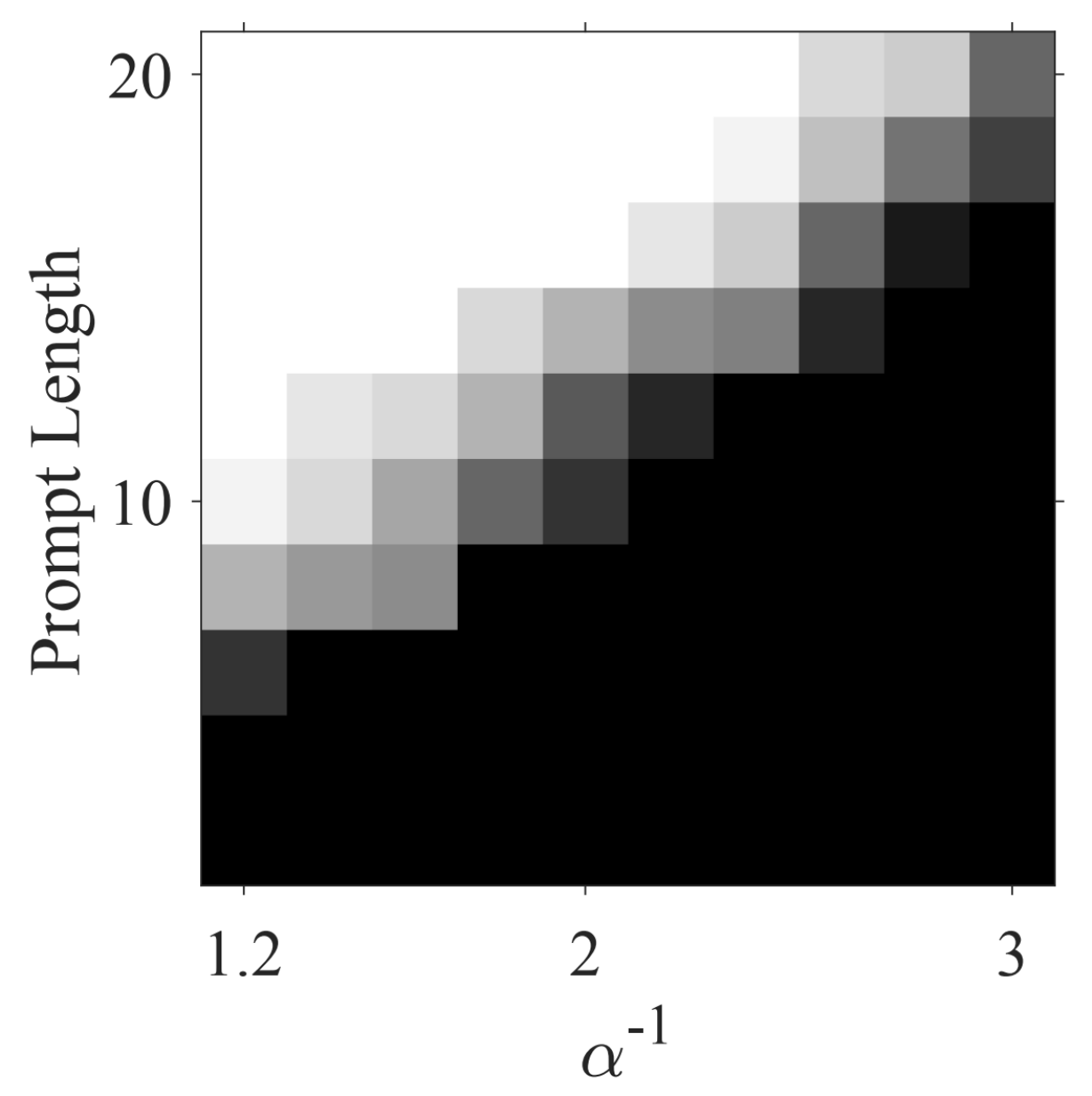}  
&\hspace*{+2mm}\includegraphics[width=.20\textwidth,height=1.6in]{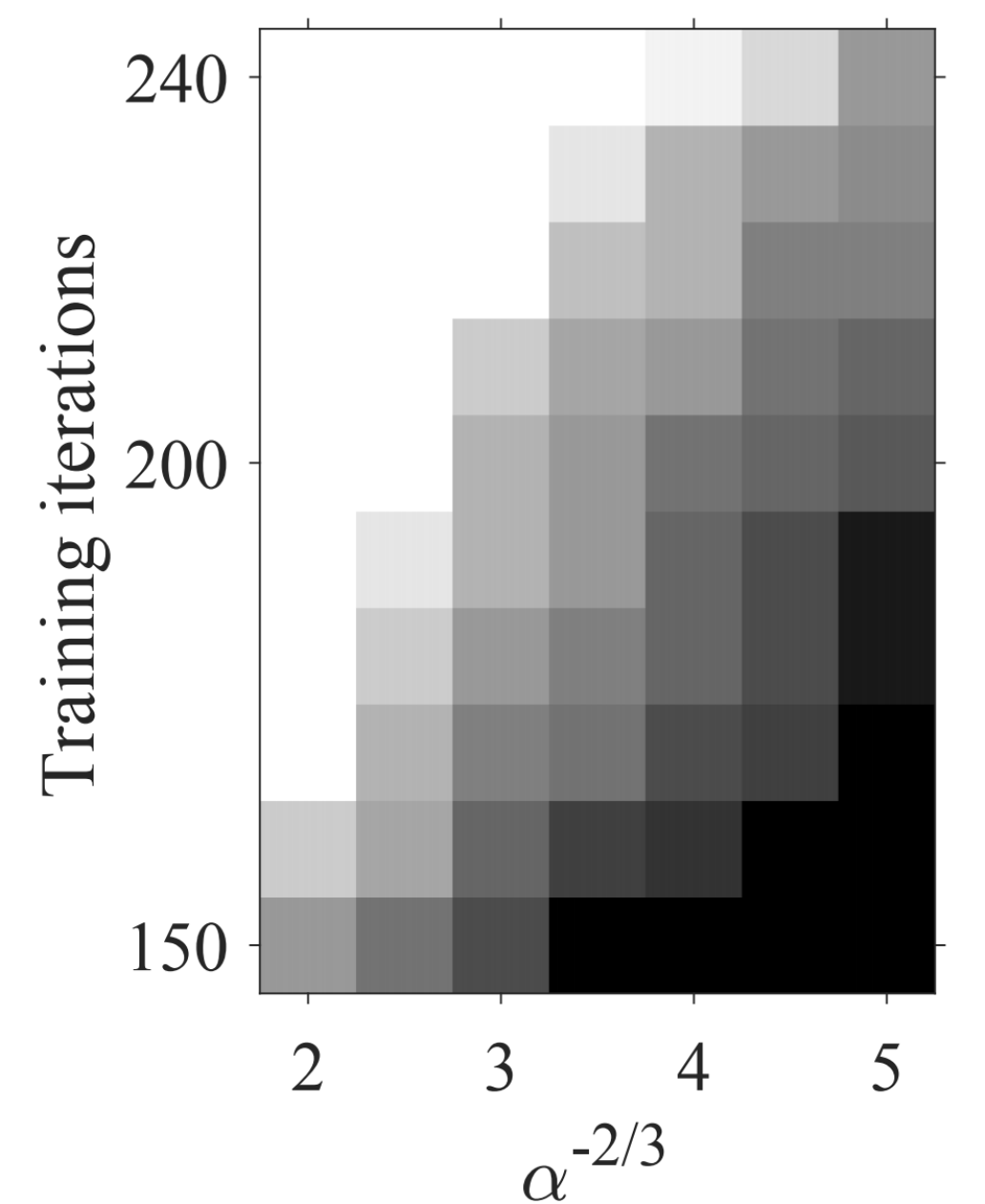}
\end{tabular}}
\vspace*{-3mm}
\caption{The prompt length against $\alpha$, and the required number of training iterations against $\alpha$.}
\vspace*{-2mm}
\label{figure: alpha}
\end{figure}

We choose $\alpha=0.6$ and use a one-layer Transformer as in (\ref{eqn: transformer}). Figure \ref{figure: alpha} shows that the required length of the training prompt is linear in $\alpha^{-1}$, while the required number of training iterations is linear in $\alpha^{-2/3}$, which verify the theoretical findings in (\ref{eqn: l_tr}) and (\ref{eqn: T}).

\subsection{The required number of training tasks}
\begin{figure}[htb]
\centering
\vspace*{-0mm}

\includegraphics[width=.26\textwidth,height=1.6in]{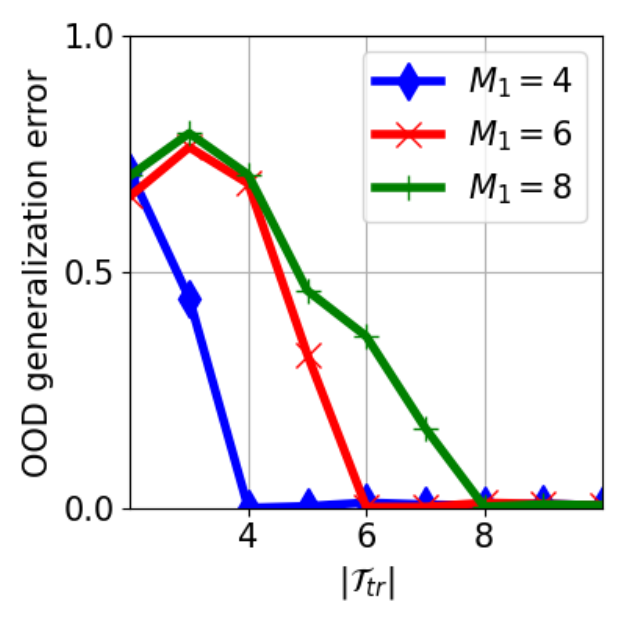}  

\vspace*{-3mm}
\caption{The required number of training tasks for different $M_1$.}
\vspace*{-2mm}
\label{figure: task}
\end{figure}

We choose $\alpha=0.6$ and use a one-layer Transformer as in (\ref{eqn: transformer}). 
For a given $\mathcal{T}$, we first generate a set of tasks that satisfies Condition \ref{cond: task} as follows. Define $\bfa_i=\bfa_{i+M_1}=\bfmu_i$ for $i\in [M_1]$, and then the $j$-th task function map the queries with $\bfa_j$ and $\bfa_{j+1}$ as IDR patterns to $+1$ and $-1$, respectively, for $j\in[M_1]$. Then, we get a task set $\mathcal{T}_{tr0}$ with $|\mathcal{T}_{tr0}|=M_1$. Then, we vary the number of training tasks in the way that (1) we sample within $\mathcal{T}_{tr0}$ to get a set $\mathcal{T}_{tr}$ with $|\mathcal{T}_{tr}|\leq M_1$ (2) we sample within $\mathcal{T}\backslash\mathcal{T}_{tr0}$ to get a set $\mathcal{T}_{tr}'$, and $\mathcal{T}_{tr}=\mathcal{T}_{tr}'\cup\mathcal{T}_{tr0}$ such that $|\mathcal{T}_{tr}|\geq M_1$. Figure \ref{figure: task} shows that for any $M_1$, the generalization error is significant as long as $|\mathcal{T}|_{tr}<M_1$, while the generalization error reaches around $0$ as long as $|\mathcal{T}|_{tr}\geq M_1$ and $\mathcal{T}$ covers all the possibilities of IDR patterns and labels. This verifies that Condition \ref{cond: task} can be met with a fraction of $(M_1-1)^{-1/2}$ of total number of in-domain tasks.

\section{Proofs of the Main Theorems}
We first provide several useful definitions and key lemmas for the proof of the main theorems. Table \ref{tbl:notations} shows a summary of notations used in the proof.

\begin{table}[h!]

  \begin{center}
        \caption{Summary of Notations}
        \label{tbl:notations}
\begin{tabularx}{\textwidth}{lX} % <-- Alignments: 1st column left, 2nd middle and 3rd right, with vertical lines in between

\toprule
Notations & Annotation \\
\hline
\small $\bfx_s^n$, $\bfy_s^n$  & $\bfx_s^n$ is the data for classification. $\bfy_s^n$ is the embedding of the label for $\bfx_s^n$.  \\
\hline
\small $\bfP^n$, $z^n$ & $\bfP^n$ is a prompt that consists of $l$ pairs of $\bfx_s^n$ and $\bfy_s^n$, $s\in[l]$. The last column of $\bfP^n$ contains $\bfp_{query}^n$, which is the query of $\bfP^n$. $z^n\in\{+1,-1\}$ is the binary label of $\bfp_{query}^n$, which is also the label of $\bfP^n$ when we formulate the problem as a supervised learning problem. \\
 \hline
 \small $F(\Psi; \bfP^n)$, $\ell(\Psi; \bfP^n, z^n)$ &  $F(\Psi; \bfP^n)$ is the Transformer output for $\bfP^n$ with $\Psi$ as the parameter. $\ell(\Psi; \bfP^n, z^n)$ is the loss function value given $\bfP^n$ and the corresponding label $z^n$.\\ 
 \hline
 \small $\bfp_s^n$, $\bfmu_j$, $\bfnu_k$ & $\bfp_s^n$ is the $s$-th example with the corresponding label in $\bfP^n$. If $s=query$, $\bfp_s^n$ is the query. $\bfmu_j$ and $\bfnu_k$ are the IDR and IDI patterns in the feature embedding of $\bfp_s^n$ as the corresponding coefficients, respectively.  \\
 \hline
 \small $\bfq$ & $\bfq$ is the label space embedding.\\
 \hline
 \small $M_1$, $M_2$, $M$ & $M_1$ is the number of IDR patterns. $M_2$ is the number of IDI patterns. $M=M_1+M_2$.  \\
 \hline
 \small   $\alpha$, $a$ & $\alpha$ is the probability of selecting examples that contain either of the two decisive IDR patterns in each $\bfP^n$. $a=1/|a_i|$ where $a_i$ is the entry of each neuron in $\bfW_O$. $a=m$. \\
 \hline
 \small $\kappa$, $\kappa'$, $K$, $K'$, $\beta$ & $\kappa$ and $\kappa’$ are the coefficients of the IDI pattern and the ODI pattern in the input $\bfx$, respectively. $\kappa$ and $\kappa’$ follow uniform distribution $\textit{U}(-K,K)$ and $\textit{U}(-K’,K’)$ with $K\leq 1/2$ and $K’\leq \mathcal{O}(1)$, respectively. $\beta$ is the norm of in-/out-of-domain-(ir)relevant (IDR/ODR/IDI/ODI) patterns.\\
 \hline
  \small $\mathcal{W}_n$, $\mathcal{U}_n$ &  
  The sets of lucky neurons. $\mathcal{W}_n$ is the set of neurons of $\bfW_O$ that can activate the terms inside $\text{Relu}(\cdot)$ in $F(\Psi; \bfP^n)$ for $z^n=+1$ at initialization. $\mathcal{U}_n$ is the set of neurons of $\bfW_O$ that can activate the Relu part of $F(\Psi; \bfP^n)$ for $z^n=-1$ at initialization.  \\
  \hline
  \small  $\mathcal{W}$, $\mathcal{U}$& $\mathcal{W}=\cup_{n\in[N]}\mathcal{W}_n$. $\mathcal{U}=\cup_{n\in[N]}\mathcal{U}_n$\\
  \hline 
  \small $\mathcal{N}_j^n$ & The set of examples in $\bfP^n$ that contains $\bfmu_j$ as the IDR pattern.\\
 \hline
  \small $\gamma_t$  &  $\gamma_t$ is the summation of attention weight on examples that have different IDR patterns from the query.  \\
  \hline
  \small  $\zeta_t$ &    $\zeta_t$ is smallest positive value inside the $\text{Relu}(\cdot)$ in $F(\Psi; \bfP^n)$ for all the $\bfW_O$ neuron and all $n\in[N]$. \\
 \hline
 \small $\mathcal{B}_b$  & $\mathcal{B}_b$ is the SGD batch at the $b$-th iteration. \\
 \hline
 \small  $l_{tr}$  &   $l_{tr}$ is the prompt length of the training data.   \\
  \hline
 \small    $l_{ts}$ &     $l_{ts}$ is the prompt length of the testing data. \\
 \hline
 \small $\mathcal{O}()$, $\Omega()$, $\Theta()$ & We follow the convention that $f(x)=O(g(x))$ (or $\Omega(g(x))$, $\Theta(g(x)))$) means that $f(x)$ increases at most, at least, or in the order of $g(x)$, respectively.\\
 \hline 
 \small $\gtrsim$, $\lesssim$ & $f(x)\gtrsim g(x)$ (or $f(x)\lesssim g(x)$ ) means that $f(x)\geq \Omega(g(x))$ (or $f(x)\lesssim \mathcal{O}(g(x))$).\\
\bottomrule
\end{tabularx}
\end{center}

\end{table}

\subsection{Proof Overview of Main Theorems}\label{subsec: proof overview}

This section illustrates how Corollary \ref{cor: attention map} and Proposition \ref{prop: mlp} contribute to the final in- and out-of-domain generalization performance of ICL. %\mw{no need of  Prop 4.1?} %prediction and the comparison with existing works in terms of the mechanism of ICL.
%decodes the obtained feature by the self-attention layer with a high weight on the embedding of the label part. 

\textbf{The establishment of generalization}

1. (\textit{Self-Attention}) % Given Corollary \ref{cor: attention map} and %the fact that, with high probability, no two same training/ODI appear in a short prompt with $l_{tr}=l_{ts}=o(M_2)$ by Theorems \ref{thm: training} and \ref{thm: prompts&queries}
%appropriate prompt length in (\ref{eqn: l_tr})%\footnote{We mean the case where the prompt length satisfies \ref{eqn: l_tr} and is no longer than $O(M_2)$. }
%,
We can deduce from Corollary \ref{cor: attention map} that, for a query with IDR pattern $\bfmu_j$ ($j\in [M_1]$) and label $+1$,  the weighted summation of contexts and the query %for $l=l_{tr}$ or $l=l_{ts}$ 
by the attention score, i.e., $\sum_{i=1}^l \bfp_i\text{attn}(\Psi; \bfP, i)$, is close to $[\bfmu_j^\top, \bfq^\top]^\top$.  This is because as long as the training/testing prompt length satisfies (\ref{eqn: l_tr}), large attention weights are assigned on $\bfp_i$ of which the IDR pattern is $\bfmu_j$, and the label embedding is $\bfq$ by (\ref{eqn: attn}). Similarly, if its label is $-1$, the weighted summation of contexts and the query outputs $[\bfmu_j^\top, -\bfq^\top]^\top$. %if the label is $-1$). %in the feature embedding part, the weighted summation of $\bfp_i$ (including $\bfp_{query}$) by attention is close to the IDR pattern of the query, and in the label embedding space, such weighted summation is close to $\bfq$ if the label is $+1$ (or $-\bfq$ if the label is $-1$).
%\mw{I do not see this. From Prof. 4.1, it seems the self attention only outputs the feature embeding. Why can it learn label embedding correctly?}
%\mw{Still do not see how Cor 4.3 itself can lead to the result of [mu q]. It only talks about the attention weights.}\HK{I add one more sentence there.}

2. (\textit{MLP}) By Proposition \ref{prop: mlp}, we know that a large enough proportion of positive (or negative) neurons $i\in[m]$ have the label embedding of $\bfW_{O_{(i,\cdot)}}^{(T)}\bfW_V^{(T)}$ close to $\pm\bfq$ (\ref{eqn: o_angle 2}). They can thus map the weighted summation of contexts and the query by attention with $+\bfq$ (or $-\bfq$) to positive (or negative) values. This leads to a correct prediction in-domain (Theorem \ref{thm: training}).

3. (\textit{Out-of-Domain Generalization})  Since  Corollary \ref{cor: attention map} also applies to ODR patterns, then for a query with an ODR pattern $\bfmu_j'$, $j\in[M_1']$, the resulting weighted summation of contexts and the query is close to $[{\bfmu_j'}^\top,\bfq^\top]^\top$ or $[{\bfmu_j'}^\top,-\bfq^\top]^\top$. %generates the feature embedding of the attention-weighed summation of the query and contexts close to the query's ODR pattern. 
Then, by combining (\ref{eqn: o_angle}), (\ref{eqn: o_angle 2}) and the condition on ODR pattern characterized in (\ref{eqn: mu'}), we can ensure that the MLP layer produces a desired prediction out of the domain (Theorem \ref{thm: prompts&queries}).% \mw{what is this? ${\bfmu_j'}^\top\bar{\bfmu}\geq \bfmu_u^\top\bar{\bfmu}$ for any $u\in[M_1]$}. \HK{a typo there. $\bar{\mu}$ is the feature embedding of WoWv. Here I want to say that the contribution in the feature embedding of the MLP layer of OOD case is not smaller than that of ID case, to give a correct prediction. Maybe too detailed here. Can I just say "xxxx, we can ensure an MLP layer output consistent with the in-domain case"? }Then, a desired prediction can be obtained (Theorem \ref{thm: prompts&queries}).

%\mw{Why does "we can ensure an MLP layer output consistent with the in-domain case" mean correct ood prediction? OOD is different from ID task, then the same output as ID a correct prediction? I actually do not see why OOD analysis needs to connect with ID here in explaining the high-level idea.}\HK{changing to "we can ensure that the MLP layer produces a desired prediction out of the domain (Theorem \ref{thm: prompts&queries})" and remove the last sentence of this paragraph?}\mw{yes}
%\textbf{Major proof idea}
%We study the gradient updates of trainable parameters $\bfW_Q$, $\bfW_K$, $\bfW_V$, and $\bfW_O$ by induction. Specifically, we build up Lemma \ref{lemma: QK} to show $\bfW_Q$ and $\bfW_K$ can only enlarge the magnitude of IDR patterns gradually along the training, which leads to Proposition \ref{prop: self-attention} and Corollary \ref{cor: attention map}. Then, we characterize the gradient of $\bfW_V$ in terms of $\bfW_O$ in Lemma \ref{lemma: V}, and the projection of iterates of $\bfW_O$ in different directions in Lemma \ref{lemma: O}, from which Proposition \ref{prop: mlp} can be obtained. Lemmas \ref{lemma: initial_WU} and \ref{lemma: update_WU} are supportive lemmas to ensure the sign of the gradients.

%\ST{please a table or list to summarize your notations.}
\subsection{Preliminaries}
\begin{lemma} \label{lemma: chernoff}
    (Multiplicative Chernoff bounds, Theorem D.4 of \citep{MRT18}) Let $X_1$, $\cdots$, $\bfX_m$ be independent random variables drawn according to some distribution $\mathcal{D}$ with mean $p$ and support included in $[0,1]$. Then, for any $\gamma\in[0,\frac{1}{p}-1]$,  the following inequality holds for $\hat{p}=\frac{1}{m}\sum_{i=1}^m X_i$:
    \begin{equation}
        \Pr(\hat{p}\geq (1+\gamma)p)\leq e^{-\frac{m p\gamma^2}{3}},
    \end{equation}
    \begin{equation}
        \Pr(\hat{p}\leq (1-\gamma)p)\leq e^{-\frac{m p\gamma^2}{2}}.
    \end{equation}
\end{lemma}

%\ST{$\Pr()$ and $\mathbb{P}\{\}$ should be consistent.}

\begin{definition}\label{def: sub-Gaussian}\citep{V10}
We say $X$ is a sub-Gaussian random variable with sub-Gaussian norm $K>0$, if $(\mathbb{E}|X|^p)^{\frac{1}{p}}\leq K\sqrt{p}$ for all $p\geq 1$. In addition, the sub-Gaussian norm of X, denoted $\|X\|_{\psi_2}$, is defined as $\|X\|_{\psi_2}=\sup_{p\geq 1}p^{-\frac{1}{2}}(\mathbb{E}|X|^p)^{\frac{1}{p}}$.
\end{definition}

\begin{lemma}  (\cite{V10} 
Proposition 5.1,  Hoeffding's inequality)  Let $X_1, X_2, \cdots, X_N$ be independent centered sub-gaussian random variables, and let $K=\max_i\|\bfX_i\|_{\psi_2}$. Then for every $\bfa=(a_1,\cdots,a_N)\in\mathbb{R}^N$ and every $t\geq0$, we have
\begin{equation}
    \Pr\Big(\Big|\sum_{i=1}^N a_i X_i\Big|\geq t\Big)\leq e\cdot \exp\left(-\frac{ct^2}{K^2\|\bfa\|^2}\right),\label{hoeffding}
\end{equation}
where $c>0$ is an absolute constant.
\end{lemma}

\begin{definition}\label{def: zeta}
    For any data index $n$ and iteration $t$, we can find $i$ such that $\bfW_{O_{(i,\cdot)}}^{(t)}\sum_{s=1}^{l+1} (\bfW_V^{(t)}\bfp_s^n)\text{softmax}({\bfp_s^n}^\top{\bfW_K^{(t)}}^\top\bfW_Q^{(t)}\bfp_{query}^n)>0$ by the initialization with high probability. Define 
    \begin{enumerate}
        \item $\zeta_{i,n,t}:=\bfW_{O_{(i,\cdot)}}^{(t)}\sum_{s=1}^{l+1} (\bfW_V^{(t)}\bfp_s^n)\text{softmax}({\bfp_s^n}^\top{\bfW_K^{(t)}}^\top\bfW_Q^{(t)}\bfp_{query}^n)$.
        \item $\zeta_{i,t}=\min_{n}\{\zeta_{i,n,t}\}$.
        \item $\zeta_{t}=\min_{i}\{\zeta_{i,t}\}$.
        \item 
    $\gamma_{t,n}=1-\sum_{s\in\mathcal{N}_*^n}\text{softmax}((\bfW_K^{(t)}\bfp_s^n )^\top(\bfW_Q^{(t)}\bfp_{query}^n ))$.
    \item $\gamma_t=\max_{n\in[N]}\{\gamma_{t,n}\}$.
    \end{enumerate}  %\ST{This is not the way of defining something. just define xxx as xxx. these equalities are confusing.}
\end{definition}

\iffalse
\begin{lemma}\label{lemma: pre}
    When $t\geq \Omega(1)$, we have that for $i\in\cup_{l=1}^{M_1}\mathcal{W}_l(t)\cup\mathcal{U}_l(t)$,
\begin{equation}
    \left\|\eta\frac{1}{B}\sum_{b=0}^{t-1}\sum_{n\in\mathcal{B}_b}\frac{\partial \ell(\tilde{\bfP}^n,z^n;\Psi)}{\partial \bfW_{O_{(i,\cdot)}}}[d_\mathcal{X}+1:d_\mathcal{X}+d_\mathcal{Y}]\right\|= \Theta(\delta/a),
\end{equation}
while
\begin{equation}
    \Big\|\eta\frac{1}{B}\sum_{b=0}^{t-1}\sum_{n\in\mathcal{B}_b}\frac{\partial \ell(\tilde{\bfP}^n,z^n;\Psi)}{\partial \bfW_{O_{(i,\cdot)}}}[1: d_\mathcal{X}]\Big\|=\Theta(\delta\beta/a).
\end{equation}
For $i\notin\cup_{l=1}^{M_1}\mathcal{W}_l(t)\cup\mathcal{U}_l(t)$, we can obtain
\begin{equation}
\begin{aligned}
    \Big\|\eta\frac{1}{B}\sum_{b=0}^{t-1}\sum_{n\in\mathcal{B}_b}\frac{\partial \ell(\tilde{\bfP}^n,z^n;\Psi)}{\partial \bfW_{O_{(i,\cdot)}}}[d_\mathcal{X}+d_\mathcal{Y}: m_b]\Big\|
    \lesssim \eta \sqrt{\frac{1}{B}}\cdot\frac{1}{a}.
\end{aligned}
\end{equation}
\end{lemma}
\fi

%\ST{please use begin proof and end proof for the deviation process}

\begin{lemma}\label{lemma: QK} (gradient updates of $\bfW_Q$ and $\bfW_K$) By the SGD training method described in Section \ref{subsec: SGD}, we have the following equations. Given the definition of in-/out-of-domain data as in (\ref{eqn: data}) and the in-/out-of-domain data distribution $\mathcal{D}$ in (\ref{eqn: x_mu_nu_tr}) and $\mathcal{D}'$ in (\ref{eqn: x_mu_nu_tt}), we study the gradient updates in the directions of queries or contexts. Note that we require $m\gtrsim M_1^2$, $B\gtrsim M_1\log M_1$, $l=l_{tr}\gtrsim 1$, $\beta\in [1,O(1)]$. \\
\noindent We first consider the case when the feature embeddings of the query $\bfx_{query}$ and the example $\bfx_q, q\in[l]$ are $\bfmu_j$. The label embedding is $\boldsymbol{0}$ for the query and $\pm\bfq$ for non-query examples. Then, for any $l,a\in[M_1]$, $k\in[M_2]$,  $t_0\geq 1$, where $\bfmu_l$ forms a task in $\mathcal{T}_{tr}$ with $\bfmu_j$ and $\bfmu_a$ does not,  %Then, for any $j,l,a\in[M_1]$, $k,b\in[M_2]$,  $t_0\geq 1$, where $\bfmu_l$ forms a task in $\mathcal{T}_{tr}$ with $\bfmu_j$ and $\bfmu_a$ does not, $\bfnu_k$ is added to $\bfmu_l$ or $\bfmu_j$ in the training data, and $\bfnu_b$ is a vector that is orthogonal to all IDR patterns and all possible $\nu_k$, 
%\ST{Don't understand. You need to provide the proof of this lemma. Start from the definition of $\beta$}
    \begin{equation}
    \begin{aligned}
        (\bfmu_j^\top, \boldsymbol{0}^\top)\eta\sum_{b=0}^{t_0}\frac{1}{B}\sum_{n\in\mathcal{B}_b}\frac{\partial \ell(\tilde{\bfP}^n,z^n;\Psi)}{\partial \bfW_Q^{(t)}}\Big|_{t=t_0}(\bfx_{query}^\top, \boldsymbol{{0}}^\top)^\top
        \gtrsim  \eta\frac{1}{M_1}\sum_{b=0}^{t_0}\zeta_b\delta \gamma_b \beta^4,\label{lm_eq: Q mu_j mu_j}
    \end{aligned}
\end{equation}
\begin{equation}
    \begin{aligned}
        &\Big|(\bfmu_l^\top, \boldsymbol{0}^\top)\eta\frac{1}{B}\sum_{b=0}^{t_0-1}\sum_{n\in\mathcal{B}_b}\frac{\partial \ell(\tilde{\bfP}^n,z^n;\Psi)}{\partial \bfW_Q^{(t)}}\Big|_{t=t_0}(\bfx_{query}^\top, \boldsymbol{{0}}^\top)^\top\Big|\\
        \lesssim    &e^{-\Theta((\frac{\eta t_0}{M_1})^2)}\Big|(\bfmu_j^\top, \boldsymbol{0}^\top)\eta\sum_{b=0}^{t_0-1}\frac{1}{B}\sum_{n\in\mathcal{B}_b}\frac{\partial \ell(\tilde{\bfP}^n,z^n;\Psi)}{\partial \bfW_Q^{(t)}}\Big|_{t=t_0}(\bfx_{query}^\top, \boldsymbol{{0}}^\top)^\top\Big|,\label{lm_eq: Q mu_j mu_l}
    \end{aligned}
\end{equation}
\begin{equation}
    \begin{aligned}
        &\Big|(\bfmu_a^\top, \boldsymbol{0}^\top)\eta\frac{1}{B}\sum_{b=0}^{t_0-1}\sum_{n\in\mathcal{B}_b}\frac{\partial \ell(\tilde{\bfP}^n,z^n;\Psi)}{\partial \bfW_Q^{(t)}}\Big|_{t=t_0}(\bfx_{query}^\top, \boldsymbol{{0}}^\top)^\top\Big|\\
        \lesssim    &\frac{1}{M_1}\Big|(\bfmu_j^\top, \boldsymbol{0}^\top)\eta\sum_{b=0}^{t_0-1}\frac{1}{B}\sum_{n\in\mathcal{B}_b}\frac{\partial \ell(\tilde{\bfP}^n,z^n;\Psi)}{\partial \bfW_Q^{(t)}}\Big|_{t=t_0}(\bfx_{query}^\top, \boldsymbol{{0}}^\top)^\top\Big|,\label{lm_eq: Q mu_j mu_a}
    \end{aligned}
\end{equation}
\begin{equation}
    \begin{aligned}
        \Big|(\bfnu_k^\top, \boldsymbol{0}^\top)\eta\frac{1}{B}\sum_{n\in\mathcal{B}_b}\frac{\partial \ell(\tilde{\bfP}^n,z^n;\Psi)}{\partial \bfW_Q^{(t)}}\Big|_{t=t_0}(\bfx_{query}^\top, \boldsymbol{{0}}^\top)^\top\Big|
        \lesssim    \frac{1}{ M_2}\Big|(\bfmu_j^\top, \boldsymbol{0}^\top)\eta\frac{1}{B}\sum_{n\in\mathcal{B}_b}\frac{\partial \ell(\tilde{\bfP}^n,z^n;\Psi)}{\partial \bfW_Q^{(t)}}\Big|_{t=t_0}(\bfx_{query}^\top, \boldsymbol{{0}}^\top)^\top\Big|,\label{lm_eq: Q mu_j nu}
    \end{aligned}
\end{equation}

\iffalse
\begin{equation}
    \begin{aligned}
        \Big|(\bfnu_b^\top, \boldsymbol{0}^\top)\eta\frac{1}{B}\sum_{n\in\mathcal{B}_b}\frac{\partial \ell(\tilde{\bfP}^n,z^n;\Psi)}{\partial \bfW_Q^{(t)}}\Big|_{t=t_0}(\bfmu_j^\top, \boldsymbol{{0}}^\top)^\top\Big|
        \lesssim    \frac{1}{ M}\Big|(\bfmu_j^\top, \bfq^\top)\eta\frac{1}{B}\sum_{n\in\mathcal{B}_b}\frac{\partial \ell(\tilde{\bfP}^n,z^n;\Psi)}{\partial \bfW_Q^{(t)}}\Big|_{t=t_0}(\bfmu_j^\top, \boldsymbol{{0}}^\top)^\top\Big|,\label{lm_eq: Q mu_j nu other}
    \end{aligned}
\end{equation}

\begin{equation}
    \begin{aligned}
        \Big\|\eta\frac{1}{B}\sum_{n\in\mathcal{B}_b}\frac{\partial \ell(\tilde{\bfP}^n,z^n;\Psi)}{\partial \bfW_Q^{(t)}}\Big|_{t=t_0+1}(\bfmu_j^\top, \boldsymbol{{0}}^\top)^\top\Big\|
        \gtrsim \eta\frac{1}{M_1}\sum_{b=0}^{t_0}\zeta_b\delta \gamma_b \beta^2,\label{lm_eq: Q mu_j norm}
    \end{aligned}
\end{equation}
\fi

\begin{equation}
    \begin{aligned}
        (\bfmu_j^\top, \boldsymbol{0}^\top)\eta\frac{1}{B}\sum_{n\in\mathcal{B}_b}\frac{\partial \ell(\tilde{\bfP}^n,z^n;\Psi)}{\partial \bfW_K^{(t)}}\Big|_{t=t_0+1}\bfp_q
        \gtrsim  \eta\frac{1}{M_1}\sum_{b=0}^{t_0}\zeta_b\delta \gamma_b \beta^4,\label{lm_eq: K mu_j mu_j}
    \end{aligned}
\end{equation}
\begin{equation}
    \begin{aligned}
        &\Big|(\bfmu_l^\top, \boldsymbol{0}^\top)\eta\frac{1}{B}\sum_{n\in\mathcal{B}_b}\frac{\partial \ell(\tilde{\bfP}^n,z^n;\Psi)}{\partial \bfW_K^{(t)}}\Big|_{t=t_0}\bfp_q\Big|\\
        \lesssim    &e^{-\Theta((\frac{\eta t_0}{M_1})^2)}\Big|(\bfmu_j^\top, \boldsymbol{0}^\top)\eta\frac{1}{B}\sum_{n\in\mathcal{B}_b}\frac{\partial \ell(\tilde{\bfP}^n,z^n;\Psi)}{\partial \bfW_K^{(t)}}\Big|_{t=t_0}\bfp_q\Big|,\label{lm_eq: K mu_j mu_l}
    \end{aligned}
\end{equation}
\begin{equation}
    \begin{aligned}
        \Big|(\bfmu_a^\top, \boldsymbol{0}^\top)\eta\frac{1}{B}\sum_{n\in\mathcal{B}_b}\frac{\partial \ell(\tilde{\bfP}^n,z^n;\Psi)}{\partial \bfW_K^{(t)}}\Big|_{t=t_0}\bfp_q\Big|
        \lesssim    \frac{1}{M_1}\Big|(\bfmu_j^\top, \boldsymbol{0}^\top)\eta\frac{1}{B}\sum_{n\in\mathcal{B}_b}\frac{\partial \ell(\tilde{\bfP}^n,z^n;\Psi)}{\partial \bfW_K^{(t)}}\Big|_{t=t_0}\bfp_q\Big|,\label{lm_eq: K mu_j mu_a}
    \end{aligned}
\end{equation}
\begin{equation}
    \begin{aligned}
        \Big|(\bfnu_k^\top, \boldsymbol{0}^\top)\eta\frac{1}{B}\sum_{n\in\mathcal{B}_b}\frac{\partial \ell(\tilde{\bfP}^n,z^n;\Psi)}{\partial \bfW_K^{(t)}}\Big|_{t=t_0}\bfp_q\Big|
        \lesssim    \frac{1}{ M_2}\Big|(\bfmu_j^\top, \bfq^\top)\eta\frac{1}{B}\sum_{n\in\mathcal{B}_b}\frac{\partial \ell(\tilde{\bfP}^n,z^n;\Psi)}{\partial \bfW_K^{(t)}}\Big|_{t=t_0}\bfp_q\Big|.\label{lm_eq: K mu_j nu}
    \end{aligned}
\end{equation}
\iffalse
\begin{equation}
    \begin{aligned}
        \Big|(\bfnu_b^\top, \bfq^\top)\eta\frac{1}{B}\sum_{n\in\mathcal{B}_b}\frac{\partial \ell(\tilde{\bfP}^n,z^n;\Psi)}{\partial \bfW_K^{(t)}}\Big|_{t=t_0}(\bfmu_j^\top, \boldsymbol{{0}}^\top)^\top\Big|
        \lesssim    \frac{1}{ M}\Big|(\bfmu_j^\top, \bfq^\top)\eta\frac{1}{B}\sum_{n\in\mathcal{B}_b}\frac{\partial \ell(\tilde{\bfP}^n,z^n;\Psi)}{\partial \bfW_K^{(t)}}\Big|_{t=t_0}(\bfmu_j^\top, \boldsymbol{{0}}^\top)^\top\Big|,\label{lm_eq: K mu_j nu other}
    \end{aligned}
\end{equation}

\begin{equation}
    \begin{aligned}
        \Big\|\eta\frac{1}{B}\sum_{n\in\mathcal{B}_b}\frac{\partial \ell(\tilde{\bfP}^n,z^n;\Psi)}{\partial \bfW_K^{(t)}}\Big|_{t=t_0+1}(\bfmu_j^\top, \pm\bfq^\top)^\top\Big\|
        \gtrsim \eta\frac{1}{M_1}\sum_{b=0}^{t_0}\zeta_b\delta \gamma_b \beta^2.\label{lm_eq: K mu_j norm}
    \end{aligned}
\end{equation}
\fi
In the above, equations (\ref{lm_eq: Q mu_j mu_j}), (\ref{lm_eq: Q mu_j mu_l}), (\ref{lm_eq: Q mu_j mu_a}), and (\ref{lm_eq: Q mu_j nu}) characterize the directions of gradient updates of $\bfW_Q$ when projected with $(\bfx_{query}^\top, \boldsymbol{0})^\top$. %The equation (\ref{lm_eq: Q mu_j norm}) shows the lower bound of the norm of gradient updates of $\bfW_Q$ when projected with $(\bfmu_j^\top, \boldsymbol{0})^\top$. 
Similarly, equations (\ref{lm_eq: K mu_j mu_j}), (\ref{lm_eq: K mu_j mu_l}), (
\ref{lm_eq: K mu_j mu_a}), and (\ref{lm_eq: K mu_j nu}) characterize the directions of gradient updates of $\bfW_K$ when projected with $\bfp_q, q\in[l]$.\\ %The equation (\ref{lm_eq: K mu_j norm}) shows the lower bound of the norm of gradient updates of $\bfW_K$ when projected with $(\bfmu_j^\top, \pm\bfq)^\top$.\\

\end{lemma}

\begin{lemma}\label{lemma: V}
(gradient updates of $\bfW_V$) For $\bfp_j^n$ defined in (\ref{eqn: data}) and $t_0\geq 1$, if $l=l_{tr}\gtrsim \max\{1,\frac{1}{\alpha\beta^2}\}$ and $BT\gtrsim \Theta(M_1^2)$, $B\gtrsim M_1$, we have that for $\bfp_j$ of which the corresponding label embedding is $\bfq$, 
    \begin{equation}
    \begin{aligned}
        &\quad\eta\frac{1}{B}\sum_{n\in\mathcal{B}_b}\sum_{b=0}^{t_0}\frac{\partial \ell(\tilde{\bfP}^n,z^n;\Psi)}{\partial \bfW_V^{(b)}}\bfp_j\\
        &=  \eta \sum_{b=0}^{t_0}(\sum_{i\in\mathcal{W}_n}V_i(b)\bfW_{O_{(i,\cdot)}}^{(b)}+\sum_{i\in\mathcal{U}_n}V_i(b)\bfW_{O_{(i,\cdot)}}^{(b)}+\sum_{i\notin\mathcal{W}_n\cup\mathcal{U}_n}V_i(b)\bfW_{O_{(i,\cdot)}}^{(b)}),\label{WV_expand}
    \end{aligned}
\end{equation}
where 
\begin{equation}
    -V_i(b)\gtrsim \beta^2(1-\gamma_t)/a,\ \ \ \ i\in\mathcal{W}_n,\label{Vi_1_lemma}
\end{equation}
\begin{equation}
    -V_i(b)\leq \frac{1}{\beta^2+1}V_j(b),\ \ \ \ i\in\mathcal{U}_n, j\in\mathcal{W}_n,\label{Vi_2_lemma}
\end{equation}
\begin{equation}
    |V_i(b)|\lesssim \sqrt{\frac{\log B}{B}}\cdot \frac{1}{a},\ \ \ \ i\notin\mathcal{W}_n\cup\mathcal{U}_n.\label{Vi_3_lemma}
\end{equation}
If the corresponding label embedding is $-\bfq$, we have the that (\ref{WV_expand}) holds with
\begin{equation}
    -V_i(b)\gtrsim \beta^2(1-\gamma_t)/a,\ \ \ \ i\in\mathcal{U}_n,\label{Vi_1_lemma_2}
\end{equation}
\begin{equation}
    -V_i(b)\leq \frac{1}{\beta^2+1}V_j(b),\ \ \ \ i\in\mathcal{W}_n, j\in\mathcal{U}_n,\label{Vi_2_lemma_2}
\end{equation}
\begin{equation}
    |V_i(b)|\lesssim \sqrt{\frac{\log B}{B}}\cdot \frac{1}{a},\ \ \ \ i\notin\mathcal{W}_n\cup\mathcal{U}_n.\label{Vi_3_lemma_2}
\end{equation}
We can also derive
\begin{equation}
    \begin{aligned}
        &\eta\frac{1}{B}\sum_{n\in\mathcal{B}_b}\frac{\partial \ell(\tilde{\bfP}^n,z^n;\Psi)}{\partial \bfW_V^{(t)}}(\bfnu_k^\top,\boldsymbol{0}^\top)^\top\\
    =: & \eta \sum_{b=0}^{t_0}(\sum_{i\in\mathcal{W}_n}V_i'(b)\bfW_{O_{(i,\cdot)}}^{(b)}+\sum_{i\in\mathcal{U}_n}V_i'(b)\bfW_{O_{(i,\cdot)}}^{(b)}+\sum_{i\notin\mathcal{W}_n\cup\mathcal{U}_n}V_i'(b)\bfW_{O_{(i,\cdot)}}^{(b)}),
    \end{aligned}
\end{equation}
where
\begin{equation}
    |V_i'(b)|\leq |V_i(b)|\cdot \frac{1}{M_2}.\label{V'_v}
\end{equation}

\end{lemma}
%\ST{Where is the proof of this lemma?}

\begin{lemma}\label{lemma: O}
(gradient updates of $\bfW_O$) We are given $\Theta(1)\geq\beta\geq 1$ and $m\gtrsim M_1^2$, $BT\gtrsim M_1\log M_1$, $B\gtrsim M_1$, $t=t_0\geq\Theta(1)$. Denote the set of examples that share the same IDR pattern as $\bfp_{query}^n$ as $\mathcal{B}_b^n$ in the $b$-th iteration. For $i\in\mathcal{W}$, $b\neq a$, and $\bfp_{query}^n$ corresponding to $\bfq$ and $\bfmu_a$, 
\begin{equation}
\begin{aligned}
    &\eta\frac{1}{|\mathcal{B}_b^n|}\sum_{n\in\mathcal{B}_b^n}\frac{\partial \ell(\tilde{\bfP}^n,z^n;\Psi)}{\partial \bfW_{O_{(i,\cdot)}}^{(t_0)}}(\bfmu_a^\top,\bfq^\top)^\top
    = \delta(\beta^2+1)\frac{\alpha\eta}{2a}(1+\frac{\eta^2 m}{a^2})^{t_0},
\end{aligned}
\end{equation}
\begin{equation}
\begin{aligned}
    &\eta\frac{1}{B}\sum_{b=0}^{t_0+1}\sum_{n\in\mathcal{B}_b}\frac{\partial \ell(\tilde{\bfP}^n,z^n;\Psi)}{\partial \bfW_{O_{(i,\cdot)}}^{(t_0)}}(\bfmu_a^\top,\bfq^\top)^\top
    \gtrsim \delta(\beta^2+1)\frac{\alpha\eta t_0}{2a}.
\end{aligned}
\end{equation}
For $i\in\mathcal{U}$, $b\neq a$, and $\bfp_{query}^n$ corresponding to $\bfq$ and $\bfmu_a$, 
\begin{equation}
\begin{aligned}
    &\eta\frac{1}{|\mathcal{B}_b^n|}\sum_{n\in\mathcal{B}_b^n}\frac{\partial \ell(\tilde{\bfP}^n,z^n;\Psi)}{\partial \bfW_{O_{(i,\cdot)}}^{(t_0)}}(\bfmu_a^\top,-\bfq^\top)^\top
    = \delta(\beta^2+1)\frac{\alpha\eta}{2a}(1+\frac{\eta^2 m}{a^2})^{t_0},
\end{aligned}
\end{equation}
\begin{equation}
\begin{aligned}
    &\eta\frac{1}{B}\sum_{b=0}^{t_0+1}\sum_{n\in\mathcal{B}_b}\frac{\partial \ell(\tilde{\bfP}^n,z^n;\Psi)}{\partial \bfW_{O_{(i,\cdot)}}^{(b)}}(\bfmu_a^\top,-\bfq^\top)^\top\\
    \gtrsim & \delta(\beta^2+1)\frac{\alpha\eta t_0}{2a}.
\end{aligned}
\end{equation}
For $i\in\mathcal{W}\cup\mathcal{U}$ and $c\in[M_2]$,
\begin{equation}
    \begin{aligned}
    \|\bfW_{O_{(i,\cdot)}}^{(t_0)}\|
    \gtrsim \sqrt{M_1}\delta(\beta^2+1)^\frac{1}{2}\frac{\alpha\eta t_0}{2a},\label{WO_norm_lemma}
    \end{aligned}
\end{equation}
\begin{equation}
    \begin{aligned}
        &\eta\frac{1}{B}\sum_{b=0}^{t_0}\sum_{n\in\mathcal{B}_b}\frac{\partial \ell(\tilde{\bfP}^n,z^n;\Psi)}{\partial \bfW_{O_{(i,\cdot)}}^{(b)}}(\bfnu_c^\top,\pm\bfq^\top)^\top
        \leq  \frac{1}{M_2}\eta\frac{1}{B}\sum_{b=0}^{t_0}\sum_{n\in\mathcal{B}_b}\frac{\partial \ell(\tilde{\bfP}^n,z^n;\Psi)}{\partial \bfW_{O_{(i,\cdot)}}^{(b)}}(\bfmu_b^\top,\bfq^\top)^\top\label{WO_v}.
        \end{aligned}
\end{equation}
For $i\notin\mathcal{W}\cup\mathcal{U}$, we have
\begin{equation}
        \eta\frac{1}{B}\sum_{b=0}^{t_0}\sum_{n\in\mathcal{B}_b}\frac{\partial \ell(\tilde{\bfP}^n,z^n;\Psi)}{\partial \bfW_{O_{(i,\cdot)}}^{(b)}}(\bfmu_a^\top,\pm\bfq^\top)^\top\leq \eta t\sqrt{\frac{\log B}{B}}\frac{1}{a}.\label{O_notin}
\end{equation}
\end{lemma}
%\ST{Where is the proof of this lemma?}

\iffalse
\begin{definition}\label{def: WU}
Let $\theta_k^i$ be the angle between $\bfmu_k$ and $\bfW_{O_{(i,\cdot)}}[:,0:d_\mathcal{X}]$, $k\in[M_1]$. Let ${\theta_1^i}'$ be the angle between $\bfq$ and $\bfW_{O_{(i,\cdot)}}[:,d_\mathcal{X}:d_{\mathcal{X}+d_\mathcal{Y}}]$. Let ${\theta_2^i}'$ be the angle between $-\bfq$ and $\bfW_{O_{(i,\cdot)}}[:,d_\mathcal{X}:d_{\mathcal{X}+d_\mathcal{Y}}]$. Define $\mathcal{W}_k(t)$, $\mathcal{U}_k(t)$ as the sets of lucky neurons at the $t$-th iteration such that
\begin{equation}
    \mathcal{W}_k(t)=\{i: \theta_k^i+ \leq \theta_k^j,\ {\theta_1^i}'+\leq {\theta_2^i}', i\in[M_1], \forall j\in[M_1]\}
\end{equation}
\begin{equation}
    \mathcal{U}_k(t)=\{i: \theta_k^i+ \leq \theta_k^j, \ {\theta_2^i}'+\leq {\theta_1^i}', i\in[M_1], \forall j\in[M_1]\}
\end{equation}
\end{definition}
\fi

\begin{definition}\label{def: WU}
Define 
\begin{equation}
    \bfV^n(t):=\sum_{s=1}^{l+1}\bfW_V^{(t)}\bfp_s^n\text{softmax}({\bfp_s^n}^\top{\bfW_K^{(t)}}^\top\bfW_Q^{(t)}\bfp_{query}^n),
\end{equation}
for $\bfP^n$. Let $\bfW_{O_{(i,\cdot)}}=(\bfO_{i,1}, \bfO_{i,2}, \boldsymbol{0}^\top)$ where $\bfO_{i,1}\in\mathbb{R}^{d_\mathcal{X}}, \bfO_{i,2}\in\mathbb{R}^{d_\mathcal{Y}}$. Let $\bfV^n(t)=(\bfV_{n,1}(t)^\top, \bfV_{n,2}(t)^\top, \boldsymbol{0}^\top)^\top$ where $\bfV_{i,1}(t)\in\mathbb{R}^{d_\mathcal{X}}, \bfV_{i,2}(t)\in\mathbb{R}^{d_\mathcal{Y}}$. Define $\mathcal{W}_n$, $\mathcal{U}_n$ as the sets of lucky neurons such that
\begin{equation}
    \mathcal{W}_n=\{i: \bfO_{i,1}^{(0)}\bfV_{n,1}(0)>0, \bfO_{i,2}^{(0)}\bfV_{n,2}(0)>0, a_i>0\},\label{W_def}
\end{equation}
\begin{equation}
    \mathcal{U}_n=\{i: \bfO_{i,1}^{(0)}\bfV_{n,1}(0)>0, \bfO_{i,2}^{(0)}\bfV_{n,2}(0)>0, a_i<0\}.\label{U_def}
\end{equation}
Define
\begin{equation}
    \mathcal{N}^{n,i}_j=\{i: i\in[l+1], \bfx_i^n=\bfmu_j+\kappa_i^n\bfnu_k+\bfn_i^n, k\in[M_2]\},
\end{equation}
\begin{equation}
    \mathcal{M}^{n,i}_k=\{i: i\in[l+1], \bfx_i^n=\bfmu_j+\kappa_i^n\bfnu_k+\bfn_i^n, j\in[M_1]\},
\end{equation}
as the set of example inputs with $\bfmu_j$ as the IDR patterns or with $\bfnu_k$ as the IDI patterns, respectively.
\begin{equation}
    \mathcal{W}=\bigcup_{n=1}^{N}\mathcal{W}_n,\ \ \ \ \mathcal{U}=\bigcup_{n=1}^{N}\mathcal{U}_n.
\end{equation}
\end{definition}
%\ST{Where is the proof of this lemma?}

%\ST{how can you guarantee this?proof of lemma 9?}
\begin{lemma}\label{lemma: initial_WU} 
By the definition of lucky neurons in (\ref{W_def}) and (\ref{U_def}), and the initialization described in Section \ref{subsec: SGD}, the number of lucky neurons $|\mathcal{W}_n|$, $|\mathcal{U}_n|$ satisfies
\begin{equation}
    |\mathcal{W}_n|,\ |\mathcal{U}_n|\geq \Omega(m).
\end{equation}
Hence, 
\begin{equation}
    |\mathcal{W}|,\ |\mathcal{U}|\geq \Omega(m).
\end{equation}
\end{lemma}

\begin{lemma}\label{lemma: update_WU}
Under the condition that $m\gtrsim M_1^2\log M_1$, we have the following results.
\begin{enumerate}
    \item When $t\geq 0$, for $V^n(t)$ where $\bfp_{query}^n$ corresponds to the label $+1$,
    \begin{equation}
    \mathbbm{1}[\bfW_{O_{(i,\cdot)}}^{(t)}\bfV^n(t)]=1, i\in\mathcal{W}_n,
\end{equation}
for $V^n(t)$ where $\bfp_{query}^n$ corresponds to the label $-1$,
\begin{equation}
    \mathbbm{1}[\bfW_{O_{(i,\cdot)}}^{(t)}\bfV^n(t)]=1, i\in\mathcal{U}_n.
\end{equation}
\item When $t\geq \Theta(1)$, for $i\in\mathcal{W}$, we have that for $V^n(t)$ where $\bfp_{query}^n$ corresponds to the label $+1$, 
\begin{equation}
    \mathbbm{1}[\bfW_{O_{(i,\cdot)}}^{(t)}\bfV^n(t)]=1.\label{WO_indicator_W}
\end{equation}
For $i\in\mathcal{U}$, we have that for $V^n(t)$ where $\bfp_{query}^n$ corresponds to the label $-1$, 
\begin{equation}
    \mathbbm{1}[\bfW_{O_{(i,\cdot)}}^{(t)}\bfV^n(t)]=1.\label{WO_indicator_U}
\end{equation}
\end{enumerate}
%(\ref{WO_indicator_W}) and (\ref{WO_indicator_U}) hold for $i\in\mathcal{W}$ and $i\in\mathcal{U}$ when $t\geq 1$, respectively.
\end{lemma}

\begin{lemma}\label{lemma: mini_task}
    With in-domain tasks defined in Definition \ref{def: task} and Condition \ref{cond: task}, the number of training tasks should satisfy $|\mathcal{T}_{tr}|\geq M_1$ to make Condition \ref{cond: task} hold. 
\end{lemma}

\subsection{Proof of Theorem \ref{thm: training}}
\begin{proof}
We first look at the required length of the prompt. Define $m_i$ as the corresponding IDR pattern in the $i$-th demonstration. Consider the categorical distribution where the probabilities of selecting $\bfmu_a$ and $\bfmu_b$ are $\alpha/2$ respectively. By the Chernoff bound of Bernoulli distribution in Lemma \ref{lemma: chernoff}, we can obtain
\begin{equation}
    \Pr\left(\frac{1}{l_{tr}}\sum_{i=1}^{l_{tr}}\mathbbm{1}[m_i=\bfmu_a] \leq (1-c)\frac{\alpha}{2}\right)\leq e^{-l_{tr} c^2\frac{\alpha}{2}}= M_1^{-C},
\end{equation}
for some $c\in(0,1)$ and $C>0$. Hence, with a high probability, combining the condition $l_{tr}\geq (\alpha\beta^2)^{-1}$ in Lemma \ref{lemma: V}, %\ST{combine what???? 203? some typos in 203???? use equation 203 to prove equation 67? }
\begin{equation}
    l_{tr}\gtrsim  \max\left\{\Omega\left(\frac{2\log M_1}{\alpha}\right),\Omega\left(\frac{1}{\alpha\beta^2}\right)\right\}.
\end{equation}
\iffalse
By the solution to the Coupon collector's problem, we know that
\begin{equation}
    B\geq M_1\log M_1,
\end{equation}
\fi
By the condition in Lemma \ref{lemma: QK}, we have that
\begin{equation}
    B\geq \Omega(M_1\log M_1).
\end{equation}
We know that there exists gradient noise caused by imbalanced IDR patterns in each batch% is in the order of $1/M_2$ scaling of each gradient by (\ref{lm_eq: Q mu_j nu}), (\ref{lm_eq: K mu_j nu}), (\ref{V'_v}), and (\ref{WO_v}). 
Therefore, by Hoeffding's inequality (\ref{hoeffding}), for any $\bfW\in\Psi$,
\begin{equation}
    \Pr\left(\Big\|\frac{1}{|\mathcal{B}_b|}\sum_{n\in\mathcal{B}_b}\frac{\partial\ell(\Psi; \bfP^n,z^n)}{\partial \bfW}-\mathbb{E}\left[\frac{\partial\ell(\Psi; \bfP^n,z^n)}{\partial \bfW}\right]\Big\|\geq \Big|\mathbb{E}\left[\frac{\partial\ell(\Psi; \bfP^n,z^n)}{\partial \bfW}\right]\epsilon\right)\leq e^{-B\epsilon^2}\leq M_1^{-C},\label{grad_noise}
\end{equation}
if $B\gtrsim \epsilon^{-2}\log M_1$. 
Therefore, we require
\begin{equation}
    B\gtrsim \max\{\epsilon^{-2}, M_1\}\log M_1.
\end{equation}
\iffalse
The gradient noise in each iteration comes from the IDI pattern and random noise terms in each query or example. By examining all gradient updates related to IDI patterns or random noise in Lemmas \ref{lemma: QK}, \ref{lemma: V}, and \ref{lemma: O}, we have that the noise level is at most a $1/M_1$ scaling of the mean gradient. Let $s_i$ be the gradient noise level compared to the mean gradient with $|s_i|\leq 1/M_1$. Hence, the final network output is within a scaling of $[1-1/M_1, 1+1/M_1]$ of using clean gradients. \\
\fi
\noindent \textbf{(a)} We have that for $i$ such that $a_i>0$ but $i\notin \mathcal{W}$ by the definition of the Relu function,
\begin{equation}
    a_i\text{Relu}(\bfW_{O_{(i,\cdot)}}^{(T)}\sum_{s=1}^{l+1}(\bfW_V^{(T)}\bfp_s^n )\text{softmax}((\bfW_K^{(T)}\bfp_s^n )^\top(\bfW_Q^{(T)}\bfp_{query}^n )))\geq 0.\label{W_notin}
\end{equation}
\textbf{(b)} Furthermore, we have that for $i\in\mathcal{W}$, and for $\bfp_s^n$ that shares the same IDR pattern as $\bfp_{query}^n$, with a high probability of $1-M_1^{-C}$,%\ST{what do you mean here? i and p shares?? why? also what probability?} \ST{How high is high? from 66?}
\begin{equation}
    \begin{aligned}
        &\eta\sum_{b=0}^{T-1}\bfW_{O_{(j,\cdot)}}^{(T)}\sum_{j\in\mathcal{W}_n}{\bfW_{O_{(j,\cdot)}}^{(b)}}^\top\\
        \geq & \eta\sum_{b=0}^{T-1} M_1 \delta(\beta^2+1)^\frac{1}{2}\frac{\alpha\eta T}{2a}\cdot \delta(\beta^2+1)^\frac{1}{2}\frac{\alpha\eta b}{2a}\\
        \gtrsim & \delta^2(\beta^2+1)\alpha^2 \frac{(\eta T)^3 M_1}{a^2},
        \end{aligned}
\end{equation}
where the first step comes from (\ref{WO_norm_lemma}) in Lemma \ref{lemma: O}, %\ST{eqaution 51?? substitute 51 here? } \ST{also the left hand side of this inequality. $W^T_O$ should be outside of the double sum. $b$ corresponds to $W^b_0$, right? $i$ is only applied to $W^b_0$} 
and the second step is by $\sum_{b=0}^{T-1}b=\Theta(T^2)$. Then, we can obtain
\begin{equation}
\begin{aligned}
    &\bfW_{O_{(i,\cdot)}}^{(T)}\bfW_V^{(T)}\bfp_s^n\\
    = & \bfW_{O_{(i,\cdot)}}^{(T)}(\delta{\bfp_s^n}+\sum_{b=0}^{T-1}\eta(\sum_{i\in\mathcal{W}_n}V_i(b)\bfW_{O_{(i,\cdot)}}^{(b)}+\sum_{i\in\mathcal{U}_n}V_i(b)\bfW_{O_{(i,\cdot)}}^{(b)}+\sum_{i\notin\mathcal{W}_n\cup\mathcal{U}_n}V_i(b)\bfW_{O_{(i,\cdot)}}^{(b)})^\top)\\
    \gtrsim & \delta^2(\beta^2+1)\frac{\alpha\eta T}{2a}+\delta^2(\beta^2+1)\alpha^2 \frac{(\eta T)^3 M_1}{a^2},\label{WOVp_1}
\end{aligned}
\end{equation}
where the first step is by (\ref{WV_expand}) in Lemma \ref{lemma: V}, and the last step comes from Lemma \ref{lemma: O}. 

Therefore,  by combining Lemma \ref{lemma: initial_WU} and Lemma \ref{lemma: update_WU}, we have that 
\begin{equation}
    \begin{aligned}
        &\sum_{i\in\mathcal{W}}a_i\text{Relu}(\bfW_{O_{(i,\cdot)}}^{(T)}\sum_{s=1}^{l+1}(\bfW_V^{(T)}\bfp_s^n )\text{softmax}((\bfW_K^{(T)}\bfp_s^n )^\top(\bfW_Q^{(T)}\bfp_{query}^n )))\\
        \gtrsim & (1-\gamma_T)\cdot \Big(\delta^2(\beta^2+1)\frac{\alpha\eta T}{2a}+\delta^2(\beta^2+1)\alpha^2 \frac{(\eta T)^3 M_1}{a^2}\Big),
    \end{aligned}\label{W(t)_lb_v1}
\end{equation}
when $\gamma_T$ is order-wise smaller than $1$.% \ST{Why is it true? which part of these lemmas did you use? did  you use the definition of relu? write the proofs line by line.} 
We next give a bound for $\gamma_T$, which is give by Definition \ref{def: zeta},
\begin{equation}
    \gamma_T\geq 1-\sum_{s\in\mathcal{N}_*^n}\text{softmax}((\bfW_K^{(T)}\bfp_s^n )^\top(\bfW_Q^{(T)}\bfp_{query}^n ))\label{1-gammaT},
\end{equation}
from Defition \ref{def: zeta} for $\bfmu_j$ as the IDR pattern in $\bfp_{query}^n$. We can tell from (\ref{W(t)_lb_v1}) and Definition \ref{def: zeta}, 
\begin{equation}
\begin{aligned}
    \zeta_b\gtrsim &\delta^2(\beta^2+1)\frac{\alpha\eta T}{2a}+\delta^2(\beta^2+1)\alpha^2 \frac{(\eta T)^3 M_1}{a^2}.\label{zeta_b}
\end{aligned}
\end{equation}
Then, if $\zeta_T\gtrsim 1$ and $T\gtrsim M_1$, by Lemma \ref{lemma: QK}, with high probability,
\begin{equation}
\begin{aligned}    
    &(\bfW_K^{(T)}\bfp_s^n )^\top(\bfW_Q^{(T)}\bfp_{query}^n )\\
    \gtrsim & (\eta\frac{1}{M_1}\sum_{b=0}^{T-1}\zeta_b\delta \gamma_b \beta^2+\delta)^2\\
    \gtrsim & (\eta\sum_{b=0}^{T-1}\gamma_b\beta^2\cdot (\delta^2\frac{\alpha\eta T}{2a }+\delta^2(\beta^2+1)\alpha^2 \frac{(\eta T)^3 M_1}{a^2})+\delta)^2\\
    :=& (\eta\sum_{b=0}^{T-1}\gamma_b\cdot \Delta_T+\delta)^2,\label{attention_T}
\end{aligned}
\end{equation}
where the first step comes from the fact that the gradient update projections of $\bfW_Q$ and $\bfW_K$ onto queries or examples are close to the corresponding IDR pattern the most by Lemma \ref{lemma: QK}. In the last inequality of (\ref{attention_T}), we only consider the term related to $T$ and $\gamma_b$. For any $\bfp_l^n$ that shares a different IDR pattern as $\bfp_{query}^n$, we have
\begin{equation}
\begin{aligned}    
    &(\bfW_K^{(T)}\bfp_l^n )^\top(\bfW_Q^{(T)}\bfp_{query}^n )\lesssim \frac{1}{M_1}(\bfW_K^{(T)}\bfp_s^n )^\top(\bfW_Q^{(T)}\bfp_{query}^n ),\label{attention_T_small}
\end{aligned}
\end{equation}
by Lemma \ref{lemma: QK}. Then, given the definition of softmax, 

\begin{equation}
\begin{aligned}
    &\sum_{s\in\mathcal{N}_j^n}\text{softmax}({\bfp_s^n}^\top{\bfW_K^{(t)}}^\top\bfW_Q^{(t)}\bfp_{query}^n)\\
    \geq & \frac{\sum_{s\in\mathcal{N}_j^n}e^{\Theta(\delta^2)+(\eta\sum_{b=0}^{T-1}\gamma_b\cdot \Delta_T)^2}}{\sum_{s\in\mathcal{N}_j^n}e^{\Theta(\delta^2)+(\eta\sum_{b=0}^{T-1}\gamma_b\cdot \Delta_T)^2}+\sum_{s\in[l]-\mathcal{N}_j^n}e^{\frac{1}{M_1}(\eta\sum_{b=0}^{T-1}\gamma_b\cdot \Delta_T)^2}}\\
    \geq & 1-\frac{2-\alpha}{\alpha}e^{-(\eta\sum_{b=0}^{T-1}\gamma_b\cdot \Delta_T)^2},\label{attn_train}
\end{aligned}
\end{equation}
where the first step is by (\ref{attention_T}) and (\ref{attention_T_small}). Combining with (\ref{1-gammaT}), we can derive
\begin{equation}
\begin{aligned}
    \gamma_T\leq &\frac{2-\alpha}{\alpha}e^{-(\eta\sum_{b=0}^{T-1}\gamma_b\cdot \Delta_T)^2}=\frac{2-\alpha}{\alpha}e^{-(\eta\sum_{b=0}^{T-2}\gamma_b\cdot \Delta_T)^2}\cdot e^{-\eta^2 \Delta_T^2(2\gamma_{T-1}\sum_{b=0}^{T-2}\gamma_b+\gamma_{T-1}^2)}\\
    =& \gamma_{T-1}\cdot e^{-\eta^2 \Delta_T^2(2\gamma_{T-1}\sum_{b=0}^{T-2}\gamma_b+\gamma_{T-1}^2)}\label{gamma_update}.
\end{aligned}
\end{equation}
When $T$ is large, $\gamma_T$ is approaching zero. Hence, the equality of (\ref{gamma_update}) is close to being achieved, in which case,
\begin{equation}
    \gamma_T\approx \gamma_{T-1}\cdot e^{-\eta^2 \Delta_T^2(2\gamma_{T-1}\sum_{b=0}^{T-2}\gamma_b+\gamma_{T-1}^2)}.\label{gamma_T_one}
\end{equation}
We can observe that when $\sum_{b=0}^{t_0-1}\eta\gamma_b\Delta_T\geq  \sqrt{\log M_1}$, $\gamma_{t_0}$ reaches $\Theta(1/M_1\cdot \frac{2-\alpha}{\alpha})$. Similarly, when $\sum_{b=0}^{t_0'-1}\eta\gamma_b\Delta_T\leq  \sqrt{\log C}$ for some $C>1$, $\gamma_{t_0'}$ is still $\Theta(1)$, which indicates $t_0'\lesssim \eta^{-1}M_1\sqrt{\log C}$ if we only care about $\eta$ and $M_1$ as variables. Therefore, we require that the final $T$ satisfies $T\gtrsim  \eta^{-1}M_1\sqrt{\log M_1}$. %Since we require $M\geq M_1^2$, we have $T\geq \Theta(\eta^{-1}((1-\alpha)/\alpha\cdot M M_1\sqrt{\log C})^\frac{1}{3})$. Therefore, we can conclude that $\gamma_T=\Theta(1/M)$. Then, for some large $C>1$,
\iffalse
\begin{equation}
    \begin{aligned}
        &\sum_{i\in\mathcal{W}}a_i\text{Relu}(\bfW_{O_{(i,\cdot)}}^{(T)}\sum_{s=1}^{l+1}(\bfW_V^{(T)}\bfp_s^n )\text{softmax}((\bfW_K^{(T)}\bfp_s^n )^\top(\bfW_Q^{(T)}\bfp_{query}^n )))\\
        \gtrsim & \frac{m}{a}(1-\epsilon_m- M_1)(1-\frac{1}{M})\cdot \Big(\frac{\eta T}{a}(1-\frac{1}{C}\alpha^{-1})+\frac{\eta^2 T^2}{a}(1-\frac{1}{C}\alpha^{-1})^2\beta^2\\
        &\cdot (1-\epsilon_m- M_1)
    + \frac{\eta^2T^2}{a}((1-\frac{1}{C}\alpha^{-1})^2+(1-\frac{1}{C}\alpha^{-1})^3\eta T(1-\epsilon_m- M_1)\beta^2 )\beta^2\frac{1}{a}\Big)\label{Wt_further}.
    \end{aligned}
\end{equation}
\fi
\\
\noindent 
\textbf{(c)} We next look at $i$ where $a_i<0$. If $i\in\mathcal{U}$, we have that for $s$ such that the $y$-embedding of $\bfp_s^n$ is $\bfq$,  the summation of corresponding softmax value is $1-\gamma_T$. Furthermore, 
\begin{equation}
    \begin{aligned}
        &\bfW_{O_{(i,\cdot)}}^{(T)}\bfW_V^{(T)}\bfp_s^n\\
        \lesssim & -\delta^2\frac{\alpha\eta T}{2a}-\delta^2\alpha^2 \frac{(\eta T)^3 M_1}{a^2},
    \end{aligned}
\end{equation}
if $\beta\leq \Theta(1)$. Hence, 
\begin{equation}
    \text{Relu}(\bfW_{O_{(i,\cdot)}}^{(T)}\sum_{s=1}^{l+1}(\bfW_V^{(T)}\bfp_s^n )\text{softmax}((\bfW_K^{(T)}\bfp_s^n )^\top(\bfW_Q^{(T)}\bfp_{query}^n )))= 0\label{notin1}.
\end{equation}
\textbf{(d)} If $i\notin \mathcal{W}\cup\mathcal{U}$ and $s\in\mathcal{W}$, we have,
\begin{equation}
 \bfW_{O_{(i,\cdot)}}^{(T)}\bfW_V^{(T)}\bfp_s^n
       \lesssim  \frac{1}{\sqrt{M_1}} \bfW_{O_{(i,\cdot)}}^{(T)}\bfW_V^{(T)}\bfp_s^n\label{notin2},
\end{equation}
by Lemma \ref{lemma: V} and $B\gtrsim M_1$. 
\iffalse
\begin{equation}
    \begin{aligned}
        &\quad\bfW_{O_{(i,\cdot)}}^{(T)}\bfW_V^{(T)}\bfp_s^n\\
       & \lesssim  \eta T\sqrt{\frac{\log BT}{BT}}\frac{1}{a}\cdot \|\bfW_V^{(T)}\bfp_s^n\|\\
       & \lesssim  \eta\sqrt{\frac{\log BT}{BT}}\frac{1}{a}\cdot (\delta+\frac{(\eta T)^2\sqrt{M_1} }{a})\\
       & \lesssim   M_1^{-2}+M_1^{-3/2}\label{notin2},
    \end{aligned}
\end{equation}
where the first step is by (\ref{O_notin}) in Lemma \ref{lemma: O}, and last step holds when $BT=\Theta(M_1^2)$. 
\fi
The final lower bound of $F(\Psi;\bfP^n)$ is based on the lower bound introduced by $i\in\mathcal{W}$. \\
\noindent Then, combining (\ref{W_notin}), (\ref{W(t)_lb_v1}), (\ref{notin1}), and (\ref{notin2}), we can derive
\begin{equation}
    \begin{aligned}
        &F(\Psi; \bfP^n)\\
    \gtrsim &(1-\gamma_T)\cdot \Big(\delta^2(\beta^2+1)\frac{\alpha\eta T}{2a}+\delta^2(\beta^2+1)\alpha^2 \frac{(\eta T)^3 M_1}{a^2}\Big).\label{F1_lower}
    \end{aligned}
\end{equation}

\noindent Therefore, as long as 
\begin{equation}
    T=\Theta(\eta^{-1}M_1\delta^{-2/3}\beta^{-2/3}\alpha^{-2/3}\sqrt{\log M_1}),\label{cond_T}
\end{equation}
for some large $C>1$, we can obtain
\begin{equation}
    F(\Psi; \bfP^n)\geq 1.\label{F_lower}
\end{equation}

\noindent Similarly, we can derive that for $z^n=-1$,
\begin{equation}
    F(\Psi;\bfP^n)\leq -1.
\end{equation}

\iffalse
{\color{red}
\begin{align}
    1-Ct^3<\epsilon, t^3>\frac{1}{C}
\end{align}

}
\fi

\textbf{Then, we study in-domain generalization.} By (\ref{grad_noise}), for any given testing prompt embedding $\bfP$ with $z=+1$, we have
\begin{equation}
    F(\Psi; \bfP)\geq 1-\epsilon,
\end{equation}
and if $z=-1$, 
\begin{equation}
    F(\Psi; \bfP)\leq -1+\epsilon.
\end{equation}
Therefore, 
\begin{equation}
    \underset{\bfx_{query}\sim\mathcal{D}, f\in\mathcal{T}}{\mathbb{E}} [\ell(\Psi; \bfP, y)]\leq \epsilon.
\end{equation} 
\end{proof}

\subsection{Proof of Theorem \ref{thm: prompts&queries}}
\begin{proof}
\iffalse
Note that we need at least one demonstration of the same testing-relevant pattern $\bfmu_l'$ as the query in the testing prompt. Define $m_i$ as the corresponding testing-relevant pattern in the $i$-th demonstration in the testing prompt. Hence, by the Chernoff bound of Bernoulli distribution in Lemma \ref{lemma: chernoff}, we can obtain
\begin{equation}
    \Pr\left(\frac{1}{l_{ts}}\sum_{i=1}^l\mathbbm{1}[m_i=\bfmu_l'] \geq (1+c)\frac{\alpha'}{2}\right)\leq e^{-l_{ts} c^2\frac{\alpha}{2}}\leq  q^{-C},
\end{equation}
for some $q, C>1$ and $c>0$. Then, we can obtain
\begin{equation}
  l_{ts}\geq \frac{2}{\alpha'}\log q.
\end{equation}
\fi
Note that we require that the fraction of contexts with the same ODR pattern as the query is at least $\alpha'$. Since we need that there exists at least one context that contains the same ODR pattern as the query, we have
\begin{equation}
    l_{ts}\geq \frac{1}{\alpha'}.
\end{equation}
Consider ${\bfp_{query}^n}'$ such that the label is $+1$. Let $\bfmu_j'=\sum_{j=1}^{M_1}c_j\bfmu_j$ where $\sum_{j=1}^{M_1}c_j^2=1$ and $\bfnu_k'=\sum_{j=1}^{M_2}g_j \bfnu_j$ where $\sum_{j=1}^{M_2}g_j^2=1$. Following the derivation of (\ref{zeta_b}) and (\ref{attention_T}), we have that for $s\in\mathcal{N}^n$, 
\begin{equation}
\begin{aligned}    
    &(\bfW_K^{(T)}{{\bfp_s^n}'})^\top\bfW_Q^{(T)}{{\bfp_{query}^n}'}\\
    \gtrsim & \sum_{j=1}^{M_1} c_j^2\cdot(\eta\sum_{b=0}^{T-1}\gamma_b\beta^2\cdot (\delta^2\frac{\alpha\eta T}{2a }+\delta^2(\beta+1)^2\alpha^2 \frac{(\eta T)^3 M_1}{a^2}))^2\\
    \gtrsim & \log M_1.
\end{aligned}
\end{equation}
For ODR patterns, by Proposition \ref{prop: self-attention}, we have for $\bfp_l^n$ that has a different ODR pattern than $\bfp_{query}^n$,
    \begin{equation}
\begin{aligned}    
    &(\bfW_K^{(T)}\bfp_l^n )^\top(\bfW_Q^{(T)}\bfp_{query}^n )\lesssim (\frac{1}{M_1}+\frac{1}{M_2})(\bfW_K^{(T)}\bfp_s^n )^\top(\bfW_Q^{(T)}\bfp_{query}^n ).
\end{aligned}
\end{equation}
Therefore, by similarly defining $\mathcal{N}_j^n=\{{\bfp_s^n}': \text{The testing-relevant pattern of }\bfP^n \text{ is }\bfmu_j'\}$, we can derive
\begin{equation}
    \sum_{s\in\mathcal{N}_j^n}\text{softmax}((\bfW_K^{(T)}{\bfp_s^n}')^\top(\bfW_Q^{(T)}{\bfp_{query}^n}'))\geq 1-\frac{2-\alpha'}{\alpha'}\Theta(\frac{1}{M_1})\geq 1-\frac{2}{\alpha'}\Theta(\frac{1}{M_1}).\label{attn_ODR}
\end{equation}
%We can always find a set of $\bfp_t$ such that ${\bfp_s^n}'$ is a linear combination of $\bfp_t$. Specifically, for ${\bfp_s^n}'=\sum_{i=1}^{M_1} c_i\bfmu_i+\sum_{j=1}^{M_2}{\kappa_s^n}' g_j\bfnu_j$, we need that $c_i=\sum_{t=1}^{Q_i}\alpha_t$ and ${\kappa_s^n}'g_j=\sum_{t=1}^{C_j}\alpha_t\kappa_t$ for some $Q_i\geq 1$ and $R_j\geq 1$. 
Note that for ${\bfp_s^n}'=\sum_{i=1}^{M_1} c_i\bfmu_i+\sum_{j=1}^{M_2}{\kappa_s^n}' g_j\bfnu_j$, when $M_1\geq M_2$, we can find a set of $\bfmu_j+{\kappa_s^n}'\bfnu_j$ from $j=1$ to $j=M_2$ with $g_j$ as the coefficients. When $M_1< M2=\Theta(M_1)$, we can find a set of $\bfmu_t+{\kappa_s^n}'\bfnu_j$ from $j=1$ to $j=M_2$ with $t\in[M_1]$, $g_j$ as the coefficients likewise. The remaining $\bfmu_i$ has coefficients of which the summation is smaller than $1$.
Therefore, we have that for a certain $i\in\mathcal{W}$  and $\bfp_s^n$ where the corresponding label-space embedding is $\bfq$, 
\begin{equation}
\begin{aligned}
    &\bfW_{O_{(i,\cdot)}}^{(T)}\bfW_V^{(T)}{\bfp_s^n}'\\
    = & \bfW_{O_{(i,\cdot)}}^{(T)}\bfW_V^{(T)}(\sum_{i=1}^{M_1}c_i{\bfmu_i}^\top+{\kappa_s^n}'\sum_{j=1}^{M_2}g_j{\bfnu_j}^\top+{\bfo_s^n}^\top, \bfq^\top)^\top\\
    \gtrsim &\sum_{i=1}^{M_1}c_i (\delta^2\beta^2\frac{\alpha\eta T}{2a}+\delta^2(\beta^2\alpha^2 \frac{(\eta T)^3 M_1}{a^2})\\
    &+(\delta^2\frac{\alpha\eta T}{2a}+\delta^2\alpha^2 \frac{(\eta T)^3 M_1}{a^2})(1-\epsilon)\\
    \geq & (\delta^2(\beta^2+1)\frac{\alpha\eta T}{2a}+\delta^2(\beta^2+1)\alpha^2 \frac{(\eta T)^3 M_1}{a^2})\cdot(1-\epsilon)\label{ood_lower},
\end{aligned}
\end{equation}
where the first equality comes from the definition of ${\bfp_s^n}'$. The first inequality of (\ref{ood_lower}) is derived from (\ref{grad_noise}). %The second inequality is because $\sum_{j=1}^{M_2}|g_j|\leq \sqrt{\sum_{j=1}^{M_2}g_j^2}=\sqrt{M_2}$ by Cauchy–Schwarz inequality. 
The last inequality is by the condition $\sum_{i=1}^{M_1}c_i\geq 1$. 
Therefore, we can derive that
\begin{equation}
\begin{aligned}
    F(\Psi; {\bfP^n}')\gtrsim &(1-\gamma_T )(\delta^2(\beta^2+1)\frac{\alpha\eta T}{2a}+\delta^2(\beta^2+1)\alpha^2 \frac{(\eta T)^3 M_1}{a^2})&\cdot(1-\epsilon)\\
    \geq &1-\epsilon,
\end{aligned}
\end{equation}
where the first step is by following (\ref{F_lower}), and the remaining steps are from basic mathematical computation. 
\noindent Likewise, for ${\bfp_{query}^n}'$ such that the label is $-1$, we can obtain
\begin{equation}
    F(\Psi; {\bfP^n}')<-(1-\epsilon).
\end{equation}
\iffalse
Then, by Hoeffding's inequality (\ref{hoeffding}), we can obtain
\begin{equation}
    \begin{aligned}
        \Pr\Big(\Big\|\frac{1}{T}\sum_{n=1}^{BT}\ell(\Psi;\bfP^n,z^n)-\mathbb{E}[\ell(\Psi;\bfP^n,z^n)]\Big\|\geq \epsilon\Big)=e^{-BT\frac{\epsilon^2}{\frac{1}{\beta^2 M_1^2}}}\leq \frac{1}{M_1^C}
    \end{aligned}
\end{equation}
for some large $C>1$, where the last step holds if 
\begin{equation}
    BT\geq \epsilon^{-2}(\beta^{-2}M_1^{-2})\log M_1,
\end{equation}
which already holds given (\ref{cond_T}). 
\fi
Therefore,  we have
\begin{equation}
    \underset{\bfx_{query}\sim\mathcal{D}', f\in\mathcal{T}'}{\mathbb{E}} [\ell(\Psi; \bfP, y)]\leq \epsilon.
\end{equation}

\end{proof}

\subsection{Proof of Theorem \ref{thm: prune}}
\begin{proof}
    We cover the proof in the proof of Proposition \ref{prop: mlp}. Please see Section \ref{subsec: proof prop mlp} for more details.
\end{proof}

\section{Proofs of Key Lemmas and propositions}

\subsection{Proof of Lemma \ref{lemma: mini_task}}
\begin{proof}
 We first show that if $|\mathcal{T}_{tr}|<M_1$, Condition \ref{cond: task} cannot hold. Then, We  show that there exists $\mathcal{T}_{tr}$ with $|\mathcal{T}_{tr}|\geq M_1$ such that Condition \ref{cond: task} holds.

(1) If $|\mathcal{T}_{tr}<M_1|$, then $|\mathcal{T}_{tr}|/M_1<1$, which is contradict to $|\mathcal{T}_{tr}|/M_1\geq 1$ in Condition \ref{cond: task}.

(2) The following example satisfies $|\mathcal{T}_{tr}|\geq M_1$. In this example, the $i$-th task function ($i\in[M_1]$) in $\mathcal{T}_{tr}$ maps the query with $\bfmu_i$ and $\bfmu_{i+1}$ as IDR patterns to $+1$ and $-1$, respectively, where we denote $\bfmu_{M_1+1}:=\bfmu_1$. Hence, the numbers of tasks that map $\bfmu_j$ to $+1$ and $-1$ are both $1$ for any $j\in[M_1]$. In this case, $|\mathcal{T}_{tr}|=M_1$. 

\end{proof}

\subsection{Proof of Proposition \ref{prop: self-attention}}
\begin{proof} We first show the results for in-domain patterns. \\
\noindent (1) We investigate the results about $\bfW_Q$ and then $\bfW_K$. For $\bfp_{query}$ with $\bfmu_j$ as the IDR pattern and $\bfa\in\{\bfmu_1,\cdots,\bfmu_{M_1}\}\backslash\{\bfmu_j\}$, by (\ref{lm_eq: Q mu_j mu_a}), we have
\begin{equation}
\begin{aligned}
    &(\bfa^\top,\boldsymbol{0}^\top)\bfW_Q^{(T)}\bfp_{query}\\
    =& (\bfa^\top,\boldsymbol{0}^\top)(\bfW_Q^{(0)}+\eta\frac{1}{B}\sum_{b=0}^{T-1}\sum_{n\in\mathcal{B}_b}\frac{\partial \ell(\tilde{\bfP}^n,z^n;\Psi)}{\partial \bfW_Q^{(t)}}\Big|_{t=b}\bfp_{query}\\
    \lesssim & \frac{1}{M_1}\cdot (\bfmu_j^\top,\boldsymbol{0}^\top)\eta\frac{1}{B}\sum_{b=0}^{T-1}\sum_{n\in\mathcal{B}_b}\frac{\partial \ell(\tilde{\bfP}^n,z^n;\Psi)}{\partial \bfW_Q^{(t)}}\Big|_{t=b}\bfp_{query},\label{a mu Q T}
\end{aligned}
\end{equation}
if $\bfa$ does not form a task in $\mathcal{T}_{tr}$ with $\bfmu_j$. If $\bfa$ forms a task in $\mathcal{T}_{tr}$ with $\bfmu_j$, and $\eta T=\Theta(M_1\sqrt{\log M_1})$, by (\ref{lm_eq: Q mu_j mu_l})
\begin{equation}
\begin{aligned}
    &(\bfa^\top,\boldsymbol{0}^\top)\bfW_Q^{(T)}\bfp_{query}\\
    \lesssim & \frac{1}{M_1}\cdot (\bfmu_j^\top,\boldsymbol{0}^\top)\eta\frac{1}{B}\sum_{b=0}^{T-1}\sum_{n\in\mathcal{B}_b}\frac{\partial \ell(\tilde{\bfP}^n,z^n;\Psi)}{\partial \bfW_Q^{(t)}}\Big|_{t=b}\bfp_{query}.\label{a mu Q T 2}
\end{aligned}
\end{equation}
For $\bfa\perp\bfmu_j$ but $\bfa\notin\{\bfmu_1,\cdots,\bfmu_{M_1}\}$, by (\ref{lm_eq: Q mu_j nu}), we have
\begin{equation}
\begin{aligned}
    &(\bfa^\top,\boldsymbol{0}^\top)\bfW_Q^{(T)}(\bfmu_j^\top,\boldsymbol{{0}}^\top)^\top\\
    =& (\bfa^\top,\boldsymbol{0}^\top)(\bfW_Q^{(0)}+\eta\frac{1}{B}\sum_{b=0}^{T-1}\sum_{n\in\mathcal{B}_b}\frac{\partial \ell(\tilde{\bfP}^n,z^n;\Psi)}{\partial \bfW_Q^{(t)}}\Big|_{t=b})\bfp_{query}\\
    \lesssim & \frac{1}{M_2}\cdot(\bfmu_j^\top,\boldsymbol{0}^\top)\eta\frac{1}{B}\sum_{b=0}^{T-1}\sum_{n\in\mathcal{B}_b}\frac{\partial \ell(\tilde{\bfP}^n,z^n;\Psi)}{\partial \bfW_Q^{(t)}}\Big|_{t=b}\bfp_{query}.\label{a nu Q T}
\end{aligned}
\end{equation}

By (\ref{lm_eq: Q mu_j mu_j}) and the initialization, we have
\begin{equation}
    (\bfmu_j^\top,\boldsymbol{0}^\top)\bfW_Q^{(T)}\bfp_{query}\geq (\bfmu_j^\top,\boldsymbol{0}^\top)\eta\frac{1}{B}\sum_{b=0}^{T-1}\sum_{n\in\mathcal{B}_b}\frac{\partial \ell(\tilde{\bfP}^n,z^n;\Psi)}{\partial \bfW_Q^{(t)}}\Big|_{t=b}\bfp_{query}\gtrsim\sqrt{\log M_1}+\delta\gtrsim \sqrt{\log M_1},\label{mu mu Q T}
\end{equation}
where the $\sqrt{\log M_1}$ in the second step comes from that $\eta T\geq \Theta
(M_1)\sqrt{\log M_1}$. Hence, by combining (\ref{a mu Q T}), (\ref{a mu Q T 2}), and (\ref{mu mu Q T}), we can derive that
\begin{equation}
\begin{aligned}
    \|\bfW_Q^{(T)}\bfp_{query}^n\|\lesssim &\sqrt{1+\frac{1}{M_1^2}\cdot M_1+\frac{1}{M_1^2}+\frac{1}{M_2^2}\cdot M_2} (\bfmu_j^\top,\boldsymbol{0}^\top)\eta\frac{1}{B}\sum_{b=0}^{T-1}\sum_{n\in\mathcal{B}_b}\frac{\partial \ell(\tilde{\bfP}^n,z^n;\Psi)}{\partial \bfW_Q^{(t)}}\Big|_{t=b}\bfp_{query}\\
    =&\sqrt{1+\frac{1}{M_1}+\frac{1}{M_2}} (\bfmu_j^\top,\boldsymbol{0}^\top)\eta\frac{1}{B}\sum_{b=0}^{T-1}\sum_{n\in\mathcal{B}_b}\frac{\partial \ell(\tilde{\bfP}^n,z^n;\Psi)}{\partial \bfW_Q^{(t)}}\Big|_{t=b}\bfp_{query},
    \end{aligned}
    \label{Q mu norm}
\end{equation}
where in the first step, the first $1/M_1^2\cdot M_1$ comes from (\ref{a mu Q T}) with $M_1-1$ choices of $\bfa$. The second $1/M_1^2$ comes from (\ref{a mu Q T 2}), i.e., $(1/M_1)^2\cdot \Theta(1)$ since there are only a constant number of such cases. The third $1/M_2^2\cdot M_2$ is from (\ref{a nu Q T}) with $M_2$ choices of $\bfa$. %The last $1/M^2$ comes from (\ref{lm_eq: Q mu_j nu other}). 
Therefore, by (\ref{mu mu Q T}) and (\ref{Q mu norm}), for $\bfa\in\{\bfmu_1,\cdots,\bfmu_{M_1}\}\backslash\{\bfmu_j\}$, we have
\begin{equation}
\begin{aligned}
    &(\bfa^\top,\boldsymbol{0}^\top)\bfW_Q^{(T)}\bfp_{query}\lesssim \frac{\sqrt{\log M_1}}{M_1}.
\end{aligned}
\end{equation}
For $\bfa\perp\bfmu_j$ but $\bfa\notin\{\bfmu_1,\cdots,\bfmu_{M_1}\}$,
\begin{equation}
\begin{aligned}
    &(\bfa^\top,\boldsymbol{0}^\top)\bfW_Q^{(T)}\bfp_{query}\lesssim \frac{\sqrt{\log M_1}}{M_2}.
\end{aligned}
\end{equation}

\iffalse
\begin{equation}
\begin{aligned}
    \|\bfW_Q^{(T)}\bfp_{query}^n\|\gtrsim& \eta\frac{1}{B}\sum_{b=0}^{T-1}\sum_{n\in\mathcal{B}_b}\frac{\partial \ell(\tilde{\bfP}^n,z^n;\Psi)}{\partial \bfW_Q^{(t)}}\Big|_{t=b}\bfp_{query}\\
    \gtrsim & \sqrt{\log M_1}\\
    \gtrsim & \sqrt{\log M_1}
\end{aligned}
\end{equation}
\begin{equation}
\begin{aligned}
    S_C(\bfW_{Q}^{(T)}\bfp_{query}^n,(\bfmu_j^\top,\boldsymbol{0}^\top)^\top)=&\frac{(\bfmu_j^\top,\boldsymbol{0}^\top)\bfW_{Q}^{(T)}\bfp_{query}}{\sqrt{\|\bfW_{Q}^{(T)}\bfp_{query}\|}}\\
    \geq & \frac{1}{\sqrt{1+\frac{1}{M_1}+\frac{1}{M_2}}}\\
    \geq & 1-\frac{\Theta(1)}{\min\{M_1,M_2\}},
\end{aligned}
\end{equation}
\fi

%where the second step considers the noise with $$ as the norm in the data, and the last step comes from that $\leq 1/M_1$ and basic computation. \\
\noindent For $\bfW_K$, we can make derivations following the above steps. For $\bfp_q$ with $\bfmu_j$ as the IDR pattern and $\bfa\in\{\bfmu_1,\cdots,\bfmu_{M_1}\}\backslash\{\bfmu_j\}$, by (\ref{lm_eq: K mu_j mu_a}), we have
\begin{equation}
\begin{aligned}
    &(\bfa^\top,\boldsymbol{0}^\top)\bfW_K^{(T)}\bfp_q\\
    \lesssim & \frac{1}{M_1}\cdot (\bfmu_j^\top,\boldsymbol{0}^\top)\eta\frac{1}{B}\sum_{b=0}^{T-1}\sum_{n\in\mathcal{B}_b}\frac{\partial \ell(\tilde{\bfP}^n,z^n;\Psi)}{\partial \bfW_K^{(t)}}\Big|_{t=b}\bfp_q,\label{a mu K T}
\end{aligned}
\end{equation}
if $\bfa$ does not form a task in $\mathcal{T}_{tr}$ with $\bfmu_j$. If $\bfa$ forms a task in $\mathcal{T}_{tr}$ with $\bfmu_j$, and $\eta T=\Theta(M_1\sqrt{\log M_1})$, by (\ref{lm_eq: K mu_j mu_l}),
\begin{equation}
\begin{aligned}
    &(\bfa^\top,\boldsymbol{0}^\top)\bfW_K^{(T)}\bfp_q\\
    \lesssim & \frac{1}{M_1}\cdot (\bfmu_j^\top,\boldsymbol{0}^\top)\eta\frac{1}{B}\sum_{b=0}^{T-1}\sum_{n\in\mathcal{B}_b}\frac{\partial \ell(\tilde{\bfP}^n,z^n;\Psi)}{\partial \bfW_K^{(t)}}\Big|_{t=b}\bfp_q.\label{a mu K T 2}
\end{aligned}
\end{equation}
For $\bfa\perp\bfmu_j$ but $\bfa\notin\{\bfmu_1,\cdots,\bfmu_{M_1}\}$, by (\ref{lm_eq: K mu_j nu}), we have
\begin{equation}
\begin{aligned}
    &(\bfa^\top,\boldsymbol{0}^\top)\bfW_K^{(T)}\bfp_q\\
    \lesssim & \frac{1}{ M_2}\cdot(\bfmu_j^\top,\boldsymbol{0}^\top)\eta\frac{1}{B}\sum_{b=0}^{T-1}\sum_{n\in\mathcal{B}_b}\frac{\partial \ell(\tilde{\bfP}^n,z^n;\Psi)}{\partial \bfW_K^{(t)}}\Big|_{t=b}\bfp_q.\label{a nu K T}
\end{aligned}
\end{equation}

By (\ref{lm_eq: K mu_j mu_j}) and the initialization, we have
\begin{equation}
    (\bfmu_j^\top,\boldsymbol{0}^\top)\bfW_K^{(T)}\bfp_q\geq \sqrt{\log M_1}+\delta\geq \sqrt{\log M_1}.\label{mu mu K T}
\end{equation}
Hence, by combining (\ref{a mu K T}), (\ref{a mu K T 2}), and (\ref{a nu K T}), we can derive that
\begin{equation}
    \|\bfW_K^{(T)}\bfp_i^n\|\lesssim \sqrt{1+\frac{1}{M_1}+\frac{1}{M_1^2}+\frac{1}{M_2}} (\bfmu_j^\top,\boldsymbol{0}^\top)\eta\frac{1}{B}\sum_{b=0}^{T-1}\sum_{n\in\mathcal{B}_b}\frac{\partial \ell(\tilde{\bfP}^n,z^n;\Psi)}{\partial \bfW_K^{(t)}}\Big|_{t=b}(\bfmu_j^\top,\pm\bfq)^\top.\label{K mu norm}
\end{equation}
Therefore, by (\ref{mu mu K T}) and (\ref{K mu norm}), for $\bfa\in\{\bfmu_1,\cdots,\bfmu_{M_1}\}\backslash\{\bfmu_j\}$, we have
\begin{equation}
\begin{aligned}
    &(\bfa^\top,\boldsymbol{0}^\top)\bfW_K^{(T)}\bfp_q\lesssim \frac{\sqrt{\log M_1}}{M_1}.
\end{aligned}
\end{equation}
For $\bfa\perp\bfmu_j$ but $\bfa\notin\{\bfmu_1,\cdots,\bfmu_{M_1}\}$,
\begin{equation}
\begin{aligned}
    &(\bfa^\top,\boldsymbol{0}^\top)\bfW_K^{(T)}\bfp_q\lesssim \frac{\sqrt{\log M_1}}{M_2}.
\end{aligned}
\end{equation}

\iffalse
\begin{equation}
\begin{aligned}
    \|\bfW_K^{(T)}\bfp_q\|\gtrsim \eta\frac{1}{B}\sum_{b=0}^{T-1}\sum_{n\in\mathcal{B}_b}\frac{\partial \ell(\tilde{\bfP}^n,z^n;\Psi)}{\partial \bfW_K^{(t)}}\Big|_{t=b}\bfp_q\gtrsim \sqrt{\log M_1}
\end{aligned}
\end{equation}
\begin{equation}
\begin{aligned}
    S_C(\bfW_{K}^{(T)}\bfp_i^n,(\bfmu_j^\top,\boldsymbol{0}^\top)^\top)
    \geq & 1-\frac{\Theta(1)}{\min\{M_1,M_2\}},
\end{aligned}
\end{equation}
\fi
(2) For out-of-domain patterns, we have the following derivation. Let $\bfmu_j'=\sum_{i=1}^{M_1}k_{ji}\bfmu_i$ where $\sum_{i=1}^{M_1} k_{ji}\geq 1$ and $\sum_{i=1}^{M_1}k_{ji}^2=1$. Then, for a query $\bfp_{query}'$, of which the corresponding ODR pattern is $\bfmu_j'$, we have that by (\ref{lm_eq: Q mu_j mu_j}), (\ref{lm_eq: Q mu_j mu_l}), (\ref{lm_eq: Q mu_j mu_a}), and (\ref{lm_eq: Q mu_j nu}),
\begin{equation}
    \begin{aligned}
        |(\bfmu_j^\top,\boldsymbol{0}^\top)\bfW_Q^{(T)}\bfp_{query}'|\geq |k_j|(\bfmu_j^\top,\boldsymbol{0}^\top)\bfW_Q^{(T)}\bfp_{query}(1-\frac{\Theta(1)}{M_1}-\frac{\Theta(1)}{M_2}),\label{Q mu' low}
    \end{aligned}
\end{equation}
\begin{equation}
    \begin{aligned}
        |(\bfmu_j^\top,\boldsymbol{0}^\top)\bfW_Q^{(t)}\bfp_{query}'|\leq |k_j|(\bfmu_j^\top,\boldsymbol{0}^\top)\bfW_Q^{(t)}\bfp_{query}(1+\frac{\Theta(1)}{M_1}+\frac{\Theta(1)}{M_2}),\label{Q mu' up}
    \end{aligned}
\end{equation}
for any $\bfp_{query}$ with $\bfmu_j$ as the IDR pattern. Meanwhile,
\begin{equation}
    \begin{aligned}
        |(\bfnu_k^\top,\boldsymbol{0}^\top)\bfW_Q^{(t)}\bfp_{query}'|\leq \frac{1}{M_2}(\bfmu_j^\top,\boldsymbol{0}^\top)\bfW_Q^{(t)}\bfp_{query}.
    \end{aligned}
\end{equation}
Therefore,
\begin{equation}
    \|\bfW_Q^{(t)}\bfp_{query}'\|\geq \sqrt{\log M_1}(1-\frac{\Theta(1)}{M_1}-\frac{\Theta(1)}{M_2})\gtrsim \sqrt{\log M_1},
\end{equation}
\begin{equation}
    \|\bfW_Q^{(t)}\bfp_{query}'\|\leq \sqrt{\log M_1}(1+\frac{\Theta(1)}{M_1}+\frac{\Theta(1)}{M_2})\lesssim \sqrt{\log M_1},
\end{equation}
\begin{equation}
    \begin{aligned}
        |({\bfmu_j'}^\top,\boldsymbol{0}^\top)\bfW_Q^{(T)}\bfp_{query}'|\gtrsim \sum_{i=1}^{M_1}|k_{ji}|\sqrt{\log M_1}\geq \sqrt{\log M_1}.
    \end{aligned}
\end{equation}
For $\bfmu_a\in\{\bfmu_1',\cdots,\bfmu_{M_1'}'\}\backslash\{\bfmu_j'\}$, let $\bfmu_a=\sum_{i=1}^{M_1} k_{ai}\bfmu_i$, we have
\begin{equation}
\begin{aligned}
    &(\bfa^\top,\boldsymbol{0}^\top)\bfW_Q^{(T)}\bfp_{query}\lesssim( \frac{1}{M_1}+\frac{1}{M_2})\sum_{i=1}^{M_1} |k_{ai}k_{ji}|\leq \sqrt{\log M_1}(\frac{1}{M_1}+\frac{1}{M_2}),
\end{aligned}
\end{equation}
where the first step is by (\ref{Q mu' low}) and (\ref{Q mu' up}), and the second step is by Cauchy-Schwarz inequality given that $\sum_{i=1}^{M_1} k_{ji}^2=\sum_{i=1} k_{ai}^2=1$. For $\bfa\perp\bfmu_j'$ but $\bfa\notin\{\bfmu_1',\cdots,\bfmu_{M_1'}'\}$,
\begin{equation}
\begin{aligned}
    &(\bfa^\top,\boldsymbol{0}^\top)\bfW_Q^{(T)}\bfp_{query}\lesssim \sqrt{\log M_1}(\frac{1}{M_1}+\frac{1}{M_2}).
\end{aligned}
\end{equation}
Likewise, we can derive the conclusion for the testing context with $\bfW_K^{(T)}$.

\end{proof}

\subsection{Proof of Corollary \ref{cor: attention map}}
\begin{proof}
    From (\ref{attention_T}) to (\ref{gamma_T_one}), we can derive the conclusion for IDR patterns. For ODR patterns, from (\ref{attn_ODR}), we can obtain the conclusion. Note that $\frac{2-\alpha}{\alpha}=\Theta(1)$ since $\alpha=\Theta(1)$.
\end{proof}

\subsection{Proof of Proposition \ref{prop: mlp}}\label{subsec: proof prop mlp}
\begin{proof}
    For any $i$, we denote $\bfW_{O_{(i,\cdot)}}^{(b)}=(\bfO_{i,1}^{(b)},\bfO_{i,2}^{(b)}, \boldsymbol{0}^\top)$ where ${\bfO_{i,1}^{(b)}}^\top\in\mathbb{R}^{d_\mathcal{X}}$ and ${\bfO_{i,2}^{(b)}}^\top\in\mathbb{R}^{d_\mathcal{Y}}$. Following the derivation of (\ref{WOVp_1}), we can obtain that for $s\in[l]$ or $s$ is the query,
\begin{equation}
\begin{aligned}
    &\bfW_{O_{(i,\cdot)}}^{(T)}\bfW_V^{(T)}(\bfp_s^n,\boldsymbol{0}^\top)^\top\\
    = & \bfW_{O_{(i,\cdot)}}^{(T)}(\delta(\bfp_s^n,\boldsymbol{0}^\top)^\top+\sum_{b=0}^{T-1}\eta(\sum_{i\in\mathcal{W}_n}V_i(b)\bfO_{i,1}^{(b)}+\sum_{i\in\mathcal{U}_n}V_i(b)\bfO_{i,1}^{(b)}+\sum_{i\notin\mathcal{W}_n\cup\mathcal{U}_n}V_i(b)\bfO_{i,1}^{(b)})^\top)\\
    \gtrsim & \delta^2\beta^2\frac{\alpha\eta T}{2a}+\delta^2\beta^2\alpha^2 \frac{(\eta T)^3 M_1}{a^2},\label{OV_mu}
\end{aligned}
\end{equation}
for any $j\in[M_1]$, and
\begin{equation}
    \bfW_{O_{(i,\cdot)}}^{(T)}\bfW_V^{(T)}(\kappa_i^n\bfnu_k^\top,\boldsymbol{0}^\top)^\top\leq \frac{1}{M_2}\cdot\bfW_{O_{(i,\cdot)}}^{(T)}\bfW_V^{(T)}(\bfmu_j^\top,\boldsymbol{0}^\top)^\top,\label{OV_nu}
\end{equation}
for $k\in[M_2]$ by (\ref{V'_v}) and (\ref{WO_v}) in Lemma \ref{lemma: V} and \ref{lemma: O}, respectively. Then, we have
\begin{equation}
\begin{aligned}
    \frac{(\frac{1}{M_1}\sum_{j=1}^{M_1}\bfmu_j^\top,\boldsymbol{0}^\top)\bfW_{O_{(i,\cdot)}}^{(T)}\bfW_V^{(T)}}{\|(\frac{1}{M_1}\sum_{j=1}^{M_1}\bfmu_j^\top,\boldsymbol{0}^\top)\|\|\bfW_{O_{(i,\cdot)}}^{(T)}\bfW_V^{(T)}\|}
    \geq & \frac{1}{\sqrt{1+\frac{1}{M_2^2}\cdot M_2}}\\
    \geq & 1-\frac{\Theta(1)}{M_2},
\end{aligned}
\end{equation}
because $BT\geq \Theta(M_1^2)$. For any $i\in\mathcal{W}$,
\begin{equation}
\begin{aligned}
    &\bfW_{O_{(i,\cdot)}}^{(T)}\bfW_V^{(T)}(\boldsymbol{0}^\top,\bfq^\top)^\top\\
    = & \bfW_{O_{(i,\cdot)}}^{(T)}(\delta(\boldsymbol{0}^\top,\bfq^\top)^\top+\sum_{b=0}^{T-1}\eta(\sum_{i\in\mathcal{W}_n}V_i(b)\bfO_{i,2}^{(b)}+\sum_{i\in\mathcal{U}_n}V_i(b)\bfO_{i,2}^{(b)}+\sum_{i\notin\mathcal{W}_n\cup\mathcal{U}_n}V_i(b)\bfO_{i,2}^{(b)})^\top)\\
    \gtrsim & \delta^2\frac{\alpha\eta T}{2a}+\delta^2\alpha^2 \frac{(\eta T)^3 M_1}{a^2}.\label{OV_q}
\end{aligned}
\end{equation}
Note that by gradient updates of $\bfW_O$ and $\bfW_V$, there are no gradient components perpendicular to $\bfp$ except some Gaussian noise. Hence,
\begin{equation}
\begin{aligned}
    \frac{(\boldsymbol{0}^\top, \bfq^\top)\bfW_{O_{(i,\cdot)}}^{(T)}\bfW_V^{(T)}}{\|(\boldsymbol{0}^\top, \bfq^\top)\|\|\bfW_{O_{(i,\cdot)}}^{(T)}\bfW_V^{(T)}\|}
    \geq & \frac{1}{\sqrt{1+\xi}}\\
    \geq & 1-\frac{\Theta(1)}{M_1}.\label{q angle pos}
\end{aligned}
\end{equation}
For any $i\in\mathcal{U}$,
\begin{equation}
\begin{aligned}
    &\bfW_{O_{(i,\cdot)}}^{(T)}\bfW_V^{(T)}(\boldsymbol{0}^\top,-\bfq^\top)^\top\\
    = & \bfW_{O_{(i,\cdot)}}^{(T)}(\delta(\boldsymbol{0}^\top,-\bfq^\top)^\top+\sum_{b=0}^{T-1}\eta(\sum_{i\in\mathcal{W}_n}V_i(b)\bfO_{i,2}^{(b)}+\sum_{i\in\mathcal{U}_n}V_i(b)\bfO_{i,2}^{(b)}+\sum_{i\notin\mathcal{W}_n\cup\mathcal{U}_n}V_i(b)\bfO_{i,2}^{(b)})^\top)\\
    \gtrsim & \delta^2\frac{\alpha\eta T}{2a}+\delta^2\alpha^2 \frac{(\eta T)^3 M_1}{a^2}.\label{OV_-q}
\end{aligned}
\end{equation}
Similarly to (\ref{q angle pos}), we have
\begin{equation}
\begin{aligned}
    \frac{(\boldsymbol{0}^\top, -\bfq^\top)\bfW_{O_{(i,\cdot)}}^{(T)}\bfW_V^{(T)}}{\|(\boldsymbol{0}^\top, -\bfq^\top)\|\|\bfW_{O_{(i,\cdot)}}^{(T)}\bfW_V^{(T)}\|}
    \geq & 1-\frac{\Theta(1)}{M_1}.
\end{aligned}
\end{equation}
Hence, for $i\in\mathcal{W}\cup\mathcal{U}$, 
\begin{equation}
    \|\bfW_{O_{(i,\cdot)}}^{(T)}\bfW_V^{(T)}\|\gtrsim \beta^{-1}=\Omega(1).
\end{equation}
By (\ref{notin2}), we have that for $i\notin\mathcal{W}\cup\mathcal{U}$,
\begin{equation}
    \|\bfW_{O_{(i,\cdot)}}^{(T)}\bfW_V^{(T)}\|\lesssim \sqrt{\frac{1}{M_1}+\frac{1}{M_2^2}\cdot M_2}=\frac{1}{\sqrt{M_2}},
\end{equation}
where $1/M_1$ is the square of (\ref{notin2}). $1/M_2^2$ is the square of the scaling in RHS of (\ref{OV_nu}), and $M_2$ is the number of IDI patterns. \\

\noindent If we prune all neurons $i\notin\mathcal{W}\cup\mathcal{U}$, we have that 
\begin{equation}
\begin{aligned}
    \underset{\bfx_{query}, f}{\mathbb{E}} [\ell(\Psi; \bfP, y)]\leq &1-(1-\frac{2}{\alpha'M_1})\frac{1-\epsilon}{(1-\frac{2}{\alpha M_1})}(1-\frac{1}{\sqrt{M_1}})\\
    \leq & 1-(1-\frac{2}{\alpha' M_1})(1-\epsilon)(1-\frac{1}{\sqrt{M_1}})\\
    \leq & 1-(1-\frac{2}{\alpha' M_1}-\epsilon-\frac{1}{\sqrt{M_1}})\\
    \leq & \mathcal{O}(\epsilon+\frac{1}{\sqrt{M_1}}+\frac{1}{\alpha' M_1})\\
    \leq & \mathcal{O}(\epsilon+\frac{1}{\sqrt{M_1}}),
\end{aligned}
\end{equation}
where the first step combines (\ref{F1_lower}), (\ref{notin2}), and $2/(\alpha M_1)$ and $2/(\alpha' M_1)$ comes from (\ref{attn_train}) and (\ref{attn_ODR}). The last step comes from $\alpha'\geq M_1^{-1/2}$. Meanwhile, if we prune $R$ fraction of neurons in $\mathcal{W}\cup\mathcal{U}$, given (\ref{grad_noise}), we have for the trained model $\Psi$, 
\begin{equation}
    F(\Psi; \bfP^n)\leq (1+\epsilon)(1-R)\cdot \frac{(1-\frac{2}{\alpha'M_1})}{(1-\frac{2}{\alpha M_1})}.
\end{equation}
Then,
\begin{equation}
\begin{aligned}
    \underset{\bfx_{query}, f}{\mathbb{E}} [\ell(\Psi; \bfP, y)]\geq &1-(1-\frac{2}{\alpha'M_1})\frac{1+\epsilon}{(1-\frac{2}{\alpha M_1})}(1-R)\\
    \geq & 1-(1-\frac{2}{\alpha'M_1})(1+\epsilon)(1+\frac{4}{\alpha M_1})(1-R)\\
    = & 1-(1-R-\frac{2}{\alpha' M_1}+\frac{2R}{\alpha' M_1})(1+\epsilon+\frac{4}{\alpha M_1}+\frac{4\epsilon}{\alpha M_1})\\
    \geq & R+\frac{2}{\alpha' M_1}-\frac{2R}{\alpha' M_1}-(1-R-\frac{2}{\alpha' M_1})(\epsilon+\frac{4+4\epsilon}{\alpha M_1})\\
    \geq & \Omega(R+\frac{1}{\alpha' M_1}),
\end{aligned}
\end{equation}
where the second step is by $(1-x)^{-1}\leq 1+2x$ for small $x>0$, and the last step is by $R=\Theta(1)$.

\end{proof}

\subsection{Proof of Lemma \ref{lemma: QK}}
\begin{proof}
\noindent We first study the gradient of $\bfW_Q^{(t+1)}$ in part (a) and the gradient of $\bfW_K^{(t+1)}$ in part (b). The proof is derived with a framework of induction combined with Lemma \ref{lemma: V} and \ref{lemma: O}. \\
\noindent (a) From the training loss function, by basic mathematical computation, we can obtain
\begin{equation}
\begin{aligned}
    &\quad\eta\frac{1}{B}\sum_{n\in\mathcal{B}_b}\frac{\partial \ell(\tilde{\bfP}^n,z^n;\Psi)}{\partial \bfW_Q}\\
    &=\eta\frac{1}{B}\sum_{n\in\mathcal{B}_b} \frac{\partial \ell(\tilde{\bfP}^n,z^n;\Psi)}{\partial F(\bfp_{query}^n)}\frac{\partial F(\bfp_{query}^n)}{\partial \bfW_Q}\\
    &=\eta\frac{1}{B}\sum_{n\in\mathcal{B}_b}(-z^n) \sum_{i=1}^m a_{i}\mathbbm{1}[\bfW_{O_{(i,\cdot)}}\sum_{s=1}^{l+1}(\bfW_V\bfp_s^n )\text{softmax}({\bfp_s^n}^\top\bfW_K^\top\bfW_Q\bfp_{query}^n )\geq0]\\
    &\quad\cdot\Big(\bfW_{O_{(i,\cdot)}}\sum_{s=1}^{l+1} (\bfW_V\bfp_s^n )\text{softmax}({\bfp_s^n}^\top\bfW_K^\top\bfW_Q\bfp_{query}^n )\\
    &\quad\cdot\sum_{r=1}^{l+1} \text{softmax}({\bfp_r^n}^\top\bfW_K^\top\bfW_Q\bfp_{query}^n )\bfW_K(\bfp_s^n-\bfp_r^n){\bfp_{query}^n}^\top\Big)\\ 
    &=\eta\frac{1}{B}\sum_{n\in\mathcal{B}_b}(-z^n) \sum_{i=1}^m a_{i}\mathbbm{1}[\bfW_{O_{(i,\cdot)}}\sum_{s=1}^{l+1}(\bfW_V\bfp_s^n )\text{softmax}({\bfp_s^n}^\top\bfW_K^\top\bfW_Q\bfp_{query}^n )\geq0]\\
    &\quad\cdot\Big(\bfW_{O_{(i,\cdot)}}\sum_{s=1}^{l+1} (\bfW_V\bfp_s^n )\text{softmax}({\bfp_s^n}^\top\bfW_K^\top\bfW_Q\bfp_{query}^n )\\
    &\quad\cdot(\bfW_K\bfp_s^n-\sum_{r=1}^{l+1} \text{softmax}({\bfp_r^n}^\top\bfW_K^\top\bfW_Q\bfp_{query}^n )\bfW_K\bfp_r^n){\bfp_{query}}^\top\Big)\label{grad_Wk}.
\end{aligned}
\end{equation}
If $t=0$, we have that 
\begin{equation}   ({\bfW_K^{(t)}}\bfp_s^n)^\top(\bfW_Q^{(t)}\bfp_{query}^n )={\bfp_s^n}^\top{\bfW_K^{(t)}}^\top\bfW_Q^{(t)}\bfp_{query}^n.
\end{equation}
When $z^n=+1$, let $\bfx_{query}^n$ be a noisy version of $\bfmu_j+\kappa_{query}^n\bfnu_k$ where $j\in\{1,2,\cdots, M_1\}$ and $k\in\{1,2,\cdots, M_2\}$. Define $m_i$ as the corresponding IDR pattern in the $i$-th demonstration. Consider the categorical distribution where the probability of selecting $\bfmu_q$ is $\alpha/2$. We know there exists a $\bfmu_j$ such that the probability of selecting $\bfmu_j$ is also $\alpha/2$. Selecting other $\bfmu_t$ for $t\neq p,j$ has a probability of $(1-\alpha)/(M_1-2)$. Selecting any IDI pattern $\bfnu_k$ has a probability of $1/M_2$. By the Chernoff bound of Bernoulli distribution in Lemma \ref{lemma: chernoff}, given $l\geq \Theta(\max\{M_1 , M_2\}\log M_1)$, we can obtain
\begin{equation}
    \Pr\left(\sum_{i=1}^l\mathbbm{1}[m_i=\bfmu_j] \leq l \cdot \frac{\alpha}{2}\right)\leq e^{-C\log M_1}= M_1^{-C},
\end{equation}
\begin{equation}
    \Pr\left(\sum_{i=1}^l\mathbbm{1}[m_i=\bfmu_s] \leq l \cdot \frac{\alpha}{2}\right)\leq e^{-C\log M_1}= M_1^{-C},
\end{equation}
\begin{equation}
    \Pr\left(\sum_{i=1}^l\mathbbm{1}[m_i=\bfmu_t] \geq l \cdot \frac{1}{M_1}\right)\leq e^{-C\log M_1 \cdot M_1\cdot \frac{1}{M_1}}= M_1^{-C},
\end{equation}
\begin{equation}
    \Pr\left(\sum_{i=1}^l\mathbbm{1}[m_i=\bfnu_k] \geq l \cdot \frac{1}{M_2}\right)\leq e^{-C\log M_1 \cdot M_2\cdot \frac{1}{M_2}}= M_1^{-C},\label{prob_kappa}
\end{equation}
for some $C>0$. Therefore, since that $\frac{1}{\sqrt{M_2}}\cdot e^{\delta^2 }\lesssim \frac{\alpha}{2}=\Theta(1)$,
\begin{equation}
   \begin{aligned}\sum_{s\in\mathcal{N}^{n,i}_j\cap\mathcal{M}^{n,i}_k}e^{\delta^2 (\beta\cdot\beta+1)}&\leq l\cdot \frac{1}{\sqrt{M_2}}e^{\delta^2 }\cdot e^{\delta^2 (\beta\cdot\beta)}\\
   &\lesssim l\cdot \frac{\alpha}{2}\cdot e^{\delta^2 (\beta\cdot\beta)}\\
   &\lesssim \sum_{s\in\mathcal{N}^{n,i}_j-\mathcal{M}^{n,i}_k}e^{\delta^2 (\beta\cdot\beta)}\label{MkNj1}.
   \end{aligned}
\end{equation}
Similarly, 
\begin{equation}
   \begin{aligned}\sum_{s\in\mathcal{M}^{n,i}_k-\mathcal{N}^{n,i}_j}e^{\delta^2 }&\leq l\cdot \frac{1}{\sqrt{M_2}}e^{\delta^2 }\cdot e^{\delta^2 (\beta\cdot\beta)}\\
   &\lesssim l\cdot \frac{\alpha}{2}\cdot e^{\delta^2 (\beta\cdot\beta)}\\
   &\lesssim \sum_{s\in[l]-\mathcal{N}^{n,i}_j-\mathcal{N}^{n,i}_k}e^{\delta^2 },\label{MkNj2}
   \end{aligned}
\end{equation}
where the last step is by the fact that there exists $\bfmu_s$ for $p\in\{1,2,\cdots,M_2\}\backslash\{j\}$ such that selecting $\bfmu_s$ has a probability of $\alpha/2$. 
Let $i\in\mathcal{W}$, $s\in\mathcal{N}_j^{n,i}-\mathcal{M}_k^{n,i}$, then
\begin{equation}
\begin{aligned}
    &\text{softmax}({\bfp_s^n}^\top{\bfW_K^{(t)}}^\top\bfW_Q^{(t)}\bfp_{query}^n)\\
    \geq &e^{\delta^2 (\beta\cdot\beta)}\cdot(\sum_{s\in\mathcal{N}^{n,i}_j-\mathcal{M}^{n,i}_k}e^{\delta^2 (\beta\cdot\beta)}+\sum_{s\in\mathcal{N}^{n,i}_j\cap\mathcal{M}^{n,i}_k}e^{\delta^2 (\beta\cdot\beta+1)}\\
    &+\sum_{s\in[l]-\mathcal{N}^{n,i}_j-\mathcal{M}^{n,i}_k}e^{\delta^2 }+\sum_{s\in\mathcal{M}^{n,i}_k-\mathcal{N}^{n,i}_j}e^{\delta^2 })^{-1}\\
    \gtrsim & \frac{e^{\delta^2 (\beta\cdot\beta)}}{\sum_{s\in\mathcal{N}^{n,i}_j-\mathcal{M}^{n,i}_k}e^{\delta^2 (\beta\cdot\beta)}+\sum_{s\in[l]-\mathcal{N}^{n,i}_j-\mathcal{N}^{n,i}_k}e^{\delta^2 }},\label{softmax_in}
\end{aligned}
\end{equation}
where the second step is by (\ref{MkNj1}) and (\ref{MkNj2}). Similarly, for $s\in\mathcal{N}^{n,i}_j\cap\mathcal{M}^{n,i}_k$, 
\begin{equation}
\begin{aligned}
    &\text{softmax}({\bfp_s^n}^\top{\bfW_K^{(t)}}^\top\bfW_Q^{(t)}\bfp_{query}^n)\\
    \gtrsim & \frac{e^{\delta^2 (\beta\cdot\beta)}}{\sum_{s\in\mathcal{N}^{n,i}_j-\mathcal{M}^{n,i}_k}e^{\delta^2 (\beta\cdot\beta)}+\sum_{s\in[l]-\mathcal{N}^{n,i}_j-\mathcal{N}^{n,i}_k}e^{\delta^2 }}.
\end{aligned}
\end{equation}
For $s\in\mathcal{M}^{n,i}_k-\mathcal{N}_j^{n,i}$,
\begin{equation}
\begin{aligned}
    &\text{softmax}({\bfp_s^n}^\top{\bfW_K^{(t)}}^\top\bfW_Q^{(t)}\bfp_{query}^n)\\
    \lesssim & \frac{e^{\delta^2 }}{\sum_{s\in\mathcal{N}^{n,i}_j-\mathcal{M}^{n,i}_k}e^{\delta^2 (\beta\cdot\beta)}+\sum_{s\in[l]-\mathcal{N}^{n,i}_j-\mathcal{N}^{n,i}_k}e^{\delta^2 }}\label{softmax_out}.
\end{aligned}
\end{equation}
For $s\in[l]-\mathcal{N}_j^{n,i}-\mathcal{M}_k^{n,i}$,
\begin{equation}
\begin{aligned}
    &\text{softmax}({\bfp_s^n}^\top{\bfW_K^{(t)}}^\top\bfW_Q^{(t)}\bfp_{query}^n)\\
    \lesssim & \frac{e^{\delta^2 }}{\sum_{s\in\mathcal{N}^{n,i}_j-\mathcal{M}^{n,i}_k}e^{\delta^2 (\beta\cdot\beta)}+\sum_{s\in[l]-\mathcal{N}^{n,i}_j-\mathcal{N}^{n,i}_k}e^{\delta^2 }}.
\end{aligned}
\end{equation}
By (\ref{W_def}) and (\ref{U_def}) in Definition \ref{def: WU}, we have that for $i\in\mathcal{W}_n$, 
\begin{equation}
    \bfW_{O_{(i,\cdot)}}^{(t)}\sum_{s=1}^{l+1}(\bfW_V^{(t)} {\bfp_s^n})\text{softmax}({\bfp_s^n}^\top{\bfW_K^{(t)}}^\top\bfW_Q^{(t)}\bfp_{query}^n)>0.
\end{equation}
\noindent Then we derive
\begin{equation}
\begin{aligned}
    &\bfW_K^{(t)}\bfp_s^n-\sum_{r=1}^{l+1} \text{softmax}({\bfp_r^n}^\top{\bfW_K^{(t)}}^\top\bfW_Q^{(t)}\bfp_{query}^n)\bfW_K^{(t)}\bfp_r^n\\
    =& \sum_{r=1}^{l+1}\text{softmax}({\bfp_r^n}^\top{\bfW_K^{(t)}}^\top\bfW_Q^{(t)}\bfp_{query}^n)(\bfW_K^{(t)}\bfp_s^n-\bfW_K^{(t)}\bfp_r^n)\\
    =& \Big(\sum_{r\in\mathcal{N}_j^{n,i}-\mathcal{M}_k^{n,i}}+\sum_{r\in\mathcal{N}_j^{n,i}\cap\mathcal{M}_k^{n,i}}+\sum_{r\in\mathcal{M}_k^{n,i}-\mathcal{N}_j^{n,i}}+\sum_{r\in[l]-\mathcal{N}_j^{n,i}-\mathcal{M}_k^{n,i}}\Big)\text{softmax}({\bfp_r^n}^\top{\bfW_K^{(t)}}^\top\bfW_Q^{(t)}\bfp_{query}^n)\\
    &\cdot(\bfW_K^{(t)}\bfp_s^n-\bfW_K^{(t)}\bfp_r^n).
\end{aligned}
\end{equation}
One can observe that 
\begin{equation}
    \begin{aligned}
    &\sum_{s\in\mathcal{N}_j^{n,i}}\text{softmax}({\bfp_s^n}^\top{\bfW_K^{(t)}}^\top\bfW_Q^{(t)}\bfp_{query}^n)(\bfW_K^{(t)}\bfp_s^n-\sum_{r=1}^{l+1} \text{softmax}({\bfp_r^n}^\top{\bfW_K^{(t)}}^\top\bfW_Q^{(t)}\bfp_{query}^n)\bfW_K^{(t)}\bfp_r^n)\\
    =&\sum_{s\in\mathcal{N}_j^{n,i}}\text{softmax}({\bfp_s^n}^\top{\bfW_K^{(t)}}^\top\bfW_Q^{(t)}\bfp_{query}^n)(\bfW_K^{(t)}\bfp_s^n-(\sum_{r\in\mathcal{N}_j^{n,i}}+\sum_{r\notin\mathcal{N}_j^{n,i}}) \text{softmax}({\bfp_r^n}^\top{\bfW_K^{(t)}}^\top\bfW_Q^{(t)}\bfp_{query}^n)\bfW_K^{(t)}\bfp_r^n)\\
    =& \sum_{r\notin\mathcal{N}_j^{n,i}}\text{softmax}({\bfp_r^n}^\top{\bfW_K^{(t)}}^\top\bfW_Q^{(t)}\bfp_{query}^n)\cdot \sum_{s\in\mathcal{N}_j^{n,i}}\text{softmax}({\bfp_s^n}^\top{\bfW_K^{(t)}}^\top\bfW_Q^{(t)}\bfp_{query}^n)\bfW_K^{(t)}\bfp_s^n\\
    &-\sum_{s\in\mathcal{N}_j^{n,i}}\text{softmax}({\bfp_s^n}^\top{\bfW_K^{(t)}}^\top\bfW_Q^{(t)}\bfp_{query}^n)\cdot \sum_{r\notin\mathcal{N}_j^{n,i}}\text{softmax}({\bfp_r^n}^\top{\bfW_K^{(t)}}^\top\bfW_Q^{(t)}\bfp_{query}^n)\bfW_K^{(t)}\bfp_r^n+\bfn\\
    =&\sum_{s\in\mathcal{N}_j^{n,i}}\text{softmax}({\bfp_s^n}^\top{\bfW_K^{(t)}}^\top\bfW_Q^{(t)}\bfp_{query}^n)\cdot \sum_{r\notin\mathcal{N}_j^{n,i}}\text{softmax}({\bfp_r^n}^\top{\bfW_K^{(t)}}^\top\bfW_Q^{(t)}\bfp_{query}^n)\bfW_K^{(t)}(\bfp_s^n-\bfp_r^n)+\bfn.
    \end{aligned}
\end{equation}
 Hence, by Definition \ref{def: zeta},
\begin{equation}
1>\sum_{r\in[l]-\mathcal{N}^{n,i}_j}\text{softmax}({\bfp_s^n}^\top{\bfW_K^{(t)}}^\top\bfW_Q^{(t)}\bfp_{query}^n)\geq \gamma_t>0.
\end{equation}
Since that the feature space embedding of $({\bfp_r^n}^\top, \boldsymbol{0}^\top)^\top$ are orthogonal to $\bfW_Q^{(t)}\bfp_{query}^n$ for $r\in[l]-\mathcal{N}^{n,i}_j$, we have that with high probability, for $s\in\mathcal{N}^{n,i}_j$, 
\begin{equation}
\begin{aligned}
    &\quad({\bfx_s^n}^\top, \boldsymbol{0}^\top)\sum_{r\in[l]-\mathcal{N}^{n,i}_j}\text{softmax}({\bfp_r^n}^\top{\bfW_K^{(t)}}^\top\bfW_Q^{(t)}\bfp_{query}^n)\cdot(\bfW_K^{(t)}\bfp_s^n-\bfW_K^{(t)}\bfp_r^n)\\
    &\geq \gamma_t\beta^2 \delta,
\end{aligned}
\end{equation}
where $\gamma_t$ comes from the definition. $\beta$ is from the definition of the data. Meanwhile, for $r$ such that $\bfmu_r$ is the IDR pattern with the probability of $\alpha/2$ to be selected,
\begin{equation}
\begin{aligned}
    &\Big|({\bfx_r^n}^\top, \boldsymbol{0}^\top)\sum_{r}\text{softmax}({\bfp_r^n}^\top{\bfW_K^{(t)}}^\top\bfW_Q^{(t)}\bfp_{query}^n)\cdot (\bfW_K^{(t)}\bfp_s^n-\bfW_K^{(t)}\bfp_r^n)\Big|\\
    \leq &({\bfx_s^n}^\top, \boldsymbol{0}^\top)\sum_{r\in[l]-\mathcal{N}^{n,i}_j}\text{softmax}({\bfp_r^n}^\top{\bfW_K^{(t)}}^\top\bfW_Q^{(t)}\bfp_{query}^n)\cdot(\bfW_K^{(t)}\bfp_s^n-\bfW_K^{(t)}\bfp_r^n)\cdot\frac{1-\alpha}{1-\alpha/2},
\end{aligned}
\end{equation}
where $(1-\alpha)/(1-\alpha/2)$ comes from the fraction of attention weights on $\bfmu_r$ in $[l]-\mathcal{N}_j^{n,i}$. If $\bfmu_r$ is the pattern that does not decide the label of the current $\bfP^n$, we have
\begin{equation}
\begin{aligned}
    &\Big|({\bfx_r^n}^\top, \boldsymbol{0}^\top)\sum_{r}\text{softmax}({\bfp_r^n}^\top{\bfW_K^{(t)}}^\top\bfW_Q^{(t)}\bfp_{query}^n)\cdot (\bfW_K^{(t)}\bfp_s^n-\bfW_K^{(t)}\bfp_r^n)\Big|\\
    \leq &\frac{\gamma_t }{l},
\end{aligned}
\end{equation}
where $l$ in the denominator comes from that with high probability, at most $1$ $\bfmu_r$ appears in one data for a certain $r$. Therefore, for $i\in\mathcal{W}_n$, we denote that $\zeta_{i,n}'=\bfW_{O_{(i,\cdot)}}^{(t)}\sum_{s\in\mathcal{N}_j^{n,i}} (\bfW_V^{(t)} {\bfp_s^n}^{(t)})\text{softmax}({\bfp_s^n}^\top{\bfW_K^{(t)}}^\top\bfW_Q^{(t)}\bfp_{query}^n)$. Then, if $\zeta_{i,n}'>0$, we have that for $\bfmu_q$ that has the same IDR pattern as $\bfx_{query}$, 
\begin{equation}
\begin{aligned}
    &(\bfmu_q^\top, \boldsymbol{{0}^\top})\bfW_{O_{(i,\cdot)}}^{(t)}\sum_{s\in\mathcal{N}_j^{n,i}} (\bfW_V^{(t)} {\bfp_s^n}^{(t)})\text{softmax}({\bfp_s^n}^\top{\bfW_K^{(t)}}^\top\bfW_Q^{(t)}\bfp_{query}^n)\\
    &\cdot(\bfW_K^{(t)}\bfp_s^n-\sum_{r=1}^{l+1} \text{softmax}({\bfp_r^n}^\top{\bfW_K^{(t)}}^\top\bfW_Q^{(t)}\bfp_{query}^n)\bfW_K^{(t)}\bfp_r^n){\bfp_{query}^n}^\top(\bfx_{query}^\top, \boldsymbol{{0}^\top})^\top\\
    \geq &\zeta_{i,n}'\delta \beta^4\gamma_t ,\\
\end{aligned}
\end{equation}
where $\gamma_t$ comes from that $\bfW_{O_{(i,\cdot)}}^{(t)}\bfp_s^n$ is much larger in average, $i\in\mathcal{W}_n$, if than other $\bfW_{O_{(i,\cdot)}}^{(t)}\bfp_t^n$ if $s\in\mathcal{N}_*^n$ while $s\notin\mathcal{N}_*^n$. 
For $j$ such that $\bfmu_j$ is the IDR pattern with the probability of $\alpha/2$ to be selected but different from $q$, 
\begin{equation}
\begin{aligned}
    &(\bfmu_j^\top, \boldsymbol{{0}^\top})\bfW_{O_{(i,\cdot)}}^{(t)}\sum_{s\in\mathcal{N}^{n,i}_j} (\bfW_V^{(t)}\bfp_s^n)\text{softmax}({\bfp_s^n}^\top{\bfW_K^{(t)}}^\top\bfW_Q^{(t)}\bfp_{query}^n)\\
    &\cdot(\bfW_K^{(t)}\bfp_s^n-\sum_{r=1}^{l+1} \text{softmax}({\bfp_r^n}^\top{\bfW_K^{(t)}}^\top\bfW_Q^{(t)}\bfp_{query}^n)\bfW_K^{(t)}\bfp_r^n){\bfp_{query}}^\top(\bfx_{query}^\top, \boldsymbol{{0}^\top})^\top\\
    \leq &\frac{1-\alpha}{1-\alpha/2}(\bfx_{query}^\top, \boldsymbol{{0}^\top})\bfW_{O_{(i,\cdot)}}^{(t)}\sum_{s\in\mathcal{N}_j^{n,i}} (\bfW_V^{(t)} {\bfp_s^n}^{(t)})\text{softmax}({\bfp_s^n}^\top{\bfW_K^{(t)}}^\top\bfW_Q^{(t)}\bfp_{query}^n).\\
\end{aligned}
\end{equation}
For $\bfmu_j$ that does not decide the label of the current $\bfP^n$, we have
\begin{equation}
\begin{aligned}
    &(\bfmu_j^\top, \boldsymbol{{0}^\top})\bfW_{O_{(i,\cdot)}}^{(t)}\sum_{s\in\mathcal{N}^{n,i}_j} (\bfW_V^{(t)}\bfp_s^n)\text{softmax}({\bfp_s^n}^\top{\bfW_K^{(t)}}^\top\bfW_Q^{(t)}\bfp_{query}^n)\\
    &\cdot(\bfW_K^{(t)}\bfp_s^n-\sum_{r=1}^{l+1} \text{softmax}({\bfp_r^n}^\top{\bfW_K^{(t)}}^\top\bfW_Q^{(t)}\bfp_{query}^n)\bfW_K^{(t)}\bfp_r^n){\bfp_{query}^n}^\top(\bfx_{query}^\top, \boldsymbol{{0}^\top})^\top\\
    \leq &\frac{\zeta_{i,n}'\delta \gamma_t \beta^4}{l}.\\
\end{aligned}
\end{equation}
To deal with $s\in[l]-\mathcal{N}_j^{n,i}$, we cover this part when summing up all the neurons. Since that each entry of $\bfW_{O_{(i,\cdot)}}$ follows $\mathcal{N}(0,\xi^2)$, we have
\begin{equation}
    \Pr(\|\bfW_{O_{(i,1:d_\mathcal{X})}}\bfx_{query}^n\|\leq \beta\xi)\leq \beta\xi,\label{hoef_xi1}
\end{equation}
by the standard property of Gaussian distribution. Meanwhile, by Hoeffding's inequality (\ref{hoeffding}), 
\begin{equation}
    \Pr(\|\bfW_{O_{(i,1:d_\mathcal{X})}}\bfx_{query}^n\|\geq \beta\xi\log M_1)\leq M_1^{-C},\label{hoef_xi2}
\end{equation}
for some $C>1$. Hence, with a high probability, by Hoeffding's inequality (\ref{hoeffding}),
\begin{equation}
    \begin{aligned}
        \Big|\frac{1}{m}\sum_{i\in\mathcal{W}_n}\bfW_{O_{(i,1:d_\mathcal{X})}}^{(t)}\bfx_{query}^n\Big|
        \lesssim  &\frac{|\mathcal{W}_n|}{m}\Phi(0)\beta\xi+\beta\xi\cdot\frac{\log M_1}{\sqrt{m}}\\
        \lesssim &\beta\xi,\label{mean_Wmu_up}
    \end{aligned}
\end{equation}
\begin{equation}
    \begin{aligned}
        \Big|\frac{1}{m}\sum_{i\in\mathcal{W}_n}\bfW_{O_{(i,1:d_\mathcal{X})}}^{(t)}\bfx_{query}^n\Big|
        \gtrsim  &\frac{|\mathcal{W}_n|}{m}\Phi(0)\beta\xi-\beta\xi\cdot\frac{\log M_1}{\sqrt{m}}\\
        \gtrsim &\beta\xi.\label{mean_Wmu_low}
    \end{aligned}
\end{equation}
For $p$ such that the probability of selecting $\bfmu_p$ is $\alpha/2$, we have
\begin{equation}
    \begin{aligned}
        \Big|\frac{1}{m}\sum_{i\in\mathcal{W}_n}\bfW_{O_{(i,1:d_\mathcal{X})}}^{(t)}\bfmu_p\Big|
        \lesssim  \Big|\frac{1}{m}\sum_{i\in\mathcal{W}_n}\bfW_{O_{(i,1:d_\mathcal{X})}}^{(t)}\bfmu_j\Big|\cdot e^{-\delta \beta^2}.
    \end{aligned}
\end{equation}
We have that for $z^n=1$,
\iffalse
\begin{equation}
    |\bfW_{O_{(i,d_\mathcal{X}+1:d_\mathcal{X}+d_\mathcal{Y})}}\bfq|>|-\bfW_{O_{(i,d_\mathcal{X}+1:d_\mathcal{X}+d_\mathcal{Y})}}\bfq|
\end{equation}
Then, 
\fi
we can then derive that for $s\in\mathcal{N}_{j}^{n,i}$, by Definition \ref{def: zeta}, for $\bfmu_q$ which is the IDR pattern of $\bfx_{query}$, 
\begin{equation}
\begin{aligned}
    &(\bfmu_q^\top, \boldsymbol{{0}^\top})\frac{1}{m}\sum_{i\in\mathcal{W}_n}\bfW_{O_{(i,\cdot)}}^{(t)}\sum_{s=1}^{l+1} (\bfW_V^{(t)}\bfp_s^n)\text{softmax}({\bfp_s^n}^\top{\bfW_K^{(t)}}^\top\bfW_Q^{(t)}\bfp_{query}^n)\\
    &\cdot(\bfW_K^{(t)}\bfp_s^n-\sum_{r=1}^{l+1} \text{softmax}({\bfp_r^n}^\top{\bfW_K^{(t)}}^\top\bfW_Q^{(t)}\bfp_{query}^n)\bfW_K^{(t)}\bfp_r^n){\bfp_{query}^n}^\top(\bfx_{query}^\top, \boldsymbol{{0}^\top})^\top\\
    \geq &\zeta_{i,t}\delta \gamma_t (1-e^{-\delta^2 \beta^2})\beta^4,\\\label{lower_l}
\end{aligned}
\end{equation}
where $\zeta_{i,t}$ is used as a lower bound after taking an average of $i\in\mathcal{W}_n$. 
%Note that here for the computation of $\bfy_s^n$ space, we consider the majority voting, which enables us to only focus on $y_s^n=z^n$ for $s\in\mathcal{N}^n$. \\
Similarly, for $j$ such that $\bfmu_j$ has a probability of $\alpha/2$ to be selected, but different from $q$ we have
\begin{equation}
    \begin{aligned}
            &(\bfmu_j^\top, \boldsymbol{{0}^\top})\frac{1}{m}\sum_{i\in\mathcal{W}_n}\bfW_{O_{(i,\cdot)}}^{(t)}\sum_{s=1}^{l+1} (\bfW_V^{(t)}\bfp_s^n)\text{softmax}({\bfp_s^n}^\top{\bfW_K^{(t)}}^\top\bfW_Q^{(t)}\bfp_{query}^n)\\
    &\cdot(\bfW_K^{(t)}\bfp_s^n-\sum_{r=1}^{l+1} \text{softmax}({\bfp_r^n}^\top{\bfW_K^{(t)}}^\top\bfW_Q^{(t)}\bfp_{query}^n)\bfW_K^{(t)}\bfp_r^n){\bfp_{query}^n}^\top(\bfx_{query}^\top, \boldsymbol{{0}^\top})^\top\\
    \lesssim     &e^{-\delta^2 \beta^2}\cdot(\bfmu_q^\top, \boldsymbol{{0}^\top})\frac{1}{m}\sum_{i\in\mathcal{W}_n}\bfW_{O_{(i,\cdot)}}^{(t)}\sum_{s=1}^{l+1} (\bfW_V^{(t)}\bfp_s^n)\text{softmax}({\bfp_s^n}^\top{\bfW_K^{(t)}}^\top\bfW_Q^{(t)}\bfp_{query}^n)\\
    &\cdot(\bfW_K^{(t)}\bfp_s^n-\sum_{r=1}^{l+1} \text{softmax}({\bfp_r^n}^\top{\bfW_K^{(t)}}^\top\bfW_Q^{(t)}\bfp_{query}^n)\bfW_K^{(t)}\bfp_r^n){\bfp_{query}^n}^\top(\bfx_{query}^\top, \boldsymbol{{0}^\top})^\top.\label{l+1_2}
    \end{aligned}
\end{equation}
\noindent For $j\in[l]-\mathcal{N}_q^{n,i}$ with probability of $(1-\alpha)/(M_1-2)$ to be selected, with high probability, at most $1$ example has $\bfmu_j$ in each data. Then,
\begin{equation}
\begin{aligned}
    &(\bfmu_j^\top, \boldsymbol{{0}^\top})\frac{1}{m}\sum_{i\in\mathcal{W}_n}\bfW_{O_{(i,\cdot)}}^{(t)}\sum_{s=1}^{l+1} (\bfW_V^{(t)}\bfp_s^n)\text{softmax}({\bfp_s^n}^\top{\bfW_K^{(t)}}^\top\bfW_Q^{(t)}\bfp_{query}^n)\\
    &\cdot(\bfW_K^{(t)}\bfp_s^n-\sum_{r=1}^{l+1} \text{softmax}({\bfp_r^n}^\top{\bfW_K^{(t)}}^\top\bfW_Q^{(t)}\bfp_{query}^n)\bfW_K^{(t)}\bfp_r^n){\bfp_{query}^n}^\top(\bfx_{query}^\top, \boldsymbol{{0}^\top})^\top\\
    \leq &\frac{1}{l}\cdot(\bfmu_q^\top, \boldsymbol{{0}^\top})\frac{1}{m}\sum_{i\in\mathcal{W}_n}\bfW_{O_{(i,\cdot)}}^{(t)}\sum_{s=1}^{l+1} (\bfW_V^{(t)}\bfp_s^n)\text{softmax}({\bfp_s^n}^\top{\bfW_K^{(t)}}^\top\bfW_Q^{(t)}\bfp_{query}^n)\\
    &\cdot(\bfW_K^{(t)}\bfp_s^n-\sum_{r=1}^{l+1} \text{softmax}({\bfp_r^n}^\top{\bfW_K^{(t)}}^\top\bfW_Q^{(t)}\bfp_{query}^n)\bfW_K^{(t)}\bfp_r^n){\bfp_{query}^n}^\top(\bfx_{query}^\top, \boldsymbol{{0}^\top})^\top.\label{l+1_3}
\end{aligned}
\end{equation}
If $i\in\mathcal{U}_n$, since that $z^n=1$, the indicator by the Relu activation returns zero. Hence, we do not need to compute this case. If $i\notin \mathcal{W}_n\cup\mathcal{U}_n$, by the uniform distribution of $a_i$, we have that, for $\bfmu_q$ which is the IDR pattern of $\bfx_{query}$, 
\begin{equation}
    \begin{aligned}
    &({\bfmu_q}^\top, \boldsymbol{0}^\top) \sum_{i\notin\mathcal{W}_n\cup\mathcal{U}_n}a_{i}\mathbbm{1}[\bfW_{O_{(i,\cdot)}}^{(t)}\sum_{s=1}^{l+1}(\bfW_V^{(t)}\bfp_s^n) \text{softmax}({\bfp_s^n}^\top{\bfW_K^{(t)}}^\top\bfW_Q^{(t)}\bfp_l^n)\geq0]\\
    &\cdot\Big(\bfW_{O_{(i,\cdot)}}^{(t)}\sum_{s=1}^{l+1} (\bfW_V^{(t)}\bfp_s^n)\text{softmax}({\bfp_s^n}^\top{\bfW_K^{(t)}}^\top\bfW_Q^{(t)}\bfp_{query}^n)\\
    &\cdot(\bfW_K^{(t)}\bfp_s^n-\sum_{r=1}^{l+1} \text{softmax}({\bfp_r^n}^\top{\bfW_K^{(t)}}^\top\bfW_Q^{(t)}\bfp_{query}^n)\bfW_K^{(t)}\bfp_r^n){\bfp_{query}^n}^\top\Big)(\bfx_{query}^\top, \boldsymbol{{0}^\top})^\top\\
    \leq & \sqrt{\frac{\log m}{m}}\cdot(\bfmu_q^\top, \boldsymbol{{0}^\top})\frac{1}{m}\sum_{i\in\mathcal{W}_n}\bfW_{O_{(i,\cdot)}}^{(t)}\sum_{s=1}^{l+1} (\bfW_V^{(t)}\bfp_s^n)\text{softmax}({\bfp_s^n}^\top{\bfW_K^{(t)}}^\top\bfW_Q^{(t)}\bfp_{query}^n)\\
    &\cdot(\bfW_K^{(t)}\bfp_s^n-\sum_{r=1}^{l+1} \text{softmax}({\bfp_r^n}^\top{\bfW_K^{(t)}}^\top\bfW_Q^{(t)}\bfp_{query}^n)\bfW_K^{(t)}\bfp_r^n){\bfp_{query}^n}^\top(\bfx_{query}^\top, \boldsymbol{{0}^\top})^\top, \label{notin_up1}
    \end{aligned}
\end{equation}
where $\sqrt{\log m/m}$ is because $a_i$ can be either $+1$ and $-1$ following a uniform distribution in this case. For $\bfmu_{j}$ that has a probability of $\alpha/2$ to be selected but different from $\bfmu_q$, we have
\begin{equation}
    \begin{aligned}
    &({\bfmu_j}^\top, \boldsymbol{0}^\top) \sum_{i\notin\mathcal{W}_t\cup\mathcal{U}}a_{i}\mathbbm{1}[\bfW_{O_{(i,\cdot)}}^{(t)}\sum_{s=1}^{l+1}(\bfW_V^{(t)}\bfp_s^n) \text{softmax}({\bfp_s^n}^\top{\bfW_K^{(t)}}^\top\bfW_Q^{(t)}\bfp_l^n)\geq0]\\
    &\cdot\Big(\bfW_{O_{(i,\cdot)}}^{(t)}\sum_{s=1}^{l+1} (\bfW_V^{(t)}\bfp_s^n)\text{softmax}({\bfp_s^n}^\top{\bfW_K^{(t)}}^\top\bfW_Q^{(t)}\bfp_{query}^n)\\
    &\cdot(\bfW_K^{(t)}\bfp_s^n-\sum_{r=1}^{l+1} \text{softmax}({\bfp_r^n}^\top{\bfW_K^{(t)}}^\top\bfW_Q^{(t)}\bfp_{query}^n)\bfW_K^{(t)}\bfp_r^n){\bfp_{query}^n}^\top\Big)(\bfx_{query}^\top, \boldsymbol{{0}^\top})^\top\\
    \leq & e^{-\delta \beta^2}\sqrt{\frac{\log m}{m}}\cdot(\bfmu_q^\top, \boldsymbol{{0}^\top})\frac{1}{m}\sum_{i\in\mathcal{W}_n}\bfW_{O_{(i,\cdot)}}^{(t)}\sum_{s=1}^{l+1} (\bfW_V^{(t)}\bfp_s^n)\text{softmax}({\bfp_s^n}^\top{\bfW_K^{(t)}}^\top\bfW_Q^{(t)}\bfp_{query}^n)\\
    &\cdot(\bfW_K^{(t)}\bfp_s^n-\sum_{r=1}^{l+1} \text{softmax}({\bfp_r^n}^\top{\bfW_K^{(t)}}^\top\bfW_Q^{(t)}\bfp_{query}^n)\bfW_K^{(t)}\bfp_r^n){\bfp_{query}^n}^\top(\bfx_{query}^\top, \boldsymbol{{0}^\top})^\top. \label{notin_up2}
    \end{aligned}
\end{equation}
For $\bfx_j^n$ with $\bfmu_{j}$ that has a probability of $(1-\alpha)/(M_1-2)$ to be selected, we have
\begin{equation}
    \begin{aligned}
    &({\bfmu_j}^\top, \boldsymbol{0}^\top) \sum_{i\notin\mathcal{W}_t\cup\mathcal{U}}a_{i}\mathbbm{1}[\bfW_{O_{(i,\cdot)}}^{(t)}\sum_{s=1}^{l+1}(\bfW_V^{(t)}\bfp_s^n) \text{softmax}({\bfp_s^n}^\top{\bfW_K^{(t)}}^\top\bfW_Q^{(t)}\bfp_l^n)\geq0]\\
    &\cdot\Big(\bfW_{O_{(i,\cdot)}}^{(t)}\sum_{s=1}^{l+1} (\bfW_V^{(t)}\bfp_s^n)\text{softmax}({\bfp_s^n}^\top{\bfW_K^{(t)}}^\top\bfW_Q^{(t)}\bfp_{query}^n)\\
    &\cdot(\bfW_K^{(t)}\bfp_s^n-\sum_{r=1}^{l+1} \text{softmax}({\bfp_r^n}^\top{\bfW_K^{(t)}}^\top\bfW_Q^{(t)}\bfp_{query}^n)\bfW_K^{(t)}\bfp_r^n){\bfp_{query}}^\top\Big)(\bfx_{query}^\top, \boldsymbol{{0}^\top})^\top\\
    \leq & \frac{1}{l}\cdot \sqrt{\frac{\log m}{m}}\cdot(\bfmu_q^\top, \boldsymbol{{0}^\top})\frac{1}{m}\sum_{i\in\mathcal{W}_n}\bfW_{O_{(i,\cdot)}}^{(t)}\sum_{s=1}^{l+1} (\bfW_V^{(t)}\bfp_s^n)\text{softmax}({\bfp_s^n}^\top{\bfW_K^{(t)}}^\top\bfW_Q^{(t)}\bfp_{query}^n)\\
    &\cdot(\bfW_K^{(t)}\bfp_s^n-\sum_{r=1}^{l+1} \text{softmax}({\bfp_r^n}^\top{\bfW_K^{(t)}}^\top\bfW_Q^{(t)}\bfp_{query}^n)\bfW_K^{(t)}\bfp_r^n){\bfp_{query}^n}^\top(\bfx_{query}^\top, \boldsymbol{{0}^\top})^\top.\label{notin_up3}
    \end{aligned}
\end{equation}
Therefore, by (\ref{lower_l}), (\ref{notin_up1}), (\ref{notin_up2}), and (\ref{notin_up3}), we have that for one $\bfx_{query}$,
\begin{equation}
    \begin{aligned}
    &\Big|\eta\frac{1}{B}\sum_{n\in\mathcal{B}_b}(-z^n) ({\bfx_{query}}^\top, \boldsymbol{0}^\top)\sum_{i=1}^{m} a_{i}\mathbbm{1}[\bfW_{O_{(i,\cdot)}}^{(t)}\sum_{s=1}^{l+1}(\bfW_V^{(t)}\bfp_s^n)\cdot\text{softmax}({\bfp_s^n}^\top{\bfW_K^{(t)}}^\top\bfW_Q^{(t)}\bfp_l^n)\geq0]\\
    &\cdot\Big(\bfW_{O_{(i,\cdot)}}^{(t)}\sum_{s=1}^{l+1} (\bfW_V^{(t)}\bfp_s^n)\text{softmax}({\bfp_s^n}^\top{\bfW_K^{(t)}}^\top\bfW_Q^{(t)}\bfp_{query}^n)\\
    &\cdot(\bfW_K^{(t)}\bfp_s^n-\sum_{r=1}^{l+1} \text{softmax}({\bfp_r^n}^\top{\bfW_K^{(t)}}^\top\bfW_Q^{(t)}\bfp_{query}^n)\bfW_K^{(t)}\bfp_r^n){\bfp_{query}^n}^\top\Big)(\bfx_{query}^\top, \boldsymbol{{0}^\top})\Big|\\
    \geq & \eta\frac{1}{BM_1}\sum_{n\in\mathcal{B}_b}\frac{1}{m}\sum_{i\in\mathcal{W}_n}\zeta_{i,t}\delta \gamma_t (1-e^{-\delta^2 \beta^2}\sqrt{\frac{\log m}{m}}-\frac{1}{l}\sqrt{\frac{\log m}{m}})\beta^4\\
    \gtrsim & \eta\frac{1}{M_1}\zeta_{t}\delta \gamma_t \beta^4\label{W_Q_l+1},
    \end{aligned}
\end{equation}
as long as 
\begin{equation}
    m\gtrsim 1,
\end{equation}
and
\begin{equation}
    B\gtrsim M_1\log M_1,\label{B_bound}
\end{equation}
to ensure that
\begin{equation}
    \Pr\left(\sum_{n=1}^B \mathbbm{1}[m_n=\bfmu_j]\leq B(1-c)\cdot\frac{1}{M_1}\right)\leq e^{-c^2B\cdot \frac{1}{M_1}}=e^{-c\log M_1}=M_1^{-C},
\end{equation}
for some $c\in(0,1)$ and $C>1$, where $m_i$ denotes the IDR pattern in the query of the $n$-th data. 
Meanwhile, for $j\in[l]-\mathcal{N}_q^{n,i}$ that has a IDR pattern which forms a task in $\mathcal{T}_{tr}$ with the IDR pattern of $\bfx_{query}$, more indicators of $i\in\mathcal{U}_n$ is activated. 
\begin{equation}
    \begin{aligned}
    &-\Big|\eta\frac{1}{B}\sum_{n\in\mathcal{B}_b}(-z^n) ({\bfmu_j}^\top, \boldsymbol{0}^\top)\sum_{i=1}^{m} a_{i}\mathbbm{1}[\bfW_{O_{(i,\cdot)}}^{(t)}\sum_{s=1}^{l+1}(\bfW_V^{(t)}\bfp_s^n)\cdot\text{softmax}({\bfp_s^n}^\top{\bfW_K^{(t)}}^\top\bfW_Q^{(t)}\bfp_l^n)\geq0]\\
    &\cdot\Big(\bfW_{O_{(i,\cdot)}}^{(t)}\sum_{s=1}^{l+1} (\bfW_V^{(t)}\bfp_s^n)\text{softmax}({\bfp_s^n}^\top{\bfW_K^{(t)}}^\top\bfW_Q^{(t)}\bfp_{query}^n)\\
    &\cdot(\bfW_K^{(t)}\bfp_s^n-\sum_{r=1}^{l+1} \text{softmax}({\bfp_r^n}^\top{\bfW_K^{(t)}}^\top\bfW_Q^{(t)}\bfp_{query}^n)\bfW_K^{(t)}\bfp_r^n){\bfp_{query}^n}^\top\Big)(\bfx_{query}^\top, \boldsymbol{{0}^\top})^\top\Big|\\
    \geq & -\frac{1}{2}e^{-\delta^2 \beta^2}\eta\frac{1}{B}\sum_{n\in\mathcal{B}_b}(-z^n) ({\bfmu_q}^\top, \boldsymbol{0}^\top)\sum_{i=1}^{m} a_{i}\mathbbm{1}[\bfW_{O_{(i,\cdot)}}^{(t)}\sum_{s=1}^{l+1}(\bfW_V^{(t)}\bfp_s^n)\cdot\text{softmax}({\bfp_s^n}^\top{\bfW_K^{(t)}}^\top\bfW_Q^{(t)}\bfp_l^n)\geq0]\\
    &\cdot\Big(\bfW_{O_{(i,\cdot)}}^{(t)}\sum_{s=1}^{l+1} (\bfW_V^{(t)}\bfp_s^n)\text{softmax}({\bfp_s^n}^\top{\bfW_K^{(t)}}^\top\bfW_Q^{(t)}\bfp_{query}^n)\\
    &\cdot(\bfW_K^{(t)}\bfp_s^n-\sum_{r=1}^{l+1} \text{softmax}({\bfp_r^n}^\top{\bfW_K^{(t)}}^\top\bfW_Q^{(t)}\bfp_{query}^n)\bfW_K^{(t)}\bfp_r^n){\bfp_{query}^n}^\top\Big)(\bfx_{query}^\top, \boldsymbol{{0}^\top})^\top\label{W_Q_j}.
    \end{aligned}
\end{equation}
For other $j\in[l]-\mathcal{N}_q^{n,i}$, 
\begin{equation}
    \begin{aligned}
    &\Big|\eta\frac{1}{B}\sum_{n\in\mathcal{B}_b}(-z^n) ({\bfmu_j}^\top, \boldsymbol{0}^\top)\sum_{i=1}^{m} a_{i}\mathbbm{1}[\bfW_{O_{(i,\cdot)}}^{(t)}\sum_{s=1}^{l+1}(\bfW_V^{(t)}\bfp_s^n)\cdot\text{softmax}({\bfp_s^n}^\top{\bfW_K^{(t)}}^\top\bfW_Q^{(t)}\bfp_l^n)\geq0]\\
    &\cdot\Big(\bfW_{O_{(i,\cdot)}}^{(t)}\sum_{s=1}^{l+1} (\bfW_V^{(t)}\bfp_s^n)\text{softmax}({\bfp_s^n}^\top{\bfW_K^{(t)}}^\top\bfW_Q^{(t)}\bfp_{query}^n)\\
    &\cdot(\bfW_K^{(t)}\bfp_s^n-\sum_{r=1}^{l+1} \text{softmax}({\bfp_r^n}^\top{\bfW_K^{(t)}}^\top\bfW_Q^{(t)}\bfp_{query}^n)\bfW_K^{(t)}\bfp_r^n){\bfp_{query}^n}^\top\Big)(\bfx_{query}^\top, \boldsymbol{{0}^\top})^\top\Big|\\
    \leq & \frac{1}{ M_1}\eta\frac{1}{B}\sum_{n\in\mathcal{B}_b}(-z^n) ({\bfmu_q^n}^\top, \boldsymbol{0}^\top)\sum_{i=1}^{m} a_{i}\mathbbm{1}[\bfW_{O_{(i,\cdot)}}^{(t)}\sum_{s=1}^{l+1}(\bfW_V^{(t)}\bfp_s^n)\cdot\text{softmax}({\bfp_s^n}^\top{\bfW_K^{(t)}}^\top\bfW_Q^{(t)}\bfp_l^n)\geq0]\\
    &\cdot\Big(\bfW_{O_{(i,\cdot)}}^{(t)}\sum_{s=1}^{l+1} (\bfW_V^{(t)}\bfp_s^n)\text{softmax}({\bfp_s^n}^\top{\bfW_K^{(t)}}^\top\bfW_Q^{(t)}\bfp_{query}^n)\\
    &\cdot(\bfW_K^{(t)}\bfp_s^n-\sum_{r=1}^{l+1} \text{softmax}({\bfp_r^n}^\top{\bfW_K^{(t)}}^\top\bfW_Q^{(t)}\bfp_{query}^n)\bfW_K^{(t)}\bfp_r^n){\bfp_{query}^n}^\top\Big)(\bfx_{query}^\top, \boldsymbol{{0}^\top})^\top\label{W_Q_j2},
    \end{aligned}
\end{equation}
where $M_1$ comes from the fact that the softmax value between $\bfp_{query}^n$ and $\bfp_r^n$ with $\bfmu_j$ as the IDR pattern of $\bfp_r^n$ is $\Theta(1-\gamma_t)/M_1$ in average of $B\gtrsim M_1\log M_1$ samples. %$\sqrt{\log Bm/Bm}$ is because the average is over $Bm$ terms with labels following a uniform distribution on $\{+1,-1\}$. 
Then, by combining (\ref{W_Q_l+1}), (\ref{W_Q_j}), and (\ref{W_Q_j2}), we have
\begin{equation}
    \begin{aligned}
        &(\bfmu_q^\top, \boldsymbol{0}^\top)\eta\frac{1}{B}\sum_{n\in\mathcal{B}_b}\frac{\partial \ell(\tilde{\bfP}^n,z^n;\Psi)}{\partial \bfW_Q}\Big|_{t=0}(\bfx_{query}^\top, \boldsymbol{{0}}^\top)^\top\\
        \gtrsim & \eta\frac{1}{M_1}\zeta_{t}\delta \gamma_t \beta^4.
    \end{aligned}
\end{equation}
By (\ref{W_Q_j}) and (\ref{W_Q_j2}), we have that for $\bfmu_j$ which forms a task in $\mathcal{T}_{tr}$ with the $\bfmu_q$,
\begin{equation}
    \begin{aligned}
        &-\Big|(\bfmu_j^\top, \boldsymbol{0}^\top)\eta\frac{1}{B}\sum_{n\in\mathcal{B}_b}\frac{\partial \ell(\tilde{\bfP}^n,z^n;\Psi)}{\partial \bfW_Q}\Big|_{t=0}(\bfx_{query}^\top, \boldsymbol{{0}}^\top)^\top\Big|\\
        \geq & -\frac{1}{2}e^{-\delta^2 \beta^2}\Big|(\bfmu_q^\top, \boldsymbol{0}^\top)\eta\frac{1}{B}\sum_{n\in\mathcal{B}_b}\frac{\partial \ell(\tilde{\bfP}^n,z^n;\Psi)}{\partial \bfW_Q}\Big|_{t=0}(\bfx_{query}^\top, \boldsymbol{{0}}^\top)^\top\Big|.
    \end{aligned}
\end{equation}
For $\bfmu_j$ which does not form a task in $\mathcal{T}_{tr}$ with the $\bfmu_q$,
\begin{equation}
    \begin{aligned}
        &\Big|(\bfmu_j^\top, \boldsymbol{0}^\top)\eta\frac{1}{B}\sum_{n\in\mathcal{B}_b}\frac{\partial \ell(\tilde{\bfP}^n,z^n;\Psi)}{\partial \bfW_Q}\Big|_{t=0}(\bfx_{query}^\top, \boldsymbol{{0}}^\top)^\top\Big|\\
        \leq & \frac{1}{M_1}\Big|(\bfmu_q^\top, \boldsymbol{0}^\top)\eta\frac{1}{B}\sum_{n\in\mathcal{B}_b}\frac{\partial \ell(\tilde{\bfP}^n,z^n;\Psi)}{\partial \bfW_Q}\Big|_{t=0}(\bfx_{query}^\top, \boldsymbol{{0}}^\top)^\top\Big|.
    \end{aligned}
\end{equation}
Similarly, for $k\in[M_2]$, 
\begin{equation}
    \begin{aligned}
        &\Big|(\bfnu_k^\top, \boldsymbol{0}^\top)\eta\frac{1}{B}\sum_{n\in\mathcal{B}_b}\frac{\partial \ell(\tilde{\bfP}^n,z^n;\Psi)}{\partial \bfW_Q}\Big|_{t=0}(\bfx_{query}^\top, \boldsymbol{{0}}^\top)^\top\Big|\\
        \lesssim & \frac{1}{M_2}\Big|(\bfmu_q^\top, \boldsymbol{0}^\top)\eta\frac{1}{B}\sum_{n\in\mathcal{B}_b}\frac{\partial \ell(\tilde{\bfP}^n,z^n;\Psi)}{\partial \bfW_Q}\Big|_{t=0}(\bfx_{query}^\top, \boldsymbol{{0}}^\top)^\top\Big|,
    \end{aligned}
\end{equation}
where $M_2$ comes from that for $\nu_k$ that is added to $\bfmu_j$, the contribution of gradient is $1/M_2$ times of replacing $\bfnu_k$ with $\bfmu_j$. Hence, $1/M_2\cdot(1+1/M_1\cdot M_1)=2/M_2=\Theta(1/M_2)$. 
\iffalse
Similarly, since that with probability $1/M_2=\Theta(1/M)$ one demonstration contains $\bfnu_j$, then by Chernoff bounds in Lemma \ref{lemma: chernoff},
\begin{equation}
    \Pr(\sum_{i=1}^{Bl}\mathbbm{1}[\bfx_i^n\text{ contains }\bfnu_k]\geq (1+\frac{M_2}{Bl}-1)\frac{Bl}{M_2})\leq e^{-Bl\cdot (\frac{M_2}{Bl})^2\cdot \frac{1}{M_2}}=e^{-\frac{M_2}{Bl}}\lesssim e^{-\frac{M}{Bl}}
\end{equation}
\fi
\iffalse
If $Bl\lesssim M_1\log^2 M$, we have that at most one demonstration contains $\bfv_j$ in the whole batch $\mathcal{B}_b$ for any $j\in[M_2]$. Therefore,
\begin{equation}
    \begin{aligned}
        \Big|(\bfnu_j^\top, \boldsymbol{0}^\top)\eta\frac{1}{B}\sum_{n\in\mathcal{B}_b}\frac{\partial \ell(\tilde{\bfP}^n,z^n;\Psi)}{\partial \bfW_Q}\Big|_{t=0}(\bfnu_j^\top, \boldsymbol{{0}}^\top)^\top\Big|
        \lesssim  \eta\frac{1}{M}\zeta_i\delta
    \end{aligned}
\end{equation}
\begin{equation}
    \begin{aligned}
        \Big|(\bfnu_l^\top, \boldsymbol{0}^\top)\eta\frac{1}{B}\sum_{n\in\mathcal{B}_b}\frac{\partial \ell(\tilde{\bfP}^n,z^n;\Psi)}{\partial \bfW_Q}\Big|_{t=0}(\bfnu_j^\top, \boldsymbol{{0}}^\top)^\top\Big|
        \lesssim  \eta\frac{1}{M}\zeta_i\delta
    \end{aligned}
\end{equation}
\begin{equation}
    \begin{aligned}
        \Big|(\bfmu_k^\top, \boldsymbol{0}^\top)\eta\frac{1}{B}\sum_{n\in\mathcal{B}_b}\frac{\partial \ell(\tilde{\bfP}^n,z^n;\Psi)}{\partial \bfW_Q}\Big|_{t=0}(\bfnu_j^\top, \boldsymbol{{0}}^\top)^\top\Big|
        \lesssim  \eta\frac{1}{M}\zeta_i\delta
    \end{aligned}
\end{equation}
\fi
For the label embedding, we have
\begin{equation}
\begin{aligned}
    &\eta\frac{1}{B}\sum_{n\in\mathcal{B}_b}\frac{\partial \ell(\tilde{\bfP}^n,z^n;\Psi)}{\partial \bfW_Q}\Big|_{t=0}[:, d_\mathcal{X}+1:d_\mathcal{X}+d_\mathcal{Y}]\\
    =&\eta\frac{1}{B}\sum_{n\in\mathcal{B}_b}(-z^n) \sum_{i=1}^{m} a_{i}\mathbbm{1}[\bfW_{O_{(i,\cdot)}}^{(t)}\sum_{s=1}^{l+1}(\bfW_V^{(t)}\bfp_s^n) \text{softmax}({\bfp_s^n}^\top{\bfW_K^{(t)}}^\top\bfW_Q^{(t)}\bfp_l^n)\geq0]\\
    &\cdot\Big(\bfW_{O_{(i,\cdot)}}^{(t)}\sum_{s=1}^{l+1} (\bfW_V^{(t)}\bfp_s^n)\text{softmax}({\bfp_s^n}^\top{\bfW_K^{(t)}}^\top\bfW_Q^{(t)}\bfp_{query}^n)\\
    &\cdot(\bfW_K^{(t)}\bfp_s^n-\sum_{r=1}^{l+1} \text{softmax}({\bfp_r^n}^\top{\bfW_K^{(t)}}^\top\bfW_Q^{(t)}\bfp_{query}^n)\bfW_K^{(t)}\bfp_r^n){\bfp_{query}}^\top\Big)[:, d_\mathcal{X}+1:d_\mathcal{X}+d_\mathcal{Y}].\\
    =&\boldsymbol{0}.
\end{aligned}
\end{equation}
We then have
\begin{equation}
\begin{aligned}
    &\Big|\bfq^\top\eta\frac{1}{B}\sum_{n\in\mathcal{B}_b}(-z^n) \sum_{i=1}^{m} a_{i}\mathbbm{1}[\bfW_{O_{(i,\cdot)}}^{(t)}\sum_{s=1}^{l+1}(\bfW_V^{(t)}\bfp_s^n)\text{softmax}({\bfp_s^n}^\top{\bfW_K^{(t)}}^\top\bfW_Q^{(t)}\bfp_l^n)\geq0]\\
    &\cdot\Big(\bfW_{O_{(i,\cdot)}}^{(t)}\sum_{s=1}^{l+1} (\bfW_V^{(t)}\bfp_s^n)\text{softmax}({\bfp_s^n}^\top{\bfW_K^{(t)}}^\top\bfW_Q^{(t)}\bfp_{query}^n)\\
    &\cdot(\bfW_K^{(t)}\bfp_s^n-\sum_{r=1}^{l+1} \text{softmax}({\bfp_r^n}^\top{\bfW_K^{(t)}}^\top\bfW_Q^{(t)}\bfp_{query}^n)\bfW_K^{(t)}\bfp_r^n)\Big)[d_\mathcal{X}+1:d_\mathcal{X}+d_\mathcal{Y}]\Big|\\
    =&0.\label{y_space_bound}
\end{aligned}
\end{equation}
\iffalse
where the last step comes from the equal probability of two signs and the upper bound of the inner product.\\
We need 
\begin{equation}
    B\geq \frac{(\epsilon_y^{-1}\xi M)^2}{\zeta_i^2}
\end{equation}
to make (\ref{y_space_bound}) upper bounded by $\epsilon_y\in(0,1/2)$. \\
\fi
\iffalse
\noindent If $\mathcal{N}^n=\emptyset$, there exists an $s_*$ such that when $s=s_*$,
\begin{equation}
    \text{softmax}({\bfp_s^n}^\top{\bfW_K^{(t)}}^\top\bfW_Q^{(t)}\bfp_{query}^n)\geq \frac{e^{\delta^2 }}{e^{\delta^2 }+l e^{\delta^2}},
\end{equation}
and when $s\neq s_*$, 
\begin{equation}
    \text{softmax}({\bfp_s^n}^\top{\bfW_K^{(t)}}^\top\bfW_Q^{(t)}\bfp_{query}^n)\leq \frac{e^{\delta^2}}{e^{\delta^2 }+le^{\delta^2}},
\end{equation}
Therefore, for $z^n=+1$ and $i\in\mathcal{W}$,
\begin{equation}
    \begin{aligned}        \bfW_{O_{(i,\cdot)}}^{(t)}\sum_{s=1}^{l+1}(\bfW_V\bfp_s^n)\text{softmax}({\bfp_s^n}^\top{\bfW_K^{(t)}}^\top\bfW_Q^{(t)}\bfp_{query}^n)
    >0
    \end{aligned}
\end{equation}
may not hold when $t=0$.\\
\fi
\noindent Hence, the conclusion holds when $t=1$. Suppose that the statement also holds when $t=t_0$. When $t=t_0+1$, the gradient update is the same as in (\ref{W_Q_l+1}) and (\ref{W_Q_j}). Note that the indicator of $\mathcal{W}_n$ will not change along the training. The only difference is the changes in $\zeta_t$ and $\gamma_t$. Thus, we can obtain that for $\bfmu_q$ with the same IDR pattern as $\bfx_{query}$,
\begin{equation}
    \begin{aligned}
        &(\bfmu_q^\top, \boldsymbol{0}^\top)\eta\frac{1}{B}\sum_{n\in\mathcal{B}_b}\frac{\partial \ell(\tilde{\bfP}^n,z^n;\Psi)}{\partial \bfW_Q}\Big|_{t=t_0+1}(\bfx_{query}^\top, \boldsymbol{{0}}^\top)^\top\\
        \gtrsim & \eta\frac{1}{M_1}\sum_{b=0}^{t_0}\zeta_b\delta \gamma_b \beta^4\\
        \gtrsim & \eta\frac{1}{M_1}\sum_{b=0}^{t_0}\zeta_b\delta \gamma_b \beta^4,
    \end{aligned}
\end{equation}
as long as (\ref{B_bound}) holds. We also have
\begin{equation}
    \eta\frac{1}{B}\sum_{n\in\mathcal{B}_b}\frac{\partial \ell(\tilde{\bfP}^n,z^n;\Psi)}{\partial \bfW_Q}\Big|_{t=t_0+1}[:, d_\mathcal{X}+1:d_\mathcal{X}+d_\mathcal{Y}]=\boldsymbol{0}.
\end{equation}
\iffalse
We know that $\bfW_Q$ is used for the computation with the $l+1$-th input. Then we have
\begin{equation}
    \begin{aligned}
        &(\bfmu_j^\top, \bfq^\top)\eta\frac{1}{B}\sum_{n\in\mathcal{B}_b}\frac{\partial \ell(\tilde{\bfP}^n,z^n;\Psi)}{\partial \bfW_Q}\Big|_{t=t_0+1}(\bfmu_j^\top, \boldsymbol{{0}}^\top)^\top\\
        \gtrsim  & \eta\frac{1}{M_1}\sum_{b=0}^{t_0}\zeta_b\delta \gamma_b \beta^2,
    \end{aligned}
\end{equation}
\begin{equation}
    \begin{aligned}
        &\Big\|\eta\frac{1}{B}\sum_{n\in\mathcal{B}_b}\frac{\partial \ell(\tilde{\bfP}^n,z^n;\Psi)}{\partial \bfW_Q}\Big|_{t=t_0+1}(\bfmu_j^\top, \boldsymbol{{0}}^\top)^\top\Big\|\\
        \gtrsim  & \eta\frac{1}{M_1}\sum_{b=0}^{t_0}\zeta_b\delta \gamma_b \beta^2,
    \end{aligned}
\end{equation}
\fi
%where the last step comes from the basic mathematical computation. \\
Similarly, for $j\neq q$ and $j\in[M_1]$ where $\bfmu_l$ does not form a task in $\mathcal{T}_{tr}$ with the $\bfmu_q$,
\begin{equation}
    \begin{aligned}
        &\Big|(\bfmu_j^\top, \boldsymbol{0}^\top)\eta\frac{1}{B}\sum_{n\in\mathcal{B}_b}\frac{\partial \ell(\tilde{\bfP}^n,z^n;\Psi)}{\partial \bfW_Q}\Big|_{t=t_0}(\bfx_{query}^\top, \boldsymbol{{0}}^\top)^\top\Big|\\
        \lesssim  &  \frac{1}{M_1}\Big|(\bfmu_q^\top, \boldsymbol{0}^\top)\eta\frac{1}{B}\sum_{n\in\mathcal{B}_b}\frac{\partial \ell(\tilde{\bfP}^n,z^n;\Psi)}{\partial \bfW_Q}\Big|_{t=t_0}(\bfx_{query}^\top, \boldsymbol{{0}}^\top)^\top\Big|.
    \end{aligned}
\end{equation}
\iffalse
When summing up over $t$ for $t\geq 1$, we have
\begin{equation}
    \begin{aligned}
        &\Big|(\bfmu_l^\top, \boldsymbol{0}^\top)\eta\frac{1}{B}\sum_{b=0}^{t_0-1}\sum_{n\in\mathcal{B}_b}\frac{\partial \ell(\tilde{\bfP}^n,z^n;\Psi)}{\partial \bfW_Q}\Big|_{t=t_0}(\bfmu_j^\top, \boldsymbol{{0}}^\top)^\top\Big|\\
        \lesssim  &  \frac{1}{l}\sqrt{\frac{\log BmT}{BmT}}\Big|(\bfmu_l^\top, \boldsymbol{0}^\top)\eta\frac{1}{B}\sum_{b=0}^{t_0-1}\sum_{n\in\mathcal{B}_b}\frac{\partial \ell(\tilde{\bfP}^n,z^n;\Psi)}{\partial \bfW_Q}\Big|_{t=t_0}(\bfmu_j^\top, \boldsymbol{{0}}^\top)^\top\Big|,
    \end{aligned}
\end{equation}
\fi
For $j\neq q$ and $j\in[M_1]$ where $\bfmu_j$ forms a task in $\mathcal{T}_{tr}$ with $\bfmu_q$,
\begin{equation}
    \begin{aligned}
        &-\Big|(\bfmu_j^\top, \boldsymbol{0}^\top)\eta\frac{1}{B}\sum_{n\in\mathcal{B}_b}\frac{\partial \ell(\tilde{\bfP}^n,z^n;\Psi)}{\partial \bfW_Q}\Big|_{t=t_0}(\bfx_{query}^\top, \boldsymbol{{0}}^\top)^\top\Big|\\
        \gtrsim  &  -e^{-\delta^2 \beta^2-(\eta\frac{1}{M_1}\sum_{b=0}^{t_0-1}\zeta_b\delta\gamma_b\beta^2)^2}\Big|(\bfmu_q^\top, \boldsymbol{0}^\top)\eta\frac{1}{B}\sum_{n\in\mathcal{B}_b}\frac{\partial \ell(\tilde{\bfP}^n,z^n;\Psi)}{\partial \bfW_Q}\Big|_{t=t_0}(\bfx_{query}^\top, \boldsymbol{{0}}^\top)^\top\Big|\\
        \gtrsim &-e^{-\Theta(\frac{\eta t_0}{M_1})^2}\Big|(\bfmu_q^\top, \boldsymbol{0}^\top)\eta\frac{1}{B}\sum_{n\in\mathcal{B}_b}\frac{\partial \ell(\tilde{\bfP}^n,z^n;\Psi)}{\partial \bfW_Q}\Big|_{t=t_0}(\bfx_{query}^\top, \boldsymbol{{0}}^\top)^\top\Big|,
    \end{aligned}
\end{equation}
where the first step comes from the fact that a negative gradient update makes the softmax value of $\bfmu_l$ much smaller. The last step is obtained in the order related to $\eta, t, M_1$ as variables.  Meanwhile, for $k\in[M_2]$, 
\begin{equation}
    \begin{aligned}
        &\Big|(\bfnu_k^\top, \boldsymbol{0}^\top)\eta\frac{1}{B}\sum_{n\in\mathcal{B}_b}\frac{\partial \ell(\tilde{\bfP}^n,z^n;\Psi)}{\partial \bfW_Q}\Big|_{t=t_0}(\bfx_{query}^\top, \boldsymbol{{0}}^\top)^\top\Big|\\
        \lesssim  &  \frac{1}{M_2}\Big|(\bfmu_q^\top, \boldsymbol{0}^\top)\eta\frac{1}{B}\sum_{n\in\mathcal{B}_b}\frac{\partial \ell(\tilde{\bfP}^n,z^n;\Psi)}{\partial \bfW_Q}\Big|_{t=t_0}(\bfx_{query}^\top, \boldsymbol{{0}}^\top)^\top\Big|.
    \end{aligned}\label{nu new}
\end{equation}

\noindent(b) Then we study the updates of $\bfW_K$. We can compute the gradient as
\begin{equation}
\begin{aligned}
    &\eta \frac{1}{B}\sum_{n\in\mathcal{B}_b} \frac{\partial \ell(\tilde{\bfP}^n, z^n, \Psi)}{\partial \bfW_K}\\
    =&\eta\frac{1}{B}\sum_{n\in\mathcal{B}_b}(-z^n)\sum_{i=1}^m a_{i}\mathbbm{1}[\bfW_{O_{(i,\cdot)}}\sum_{s=1}^{l+1}(\bfW_V\bfp_s^n)\cdot\text{softmax}({\bfp_s^n}^\top\bfW_K^\top\bfW_Q\bfp_{query}^n)\geq0]\\
    &\cdot\Big(\bfW_{O_{(i,\cdot)}}\sum_{s=1}^{l+1} (\bfW_V\bfp_s^n)\text{softmax}({\bfp_s^n}^\top\bfW_K^\top\bfW_Q\bfp_{query})\bfW_Q^\top\bfp_{query}^n\\
    &\cdot(\bfp_s^n-\sum_{r=1}^{l+1} \text{softmax}({\bfp_r^n}^\top\bfW_K^\top\bfW_Q\bfp_{query}^n)\bfp_r^n)^\top\Big).\label{grad_O}
\end{aligned}
\end{equation}
If we investigate $\bfW_K^{(t)}\bfp_s^n$, we can tell that the output is a weighed summation of multiple ${\bfW_Q^{(t)}}\bfp_{query}^n$. Similarly, the output of $\bfW_Q^{(t)}\bfp_{query}^n$ is a weighed summation of multiple ${\bfW_K^{(t)}}\bfp_s$. Given the initialization $\bfW_Q^{(0)}$ and $\bfW_K^{(0)}$, the update of $\bfW_K^{(t)}\bfp_s^n$ and $\bfW_Q^{(t)}\bfp_{query}^n$ only contains the contribution from the feature space embeddings at the initialization. One difference is that since that $\bfq$ appears with $1/2$ probability in all $\bfp_s^n$, 
\begin{equation}
    \begin{aligned}
        &\Big|\eta\frac{1}{B}\sum_{n\in\mathcal{B}_b}(-z^n)\sum_{i=1}^m a_{i}\mathbbm{1}[\bfW_{O_{(i,\cdot)}}\sum_{s=1}^{l+1}(\bfW_V\bfp_s^n)\cdot\text{softmax}({\bfp_s^n}^\top\bfW_K^\top\bfW_Q\bfp_{query}^n)\geq0]\\
    &\cdot\Big(\bfW_{O_{(i,\cdot)}}\sum_{s=1}^{l+1} (\bfW_V\bfp_s^n)\text{softmax}({\bfp_s^n}^\top\bfW_K^\top\bfW_Q\bfp_{query})\bfW_Q^\top\bfp_{query}^n\\
    &\cdot(\bfp_s^n-\sum_{r=1}^{l+1} \text{softmax}({\bfp_r^n}^\top\bfW_K^\top\bfW_Q\bfp_{query}^n)\bfp_r^n)^\top\Big)[d_\mathcal{X}+1:d_\mathcal{X}+d_\mathcal{Y}]\bfq\Big|\\
    \leq &\sqrt{\frac{\log B}{B}} (\bfmu_j^\top, \boldsymbol{0}^\top)\eta\frac{1}{B}\sum_{n\in\mathcal{B}_b}\frac{\partial \ell(\tilde{\bfP}^n,z^n;\Psi)}{\partial \bfW_K}\Big|_{t=t_0+1}(\bfmu_j^\top, \boldsymbol{{0}}^\top)^\top.\label{W_K_labe_part}
    \end{aligned}
\end{equation}
Following the steps in Part (a), we can obtain that for $\bfmu_q$ as the IDR pattern of $\bfx_q$, $e\in[l]$,
\begin{equation}
    \begin{aligned}
        &(\bfmu_q^\top, \boldsymbol{0}^\top)\eta\frac{1}{B}\sum_{n\in\mathcal{B}_b}\frac{\partial \ell(\tilde{\bfP}^n,z^n;\Psi)}{\partial \bfW_K}\Big|_{t=t_0+1}(\bfx_e^\top, \boldsymbol{{0}}^\top)^\top\\
        \gtrsim & \eta\frac{1}{M_1}\sum_{b=0}^{t_0}\zeta_b\delta \gamma_b \beta^2,
    \end{aligned}
\end{equation}
\iffalse
\begin{equation}
    \eta\frac{1}{B}\sum_{n\in\mathcal{B}_b}\frac{\partial \ell(\tilde{\bfP}^n,z^n;\Psi)}{\partial \bfW_K}\Big|_{t=t_0+1}[ d_\mathcal{X}+1:d_\mathcal{X}+d_\mathcal{Y},:]=\boldsymbol{0},
\end{equation}
\fi
and combining (\ref{W_K_labe_part}),
\begin{equation}
    \begin{aligned}
        &(\bfmu_q^\top, \boldsymbol{0}^\top)\eta\frac{1}{B}\sum_{n\in\mathcal{B}_b}\frac{\partial \ell(\tilde{\bfP}^n,z^n;\Psi)}{\partial \bfW_K}\Big|_{t=t_0+1}\bfp_q\\
        \gtrsim & \eta\frac{1}{M_1}\sum_{b=0}^{t_0}\zeta_b\delta \gamma_b \beta^2(1-\sqrt{\frac{\log B}{B}})\\
        \gtrsim &\eta\frac{1}{M_1}\sum_{b=0}^{t_0}\zeta_b\delta \gamma_b \beta^2,
    \end{aligned}
\end{equation}
where the last step holds as long as (\ref{B_bound}). 
\iffalse
\begin{equation}
    \begin{aligned}
        &\Big\|\eta\frac{1}{B}\sum_{n\in\mathcal{B}_b}\frac{\partial \ell(\tilde{\bfP}^n,z^n;\Psi)}{\partial \bfW_K}\Big|_{t=t_0+1}(\bfmu_j^\top, \bfq^\top)^\top\Big\|\\
        \gtrsim & \eta\frac{1}{M_1}\sum_{b=0}^{t_0}\zeta_b\delta \gamma_b \beta^2.
    \end{aligned}
\end{equation}
\fi
For $\bfmu_j$ which forms a task in $\mathcal{T}_{tr}$ with the $\bfmu_q$,
\begin{equation}
    \begin{aligned}
        &-\Big|(\bfmu_j^\top, \boldsymbol{0}^\top)\eta\frac{1}{B}\sum_{n\in\mathcal{B}_b}\frac{\partial \ell(\tilde{\bfP}^n,z^n;\Psi)}{\partial \bfW_K}\Big|_{t=t_0}\bfp_q\Big|\\
        \gtrsim  &   -e^{-\Theta(\frac{\eta t_0}{M_1})^2}\Big|(\bfmu_q^\top, \boldsymbol{0}^\top)\eta\frac{1}{B}\sum_{n\in\mathcal{B}_b}\frac{\partial \ell(\tilde{\bfP}^n,z^n;\Psi)}{\partial \bfW_K}\Big|_{t=t_0}\bfp_q\Big|.
    \end{aligned}
\end{equation}
For $\bfmu_j$ which does not form a task in $\mathcal{T}_{tr}$ with the $\bfmu_q$,
\begin{equation}
    \begin{aligned}
        &\Big|(\bfmu_j^\top, \boldsymbol{0}^\top)\eta\frac{1}{B}\sum_{n\in\mathcal{B}_b}\frac{\partial \ell(\tilde{\bfP}^n,z^n;\Psi)}{\partial \bfW_K}\Big|_{t=t_0}\bfp_q\Big|\\
        \leq &   \frac{1}{M_1}\Big|(\bfmu_j^\top, \boldsymbol{0}^\top)\eta\frac{1}{B}\sum_{n\in\mathcal{B}_b}\frac{\partial \ell(\tilde{\bfP}^n,z^n;\Psi)}{\partial \bfW_K}\Big|_{t=t_0}\bfp_q\Big|.
    \end{aligned}
\end{equation}
Meanwhile, for $k\in[M_2]$, similar to (\ref{nu new}),
\begin{equation}
    \begin{aligned}
        &\Big|(\bfnu_k^\top, \boldsymbol{0}^\top)\eta\frac{1}{B}\sum_{n\in\mathcal{B}_b}\frac{\partial \ell(\tilde{\bfP}^n,z^n;\Psi)}{\partial \bfW_K}\Big|_{t=t_0}\bfp_q\Big|\\
        \leq &   \frac{1}{M_2}\Big|(\bfmu_j^\top, \boldsymbol{0}^\top)\eta\frac{1}{B}\sum_{n\in\mathcal{B}_b}\frac{\partial \ell(\tilde{\bfP}^n,z^n;\Psi)}{\partial \bfW_K}\Big|_{t=t_0}\bfp_q\Big|.
    \end{aligned}
\end{equation}

\end{proof}

\subsection{Proof of Lemma \ref{lemma: V}}
\begin{proof}
\begin{equation}
\begin{aligned}
    &\quad\eta\frac{1}{B}\sum_{n\in\mathcal{B}_b}\frac{\partial \ell(\tilde{\bfP}^n,z^n;\Psi)}{\partial \bfW_V}\\
    &=\eta\frac{1}{B}\sum_{n\in\mathcal{B}_b} \frac{\partial \ell(\tilde{\bfP}^n,z^n;\Psi)}{\partial F(\bfp_{query}^n)}\frac{\partial F(\bfp_{query}^n)}{\partial \bfW_V}\\
    &=\eta\frac{1}{B}\sum_{n\in\mathcal{B}_b}(-z^n) \sum_{i=1}^m a_{i}\mathbbm{1}[\bfW_{O_{(i,\cdot)}}\sum_{s=1}^{l+1}(\bfW_V\bfp_s^n)\text{softmax}({\bfp_s^n}^\top\bfW_K^\top\bfW_Q\bfp_{query}^n)\geq0]\\
    &\cdot\bfW_{O_{(i,\cdot)}}^\top\sum_{s=1}^{l+1}\text{softmax}({\bfp_s^n}^\top\bfW_K^\top\bfW_Q\bfp_{query}^n){\bfp_s^n}^\top.\label{grad_V}
\end{aligned}
\end{equation}
Let $\bfx_i^n$ and $\bfx_j^n$ correspond to IDR patterns $\bfmu_a$ and $\bfmu_b$, respectively. For $\bfp_{query}^n$ which corresponds to the IDR feature $\bfmu_{a}$, 
\begin{equation}
    \begin{aligned}
        \sum_{s=1}^{l+1}\text{softmax}({\bfp_s^n}^\top\bfW_K^\top\bfW_Q\bfp_{query}^n){\bfp_s^n}^\top({\bfx_i^n}^\top,\bfq^\top)^\top&\gtrsim  \beta^2(1-\gamma_t)\cdot  1- \frac{1}{\frac{\alpha}{2}l}\\
        &\gtrsim \beta^2(1-\gamma_t)\cdot 1,\label{WV_part1}
    \end{aligned}
\end{equation}
where the first step holds since that by (\ref{prob_kappa}), with high probability, no other $\bfx_k^n$ where $k\neq l+1$ shares the same IDI pattern as $\bfx_{query}^n$. The last step holds if
\begin{equation}
    l_{tr}\gtrsim \frac{1}{\alpha\beta^2}.\label{l_tr_2}
\end{equation}
Meanwhile, by a different IDR pattern of $\bfx_j^n$,
\begin{equation}
    \begin{aligned}
        \sum_{s=1}^{l+1}\text{softmax}({\bfp_s^n}^\top\bfW_K^\top\bfW_Q\bfp_{query}^n){\bfp_s^n}^\top({\bfx_j^n}^\top,\bfq^\top)^\top\lesssim  \beta^2 \gamma_t.\label{WV_part2}
    \end{aligned}
\end{equation}
When $t=0$, for all $i\in\mathcal{W}_n$, we have that by Lemma \ref{lemma: update_WU}, for $\bfp_{query}^n$ that corresponds to $\bfmu_a$,
\begin{equation}
    \bfW_{O_{(i,\cdot)}}^{(t)}\sum_{s=1}^{l+1}(\bfW_V\bfp_s^n)\text{softmax}({\bfp_s^n}^\top\bfW_K^\top\bfW_Q\bfp_{query}^n)>0.\label{indicator}
\end{equation}

\iffalse
\begin{equation}
\begin{aligned}
    &\bfW_{O_{(i,\cdot)}}^{(t)}\sum_{s=1}^{l+1}(\bfW_V^{(t)}\bfp_s^n)\text{softmax}({\bfp_s^n}^\top{\bfW_K^{(t)}}^\top\bfW_Q^{(t)}\bfp_{query}^n)\\
    \gtrsim & \delta\frac{1}{a}((1-2\gamma_t)-\gamma_t\cdot 1)\\
    \gtrsim & \delta\frac{1}{a}(1-3\gamma_t)>0,
\end{aligned}
\end{equation}
where $1-3\gamma_t>0$ when $t\geq 1$ after pre-training.
\fi
Therefore, for any $\bfp_j^n=({\bfx_j^n}^\top,{\bfy_j^n}^\top)^\top$ where $f^{(n)}(\tilde{\bfx}_j^n)=+1$, and 
\begin{equation}
    \bfx_j^n=\bfmu_a+\kappa_j^n\bfnu_b,
\end{equation}
we have
\begin{equation}
    \begin{aligned}
        &\eta\frac{1}{B}\sum_{n\in\mathcal{B}_b}\frac{\partial \ell(\tilde{\bfP}^n,z^n;\Psi)}{\partial \bfW_V^{(t)}}\bfp_j^n\\
        =& \eta\frac{1}{B}\sum_{n\in\mathcal{B}_b}(-z^n) \sum_{k=1}^m a_{k}\mathbbm{1}[\bfW_{O_{(k,\cdot)}}\sum_{s=1}^{l+1}(\bfW_V\bfp_s^n)\text{softmax}({\bfp_s^n}^\top\bfW_K^\top\bfW_Q\bfp_{query}^n)\geq0]\\
    &\cdot\bfW_{O_{(k,\cdot)}}^\top\sum_{s=1}^{l+1}\text{softmax}({\bfp_s^n}^\top\bfW_K^\top\bfW_Q\bfp_{query}^n){\bfp_s^n}^\top\bfp_j^n.
    \end{aligned}
\end{equation}
We then have that by combining (\ref{WV_part1}) and (\ref{WV_part2}),
\begin{equation}
    \begin{aligned}
        &-\eta\frac{1}{B}\sum_{n\in\mathcal{B}_b}(-z^n) \sum_{i\in\mathcal{W}_n} a_{i}\mathbbm{1}[\bfW_{O_{(i,\cdot)}}\sum_{s=1}^{l+1}(\bfW_V\bfp_s^n)\text{softmax}({\bfp_s^n}^\top\bfW_K^\top\bfW_Q\bfp_{query}^n)\geq0]\\
    &\cdot\sum_{s=1}^{l+1}\text{softmax}({\bfp_s^n}^\top\bfW_K^\top\bfW_Q\bfp_{query}^n){\bfp_s^n}^\top\bfp_j^n\\
    \gtrsim & \eta\beta^2(1-\gamma_t).
    \end{aligned}
\end{equation}
Since that for $\bfp_s^n$ and $\bfp_j^n$ with different label embeddings, their inner product is smaller than $-1+\beta$ if they share the same IDR pattern, or smaller than $-1$ if they share different IDR patterns. On average, in a batch, this product is close to $-1$. Hence
\begin{equation}
    \begin{aligned}
        &-\eta\frac{1}{B}\sum_{n\in\mathcal{B}_b}(-z^n) \sum_{i\in\mathcal{U}_n}^m a_{i}\mathbbm{1}[\bfW_{O_{(k,\cdot)}}\sum_{s=1}^{l+1}(\bfW_V\bfp_s^n)\text{softmax}({\bfp_s^n}^\top\bfW_K^\top\bfW_Q\bfp_{query}^n)\geq0]\\
    &\cdot\sum_{s=1}^{l+1}\text{softmax}({\bfp_s^n}^\top\bfW_K^\top\bfW_Q\bfp_{query}^n){\bfp_s^n}^\top\bfp_j^n\\
    \leq  & \frac{1}{\beta^2+1} \cdot (-\eta\frac{1}{B}\sum_{n\in\mathcal{B}_b}(-z^n) \sum_{i\in\mathcal{W}_n} a_{i}\mathbbm{1}[\bfW_{O_{(i,\cdot)}}\sum_{s=1}^{l+1}(\bfW_V\bfp_s^n)\text{softmax}({\bfp_s^n}^\top\bfW_K^\top\bfW_Q\bfp_{query}^n)\geq0]\\
    &\cdot\sum_{s=1}^{l+1}\text{softmax}({\bfp_s^n}^\top\bfW_K^\top\bfW_Q\bfp_{query}^n){\bfp_s^n}^\top\bfp_j^n).
    \end{aligned}
\end{equation}
Meanwhile, since that 
\begin{equation}
    \begin{aligned}
        &\eta\frac{1}{B}\sum_{n\in\mathcal{B}_b}(-z^n) \sum_{i\notin\mathcal{W}_n\cup\mathcal{U}_n}^m a_{i}\mathbbm{1}[\bfW_{O_{(i,\cdot)}}\sum_{s=1}^{l+1}(\bfW_V\bfp_s^n)\text{softmax}({\bfp_s^n}^\top\bfW_K^\top\bfW_Q\bfp_{query}^n)\geq0]\\
    &\cdot\sum_{s=1}^{l+1}\text{softmax}({\bfp_s^n}^\top\bfW_K^\top\bfW_Q\bfp_{query}^n){\bfp_s^n}^\top\bfp_j^n\\
    \lesssim  & \sqrt{\frac{\log B}{B}} \cdot\eta\frac{1}{B}\sum_{n\in\mathcal{B}_b}(-z^n) \sum_{j\in\mathcal{W}_n}^m a_{j}\mathbbm{1}[\bfW_{O_{(j,\cdot)}}\sum_{s=1}^{l+1}(\bfW_V\bfp_s^n)\text{softmax}({\bfp_s^n}^\top\bfW_K^\top\bfW_Q\bfp_{query}^n)\geq0]\\
    &\sum_{s=1}^{l+1}\text{softmax}({\bfp_s^n}^\top\bfW_K^\top\bfW_Q\bfp_{query}^n){\bfp_s^n}^\top\bfp_j^n,
    \end{aligned}
\end{equation}
where $\sqrt{\log B/B}$ is because that $z^n$ is selected from $\{+1,-1\}$ with equal probability. Hence, we can denote and derive that when $t=t_0+1$,
\begin{equation}
    \begin{aligned}
        &\eta\frac{1}{B}\sum_{n\in\mathcal{B}_b}\sum_{b=0}^{t_0}\frac{\partial \ell(\tilde{\bfP}^n,z^n;\Psi)}{\partial \bfW_V^{(b)}}\bfp_j^n\\
        = & \eta \sum_{b=0}^{t_0}(\sum_{i\in\mathcal{W}_n}V_i(b)\bfW_{O_{(i,\cdot)}}^{(b)}+\sum_{i\in\mathcal{U}_n}V_i(b)\bfW_{O_{(i,\cdot)}}^{(b)}+\sum_{i\notin\mathcal{W}_n\cup\mathcal{U}_n}V_i(b)\bfW_{O_{(i,\cdot)}}^{(b)}),
    \end{aligned}
\end{equation}
where 
\begin{equation}
    -V_i(b)\gtrsim \beta^2(1-\gamma_t)1/a,\ \ \ \ i\in\mathcal{W}_n,
\end{equation}
\begin{equation}
    -V_i(b)\leq \frac{1}{\beta^2+1}V_j(b),\ \ \ \ i\in\mathcal{U}_n, j\in\mathcal{W}_n,
\end{equation}
\begin{equation}
    |V_i(b)|\lesssim \sqrt{\frac{\log B}{B}}\cdot \frac{1}{a},\ \ \ \ i\notin\mathcal{W}_n\cup\mathcal{U}_n.
\end{equation}
\iffalse
We require that
\begin{equation}
    \eta t(1-2\gamma_t)(\beta^2-1)->0
\end{equation}
\fi
Similarly, for any $\bfp_j^n=({\bfx_j^n}^\top,{\bfy_j^n}^\top)^\top$ where $f^{(n)}(\tilde{\bfx}_j^n)=-1$,
\begin{equation}
    \bfx_j^n=\bfmu_a+\kappa_j^n\bfnu_b,
\end{equation}
we have
\begin{equation}
    \begin{aligned}
        &\eta\frac{1}{B}\sum_{n\in\mathcal{B}_b}\frac{\partial \ell(\tilde{\bfP}^n,z^n;\Psi)}{\partial \bfW_V^{(t)}}\bfp_j^n\\
        =& \eta\frac{1}{B}\sum_{n\in\mathcal{B}_b}(-z^n) \sum_{k=1}^m a_{k}\mathbbm{1}[\bfW_{O_{(k,\cdot)}}\sum_{s=1}^{l+1}(\bfW_V\bfp_s^n)\text{softmax}({\bfp_s^n}^\top\bfW_K^\top\bfW_Q\bfp_{query}^n)\geq0]\\
    &\cdot\bfW_{O_{(k,\cdot)}}^\top\sum_{s=1}^{l+1}\text{softmax}({\bfp_s^n}^\top\bfW_K^\top\bfW_Q\bfp_{query}^n){\bfp_s^n}^\top\bfp_j^n,
    \end{aligned}
\end{equation}
\begin{equation}
    \begin{aligned}
        &\eta\frac{1}{B}\sum_{n\in\mathcal{B}_b}\sum_{b=0}^{t_0}\frac{\partial \ell(\tilde{\bfP}^n,z^n;\Psi)}{\partial \bfW_V^{(b)}}\bfp_j^n\\
        = & \eta \sum_{b=0}^{t_0}(\sum_{i\in\mathcal{W}_n}V_i(b)\bfW_{O_{(i,\cdot)}}^{(b)}+\sum_{i\in\mathcal{U}_n}V_i(b)\bfW_{O_{(i,\cdot)}}^{(b)}+\sum_{i\notin\mathcal{W}\cup\mathcal{U}}V_i(b)\bfW_{O_{(i,\cdot)}}^{(b)}),
    \end{aligned}
\end{equation}
where 
\begin{equation}
    -V_i(b)\gtrsim \beta^2(1-\gamma_t)1/a,\ \ \ \ i\in\mathcal{U}_n,
\end{equation}
\begin{equation}
    -V_i(b)\leq \frac{1}{\beta^2+1}V_j(b),\ \ \ \ i\in\mathcal{W}_n, j\in\mathcal{U}_n,
\end{equation}
\begin{equation}
    |V_i(b)|\lesssim \sqrt{\frac{\log B}{B}}\cdot \frac{1}{a},\ \ \ \ i\notin\mathcal{W}_n\cup\mathcal{U}_n.
\end{equation}

We can also derive
\begin{equation}
    \begin{aligned}
        &\eta\frac{1}{B}\sum_{b=0}^{t_0}\sum_{n\in\mathcal{B}_b}\frac{\partial \ell(\tilde{\bfP}^n,z^n;\Psi)}{\partial \bfW_V^{(t)}}(\bfv_k^\top,\boldsymbol{0}^\top)^\top\\
        =& \eta\frac{1}{B}\sum_{b=0}^{t_0}\sum_{n\in\mathcal{B}_b}(-z^n) \sum_{k=1}^m a_{k}\mathbbm{1}[\bfW_{O_{(k,\cdot)}}\sum_{s=1}^{l+1}(\bfW_V\bfp_s^n)\text{softmax}({\bfp_s^n}^\top\bfW_K^\top\bfW_Q\bfp_{query}^n)\geq0]\\
    &\cdot\bfW_{O_{(k,\cdot)}}^\top\sum_{s=1}^{l+1}\text{softmax}({\bfp_s^n}^\top\bfW_K^\top\bfW_Q\bfp_{query}^n){\bfp_s^n}^\top(\bfnu_k^\top,\boldsymbol{0}^\top)^\top\\
    := & \eta \sum_{b=0}^{t_0}(\sum_{i\in\mathcal{W}_n}V_i'(b)\bfW_{O_{(i,\cdot)}}^{(b)}+\sum_{i\in\mathcal{U}_n}V_i'(b)\bfW_{O_{(i,\cdot)}}^{(b)}+\sum_{i\notin\mathcal{W}\cup\mathcal{U}}V_i'(b)\bfW_{O_{(i,\cdot)}}^{(b)}),
    \end{aligned}
\end{equation}
where
\begin{equation}
    |V_i'(b)|\leq |V_i(b)|\cdot \frac{1}{M_2},
\end{equation}
since that $1/M_2$ fraction of $\bfp_s^n$ has $\bfnu_K$ as the IDI pattern in average.

\end{proof}

\subsection{Proof of Lemma \ref{lemma: O}}
\begin{proof}
\begin{equation}
\begin{aligned}
    &\quad\eta\frac{1}{B}\sum_{n\in\mathcal{B}_b}\frac{\partial \ell(\tilde{\bfP}^n,z^n;\Psi)}{\partial \bfW_{O_{(i,\cdot)}}}\\
    &=\eta\frac{1}{B}\sum_{n\in\mathcal{B}_b} \frac{\partial \ell(\tilde{\bfP}^n,z^n;\Psi)}{\partial F(\bfp_{query}^n)}\frac{\partial F(\bfp_{query}^n)}{\partial \bfW_{O_{(i,\cdot)}}}\\
    &=\eta\frac{1}{B}\sum_{n\in\mathcal{B}_b}(-z^n)  a_{i}\mathbbm{1}[\bfW_{O_{(i,\cdot)}}\sum_{s=1}^{l+1}(\bfW_V\bfp_s^n)\text{softmax}({\bfp_s^n}^\top\bfW_K^\top\bfW_Q\bfp_{query}^n)\geq0]\\
    &\cdot\sum_{s=1}^{l+1} (\bfW_V\bfp_s^n)\text{softmax}({\bfp_s^n}^\top\bfW_K^\top\bfW_Q\bfp_{query}^n).\label{O grad}
\end{aligned}
\end{equation}
We have that
\begin{equation}
\begin{aligned}
    &\bfW_V^{(t)}\bfp_s^n\\
    =&\delta({\bfp_s^n}^\top, \boldsymbol{0}^\top)^\top+\sum_{b=0}^{t-1}\eta(\sum_{i\in\mathcal{W}_n}V_i(b)\bfW_{O_{(i,\cdot)}}^{(b)}+\sum_{i\in\mathcal{U}_n}V_i(b)\bfW_{O_{(i,\cdot)}}^{(b)}+\sum_{i\notin\mathcal{W}_n\cup\mathcal{U}_n}V_i(b)\bfW_{O_{(i,\cdot)}}^{(b)})^\top.\label{WvP_o_grad}
\end{aligned}
\end{equation}
Consider a certain $\bfp_s^n=({\bfx_s^n}^\top,{\bfy_s^n}^\top,\boldsymbol{0}^\top)^\top$ where $f^{(n)}(\tilde{\bfx}_s^n)=+1$, and
\begin{equation}
    \bfx_s^n=\bfmu_a+\kappa_s^n\bfnu_b.
\end{equation}
When $t=0$, we can obtain that for $i\in\mathcal{W}_n$ and $\bfmu_a$ as the IDR pattern of $\bfp_{query}^n$,
\begin{equation}
    \begin{aligned}
        &\eta\frac{1}{B}\sum_{n\in\mathcal{B}_b}\frac{\partial \ell(\tilde{\bfP}^n,z^n;\Psi)}{\partial \bfW_{O_{(i,\cdot)}}}(\bfmu_a^\top,\bfq^\top)^\top\\
        =&\eta\frac{1}{B}\sum_{n\in\mathcal{B}_b}\frac{1}{a}\sum_{s=1}^{l+1}\text{softmax}({\bfp_s^n}^\top\bfW_K^\top\bfW_Q\bfp_{query}^n)(\delta{\bfp_s^n}^\top(\bfmu_a^\top,\bfq^\top)^\top+\sum_{b=0}^{t-1}\eta(\sum_{i\in\mathcal{W}_n}V_i(b)\bfW_{O_{(i,\cdot)}}^{(b)}\\
        &+\sum_{i\in\mathcal{U}_n}V_i(b)\bfW_{O_{(i,\cdot)}}^{(b)}+\sum_{i\notin\mathcal{W}_n\cup\mathcal{U}_n}V_i(b)\bfW_{O_{(i,\cdot)}}^{(b)})^\top(\bfmu_a^\top,\bfq^\top)^\top)\\
        \geq & \frac{\alpha\eta}{2a}\delta(\beta^2+1).\\
        \end{aligned}
\end{equation}
Then, we have the following results by Lemma \ref{lemma: V}, and the magnitude of $\bfmu_a$, $\bfmu_b$, and $\bfq$, 
\begin{equation}
    \begin{aligned}
        &\eta\frac{1}{B}\sum_{n\in\mathcal{B}_b}\frac{\partial \ell(\tilde{\bfP}^n,z^n;\Psi)}{\partial \bfW_{O_{(i,\cdot)}}}(\bfmu_a^\top,-\bfq^\top)^\top
        \leq  \frac{\beta^2-1}{\beta^2+1}\eta\frac{1}{B}\sum_{n\in\mathcal{B}_b}\frac{\partial \ell(\tilde{\bfP}^n,z^n;\Psi)}{\partial \bfW_{O_{(i,\cdot)}}}(\bfmu_a^\top,\bfq^\top)^\top,\\
        \end{aligned}
\end{equation}
\begin{equation}
    \begin{aligned}
        &\eta\frac{1}{B}\sum_{n\in\mathcal{B}_b}\frac{\partial \ell(\tilde{\bfP}^n,z^n;\Psi)}{\partial \bfW_{O_{(i,\cdot)}}}(\bfmu_b^\top,\bfq^\top)^\top
        \leq  \frac{1}{\beta^2+1}\eta\frac{1}{B}\sum_{n\in\mathcal{B}_b}\frac{\partial \ell(\tilde{\bfP}^n,z^n;\Psi)}{\partial \bfW_{O_{(i,\cdot)}}}(\bfmu_a^\top,\bfq^\top)^\top,\\
        \end{aligned}
\end{equation}
\begin{equation}
    \begin{aligned}
        &\eta\frac{1}{B}\sum_{n\in\mathcal{B}_b}\frac{\partial \ell(\tilde{\bfP}^n,z^n;\Psi)}{\partial \bfW_{O_{(i,\cdot)}}}(\bfmu_b^\top,-\bfq^\top)^\top
        \leq  -\frac{1}{\beta^2+1}\eta\frac{1}{B}\sum_{n\in\mathcal{B}_b}\frac{\partial \ell(\tilde{\bfP}^n,z^n;\Psi)}{\partial \bfW_{O_{(i,\cdot)}}}(\bfmu_a^\top,\bfq^\top)^\top,\\
        \end{aligned}
\end{equation}
\begin{equation}
    \begin{aligned}
        &\eta\frac{1}{B}\sum_{n\in\mathcal{B}_b}\frac{\partial \ell(\tilde{\bfP}^n,z^n;\Psi)}{\partial \bfW_{O_{(i,\cdot)}}}(\bfnu_c^\top,\boldsymbol{0}^\top)^\top
        \leq  \frac{1}{M_2}\eta\frac{1}{B}\sum_{n\in\mathcal{B}_b}\frac{\partial \ell(\tilde{\bfP}^n,z^n;\Psi)}{\partial \bfW_{O_{(i,\cdot)}}}(\bfmu_a^\top,\bfq^\top)^\top.\\
        \end{aligned}
\end{equation}
Denote the set of data that share one same IDR pattern as $\bfp_{query}^n$ as $\mathcal{B}_b^n$ in the $b$-th iteration. Therefore, when $t=1$, we have
\begin{equation}
    \begin{aligned}
        &\quad\eta\frac{1}{|\mathcal{B}_b^n|}\sum_{n\in\mathcal{B}_b^n}\frac{\partial \ell(\tilde{\bfP}^n,z^n;\Psi)}{\partial \bfW_{O_{(i,\cdot)}}}(\bfmu_a^\top,\bfq^\top)^\top\\
        &=\eta\frac{1}{B}\sum_{n\in\mathcal{B}_b}\frac{1}{a}\sum_{s=1}^{l+1}\text{softmax}({\bfp_s^n}^\top\bfW_K^\top\bfW_Q\bfp_{query}^n)(\delta{\bfp_s^n}^\top(\bfmu_a^\top,\bfq^\top)^\top+\sum_{b=0}^{t-1}\eta(\sum_{i\in\mathcal{W}_n}V_i(b)\bfW_{O_{(i,\cdot)}}^{(b)}\\
        &\quad +\sum_{i\in\mathcal{U}_n}V_i(b)\bfW_{O_{(i,\cdot)}}^{(b)}+\sum_{i\notin\mathcal{W}_n\cup\mathcal{U}_n}V_i(b)\bfW_{O_{(i,\cdot)}}^{(b)})^\top(\bfmu_a^\top,\bfq^\top)^\top\\
       & \geq  \frac{\alpha}{2}(\frac{\eta}{a}(\delta(\beta^2+1)-\frac{\eta}{a}\cdot\eta \cdot\sqrt{\frac{\log B}{B}}\frac{m}{a}\cdot\xi\log M_1\\
        &\quad +\frac{\eta}{a}\cdot\eta \frac{m}{a}(\beta^2(1-\gamma_t))(\frac{\eta }{a}\delta(\beta^2+1)-\xi))\\
        & \gtrsim  \frac{\alpha}{2}(\frac{\eta}{a}(\delta(\beta^2+1)+\frac{\eta}{a}\cdot\eta \frac{m}{a}(\beta^2(1-\gamma_t))\frac{\eta }{a}\delta(\beta^2+1))\\
        & \gtrsim  \delta(\beta^2+1)\frac{\alpha\eta}{2a}(1+\frac{\eta^2 m}{a^2}),
    \end{aligned}
\end{equation}
%as long as $\log M_1\geq \delta^21$ and $B\geq \beta^{-4}$. 
where the first inequality comes from that the update in the previous step makes the output of $\bfW_{O_{(i,\cdot)}}^{(b)}$ for $i\in\mathcal{W}_n$ be positive. The second step holds when $B\gtrsim M_1$. We also have
\begin{equation}
    \begin{aligned}
        &\quad \eta\frac{1}{|\mathcal{B}_b-\mathcal{B}_b^n|}\sum_{n\in\mathcal{B}_b-\mathcal{B}_b^n}\frac{\partial \ell(\tilde{\bfP}^n,z^n;\Psi)}{\partial \bfW_{O_{(i,\cdot)}}}(\bfmu_a^\top,\bfq^\top)^\top\\
        &\gtrsim  \delta(\beta^2+1)\frac{\alpha\eta}{2a}(1+\frac{\eta^2 m}{a^2}).\label{mu_b_t1}
    \end{aligned}
\end{equation}
For $i\in \mathcal{U}_n$, we also have
\begin{equation}
    \begin{aligned}
        &\quad \eta\frac{1}{|\mathcal{B}_b^n|}\sum_{n\in\mathcal{B}_b^n}\frac{\partial \ell(\tilde{\bfP}^n,z^n;\Psi)}{\partial \bfW_{O_{(i,\cdot)}}}(\bfmu_a^\top,-\bfq^\top)^\top\\
        &\gtrsim  \delta(\beta^2+1)\frac{\alpha\eta}{2a}(1+\frac{\eta^2 m}{a^2}),
    \end{aligned}
\end{equation}
\begin{equation}
    \begin{aligned}
        &\quad \eta\frac{1}{|\mathcal{B}_b-\mathcal{B}_b^n|}\sum_{n\in\mathcal{B}_b-\mathcal{B}_b^n}\frac{\partial \ell(\tilde{\bfP}^n,z^n;\Psi)}{\partial \bfW_{O_{(i,\cdot)}}}(\bfmu_a^\top,-\bfq^\top)^\top\\
        &\gtrsim  \delta(\beta^2+1)\frac{\alpha\eta}{2a}(1+\frac{\eta^2 m}{a^2}),
    \end{aligned}
\end{equation}
if $\bfp_j^n$ corresponds to label $-1$ in this task. For $i\notin\mathcal{W}_n(t)\cup\mathcal{U}_n(t)$, we have
\begin{equation}
    \begin{aligned}
        &\eta\frac{1}{B}\sum_{n\in\mathcal{B}_b}\frac{\partial \ell(\tilde{\bfP}^n,z^n;\Psi)}{\partial \bfW_{O_{(i,\cdot)}}^{(t)}}({\bfp_j^n}^\top,\boldsymbol{0})^\top \leq \eta \sqrt{\frac{\log B}{B}}\frac{1}{a}.
    \end{aligned}
\end{equation}

\noindent Suppose that the conclusion holds when $t\leq t_0$. Then when $t=t_0+1$, we have that for $i\in\mathcal{W}_n$, $b\neq a$, and $\bfp_{query}^n$ corresponding to $\bfq$ and $\bfmu_a$, 
\begin{equation}
\begin{aligned}
    &\quad \eta\frac{1}{|\mathcal{B}_b^n|}\sum_{n\in\mathcal{B}_b^n}\frac{\partial \ell(\tilde{\bfP}^n,z^n;\Psi)}{\partial \bfW_{O_{(i,\cdot)}}^{(t)}}(\bfmu_a^\top,\bfq^\top)^\top\\
   & =  \eta\frac{1}{B}\sum_{n\in\mathcal{B}_b}\frac{1}{a}\sum_{s=1}^{l+1}\text{softmax}({\bfp_s^n}^\top\bfW_K^\top\bfW_Q\bfp_{query}^n)(\delta{\bfp_s^n}^\top(\bfmu_a^\top,\bfq^\top)^\top+\sum_{b=0}^{t-1}\eta(\sum_{i\in\mathcal{W}_n}V_i(b)\bfW_{O_{(i,\cdot)}}^{(b)}\\
    &\quad +\sum_{i\in\mathcal{U}_n}V_i(b)\bfW_{O_{(i,\cdot)}}^{(b)}+\sum_{i\notin\mathcal{W}_n\cup\mathcal{U}_n}V_i(b)\bfW_{O_{(i,\cdot)}}^{(b)})^\top(\bfmu_a^\top,\bfq^\top)^\top\\
    &\gtrsim  \delta(\beta^2+1)\frac{\alpha\eta}{2a}+\frac{\alpha\eta }{2a}\cdot\eta\sum_{b=0}^{t_0}\delta(\beta^2+1)\frac{\eta m}{a^2}(1+\frac{\eta^2 m}{a^2})^{b}\\
   & =  \delta(\beta^2+1)\frac{\alpha\eta}{2a}(1+\frac{\eta^2 m}{a^2}\cdot\frac{(1+\frac{\eta^2 m}{a^2})^{t_0+1}-1}{\frac{\eta^2 m}{a^2}})\\
   & =  \delta(\beta^2+1)\frac{\alpha\eta}{2a}(1+\frac{\eta^2 m}{a^2})^{t_0+1},
\end{aligned}
\end{equation}
where the first inequality is by plugging the condition in the induction. The last two steps come from basic mathematical computation. Then,
\begin{equation}
\begin{aligned}
    &\eta\frac{1}{|\mathcal{B}_b^n|}\sum_{b=0}^{t_0+1}\sum_{n\in\mathcal{B}_b^n}\frac{\partial \ell(\tilde{\bfP}^n,z^n;\Psi)}{\partial \bfW_{O_{(i,\cdot)}}^{(t)}}(\bfmu_a^\top,\bfq^\top)^\top\\
    \gtrsim & \delta(\beta^2+1)\sum_{b=0}^{t_0+1}\frac{\alpha\eta}{2a}(1+\frac{\eta^2 m}{a^2})^b\\
    \gtrsim & \delta(\beta^2+1)\frac{\alpha\eta}{2a}(t_0+1),
\end{aligned}
\end{equation}
where lower bound in the last step is also a tight estimation of the second to last step if $\eta^2 T m/a^2\ll1$. Then, we have
\iffalse
\begin{equation}
    \begin{aligned}
        &\eta\frac{1}{B}\sum_{b=0}^{t_0+1}\sum_{n\in\mathcal{B}_b^n}\frac{\partial \ell(\tilde{\bfP}^n,z^n;\Psi)}{\partial \bfW_{O_{(i,\cdot)}}}(\bfmu_a^\top,\bfq^\top)^\top\\
        \gtrsim & \frac{1}{M_1}\cdot(\frac{\eta}{a}\sum_{b=0}^{t_0+1}(1-\gamma_b)\delta(\beta^2+1)+\frac{\eta}{a}\sum_{b=0}^{t_0+1}\eta m\sum_{c=0}^b\frac{\eta}{a}\sum_{e=0}^{c}(1-\gamma_e)\delta(\beta^2+1)),
    \end{aligned}
\end{equation}
\fi
\begin{equation}
    \begin{aligned}
        &\eta\frac{1}{B}\sum_{b=0}^{t_0+1}\sum_{n\in\mathcal{B}_b^n}\frac{\partial \ell(\tilde{\bfP}^n,z^n;\Psi)}{\partial \bfW_{O_{(i,\cdot)}}}(\bfmu_a^\top,\bfq^\top)^\top\\
        \gtrsim & \frac{1}{M_1}\cdot\delta\beta^2(\beta^2+1)\frac{\alpha\eta}{2a}(t_0+1).\label{WO mu_a q 1}
    \end{aligned}
\end{equation}
%Since $i\in\mathcal{W}_n$ can also give a positive lower bound by (\ref{mu_b_t1}), we know that the component of $(\bfmu_b^\top,\bfq^\top)^\top$ also grows in $\bfW_{O_{(i,\cdot)}}$ for $\bfmu_b$ that is not the IDR pattern of $\bfp_{query}^n$. 
By Lemma \ref{lemma: update_WU}, when $t\geq \Theta(1)$, we have
\begin{equation}
    \begin{aligned}
        &\eta\frac{1}{B}\sum_{b=0}^{t_0+1}\sum_{n\in\mathcal{B}_b-\mathcal{B}_b^n}\frac{\partial \ell(\tilde{\bfP}^n,z^n;\Psi)}{\partial \bfW_{O_{(i,\cdot)}}}(\bfmu_b^\top,\bfq^\top)^\top\\
        \gtrsim & \frac{M_1-1}{M_1}\cdot\delta(\beta^2+1)\frac{\alpha\eta}{2a}(t_0+1).\label{WO mu_a q 2}
    \end{aligned}
\end{equation}
Hence, 
\begin{equation}
    \begin{aligned}
        &\eta\frac{1}{B}\sum_{b=0}^{t_0+1}\sum_{n\in\mathcal{B}_b}\frac{\partial \ell(\tilde{\bfP}^n,z^n;\Psi)}{\partial \bfW_{O_{(i,\cdot)}}}(\bfmu_b^\top,\bfq^\top)^\top\\
        \gtrsim & \delta(\beta^2+1)\frac{\alpha\eta}{2a}(t_0+1),\label{WO_p}
    \end{aligned}
\end{equation}
which holds for $i\in\cup_{n\in[N]}\mathcal{W}_n=\mathcal{W}$. Meanwhile,
\begin{equation}
    \begin{aligned}
        &\eta\frac{1}{B}\sum_{b=0}^{t_0+1}\sum_{n\in\mathcal{B}_b}\frac{\partial \ell(\tilde{\bfP}^n,z^n;\Psi)}{\partial \bfW_{O_{(i,\cdot)}}}(\bfnu_c^\top,\boldsymbol{0}^\top)^\top
        \leq  \frac{1}{BT}\eta\frac{1}{B}\sum_{b=0}^{t_0+1}\sum_{n\in\mathcal{B}_b}\frac{\partial \ell(\tilde{\bfP}^n,z^n;\Psi)}{\partial \bfW_{O_{(i,\cdot)}}}(\bfmu_b^\top,\bfq^\top)^\top.
        \end{aligned}
\end{equation}
\noindent For $i\in\mathcal{U}_n$ and $\bfp_{query}^n$ corresponding to $-\bfq$ and $\bfmu_a$, similarly to (\ref{WO mu_a q 1}), (\ref{WO mu_a q 2}), and (\ref{WO_p}), we have
\begin{equation}
    \begin{aligned}
        &\eta\frac{1}{B}\sum_{b=0}^{t_0+1}\sum_{n\in\mathcal{B}_b^n}\frac{\partial \ell(\tilde{\bfP}^n,z^n;\Psi)}{\partial \bfW_{O_{(i,\cdot)}}}(\bfmu_a^\top,-\bfq^\top)^\top\\
        \gtrsim & \frac{1}{M_1}\cdot\delta(\beta^2+1)\frac{\alpha\eta}{2a}(t_0+1),
    \end{aligned}
\end{equation}
and when $t\geq \Theta(1)$,
\begin{equation}
    \begin{aligned}
        &\eta\frac{1}{B}\sum_{b=0}^{t_0+1}\sum_{n\in\mathcal{B}_b-\mathcal{B}_b^n}\frac{\partial \ell(\tilde{\bfP}^n,z^n;\Psi)}{\partial \bfW_{O_{(i,\cdot)}}}(\bfmu_b^\top,-\bfq^\top)^\top\\
        \gtrsim & \frac{M_1-1}{M_1}\delta(\beta^2+1)\frac{\alpha\eta}{2a}(t_0+1),
    \end{aligned}
\end{equation}
\begin{equation}
    \begin{aligned}
        &\eta\frac{1}{B}\sum_{b=0}^{t_0+1}\sum_{n\in\mathcal{B}_b}\frac{\partial \ell(\tilde{\bfP}^n,z^n;\Psi)}{\partial \bfW_{O_{(i,\cdot)}}}(\bfmu_b^\top,-\bfq^\top)^\top\\
        \gtrsim & \delta(\beta^2+1)\frac{\alpha\eta}{2a}(t_0+1),
    \end{aligned}
\end{equation}
which also holds for $i\in\cup_{n\in[N]}\mathcal{U}_n=\mathcal{U}$. Meanwhile,
\begin{equation}
    \begin{aligned}
        &\eta\frac{1}{B}\sum_{b=0}^{t_0+1}\sum_{n\in\mathcal{B}_b}\frac{\partial \ell(\tilde{\bfP}^n,z^n;\Psi)}{\partial \bfW_{O_{(i,\cdot)}}}(\bfnu_c^\top,\pm\bfq^\top)^\top
        \leq  \frac{1}{M_2}\eta\frac{1}{B}\sum_{b=0}^{t_0+1}\sum_{n\in\mathcal{B}_b}\frac{\partial \ell(\tilde{\bfP}^n,z^n;\Psi)}{\partial \bfW_{O_{(i,\cdot)}}}(\bfmu_b^\top,\bfq^\top)^\top.
        \end{aligned}
\end{equation}
\noindent Then, for $i\in\mathcal{W}_n\cup\mathcal{U}_n$,
\begin{equation}
    \begin{aligned}
    \|\bfW_{O_{(i,\cdot)}}^{(t_0+1)}\|
    \gtrsim \sqrt{M_1}\delta(\beta^2+1)^\frac{1}{2}\frac{\alpha\eta}{2a}(t_0+1).\label{WO_norm}
    \end{aligned}
\end{equation}
For $i\notin\mathcal{W}\cup\mathcal{U}$, we have
\begin{equation}
        \eta\frac{1}{B}\sum_{b=0}^{t_0+1}\sum_{n\in\mathcal{B}_b}\frac{\partial \ell(\tilde{\bfP}^n,z^n;\Psi)}{\partial \bfW_{O_{(i,\cdot)}}}(\bfmu_a^\top,-\bfq^\top)^\top\leq \eta \sqrt{\frac{\log B(t_0+1)}{B(t_0+1)}}\frac{1}{a}.\label{WO_lower}
\end{equation}
%By the derivation of (\ref{indicator_final}), we have that (\ref{WO_p}), (\ref{WO_norm}), and (\ref{WO_lower}) holds for $i\in\mathcal{W}$. 
\end{proof}

\subsection{Proof of Lemma \ref{lemma: initial_WU}}
\begin{proof}
    We know that the Gaussian initialization of $\bfW_{O_{(i,\cdot)}}^{(0)}$ generates a uniform distribution on the $d_\mathcal{X}-1$-sphere for the first $d_\mathcal{X}$ dimensions. Therefore,
\begin{equation}
    \Pr(i\in\mathcal{W}_n)=A_{d_\mathcal{X}}^{cap}(\phi)/A_{d_\mathcal{X}},
\end{equation}
where $A_{d_\mathcal{X}}$ is the surface area of an $d_\mathcal{X}-1$-sphere. $A_{d_\mathcal{X}}^{cap}(\phi)$ is the surface area of a $d_\mathcal{X}-1$-spherical cap with $\phi$ as the colatitude angle. By Equation 1 in \citep{L10}, we have
\begin{equation}
    \Pr(i\in\mathcal{W}_n)=\frac{1}{2}I_{\sin^2{\phi}}(\frac{d_\mathcal{X}-1}{2}, \frac{1}{2})=\frac{\int_{0}^{\sin^2{\phi}}t^{\frac{d_\mathcal{X}-3}{2}}(1-t)^{-\frac{1}{2}}dt}{2\int_{0}^{1}t^{\frac{d_\mathcal{X}-3}{2}}(1-t)^{-\frac{1}{2}}dt},
\end{equation}
where $I_{\cdot}(\cdot,\cdot)$ is the regularized incomplete beta function. Since that 
\begin{equation}
    \phi\leq \pi/2-\Theta(1/M_1),
\end{equation}
to avoid concentration error of $\Theta(\sqrt{1/m})$ if $m\gtrsim M_1^2$, we have that when $d_\mathcal{X}=M_1+M_2=M=\Theta(M)$,
\begin{equation}
    \begin{aligned}
        &\frac{\int_{0}^{\sin^2{\phi}}t^{\frac{d_\mathcal{X}-3}{2}}(1-t)^{-\frac{1}{2}}dt}{\int_{0}^{1}t^{\frac{d_\mathcal{X}-3}{2}}(1-t)^{-\frac{1}{2}}dt}\\
        \geq & \frac{\int_{0}^{\cos^2{1/M}}t^{\frac{d_\mathcal{X}-3}{2}}(1-t)^{-\frac{1}{2}}dt}{\int_{0}^{1}t^{\frac{d_\mathcal{X}-3}{2}}(1-t)^{-\frac{1}{2}}dt}\\
        \geq & 1-\frac{\int_{1-1/M^2}^{1}t^{\frac{d_\mathcal{X}-3}{2}}(1-t)^{-\frac{1}{2}}dt}{\int_{0}^{1}t^{\frac{d_\mathcal{X}-3}{2}}(1-t)^{-\frac{1}{2}}dt}\\
        \geq & 1-\frac{\int_{1-1/M^2}^{1}(1-t)^{-\frac{1}{2}}dt}{\int_{1-1/M}^{1}\Theta(1)\cdot(1-t)^{-\frac{1}{2}}dt}\\
        =& 1-\frac{\frac{2}{M}}{\Theta(1)\cdot(\frac{2}{\sqrt{M}}-\frac{2}{M})}\\
        \geq & \Theta(1),\label{initial_first_dx}
    \end{aligned}
\end{equation}
where the second inequality comes from that $\cos^2(1/M)=(1+\cos(2/M))/2\geq 1-1/M^2\geq 1-1/M$, and the third to last step is by $(1-1/M)^{\frac{d_\mathcal{X}-3}{2}}\geq \Theta(1)$, and the last step is by $M\geq \Theta(1)$. For the second $d_\mathcal{Y}$ dimensions of $\bfW_{O_{(i,\cdot)}}^{(0)}$, we can derive a similar result by replacing $d_\mathcal{X}$ with $d_\mathcal{Y}$ in (\ref{initial_first_dx}). This implies that
\begin{equation}
    |\mathcal{W}_n|\geq \Omega(1)\cdot \Omega(1)\cdot m\geq \Omega(m).
\end{equation}
Likewise, the conclusion holds for $\mathcal{U}_n$. Since that $\mathcal{W}=\cup_{n\in[N}\mathcal{W}_n$, $\mathcal{U}=\cup_{n\in[N}\mathcal{U}_n$, we have
\begin{equation}
    |\mathcal{W}|, |\mathcal{U}|\geq \Omega(m).
\end{equation}

\end{proof}

\subsection{Proof of Lemma \ref{lemma: update_WU}}
\begin{proof}
We prove this lemma in two steps. In the first step, we prove the conclusion by replacing $\mathcal{W}$ with $\mathcal{W}_n$, and replacing $\mathcal{U}$ with $\mathcal{U}_n$. We will also cover the proof of $\bfW_{O_{(i,\cdot)}}^{(0)}{\bfW_{O_{(i,\cdot)}}^{(t)}}^\top>0$ and $\sum_{b=0}^{t-1}\eta \sum_{i\in\mathcal{W}_n}V_i(b)\bfW_{O_{(i,\cdot)}}^{(b)}{\bfW_{O_{i,\cdot}}^{(t)}}^\top>0$ in the induction as a support. In the second step, we prove the results for $\mathcal{W}$ and $\mathcal{U}$. \\
\noindent (1) When $t=0$. For any $i\in\mathcal{W}_n$, we have that by definition of $\mathcal{W}_n$
\begin{equation}
    \bfW_{O_{(i,\cdot)}}^{(0)}\bfV^n(0)>0,
\end{equation}
\begin{equation}
    \bfW_{O_{(i,\cdot)}}^{(0)}{\bfW_{O_{i,\cdot}}^{(t)}}^\top>0.
\end{equation}
Hence, the conclusion holds. When $t=1$, we have
\begin{equation}
    \begin{aligned}
    &\bfW_{O_{(i,\cdot)}}^{(t)}\bfV^n(t)\\
        =&\bfW_{O_{(i,\cdot)}}^{(0)}{\bfW_{O_{(i,\cdot)}}^{(t)}}^\top+\eta\frac{1}{B}\sum_{n\in\mathcal{B}_b}\frac{1}{a}\sum_{s=1}^{l+1}\text{softmax}({\bfp_s^n}^\top{\bfW_K^{(t)}}^\top\bfW_Q^{(t)}\bfp_{query}^n)(\delta{\bfp_s^n}^\top{\bfW_{O_{(i,\cdot)}}^{(t)}}^\top+\sum_{b=0}^{t-1}\eta(\sum_{i\in\mathcal{W}_n}V_i(b)\bfW_{O_{(i,\cdot)}}^{(b)}\\
        &+\sum_{i\in\mathcal{U}_n}V_i(b)\bfW_{O_{(i,\cdot)}}^{(b)}+\sum_{i\notin\mathcal{W}_n\cup\mathcal{U}_n}V_i(b)\bfW_{O_{(i,\cdot)}}^{(b)})^\top{\bfW_{O_{(i,\cdot)}}^{(t)}}^\top).\label{WU_update_to_prove}
    \end{aligned}
\end{equation}
By (\ref{O grad}) and definition of $\bfW_n$, we have
\begin{equation}
\begin{aligned}
    &\bfW_{O_{(i,\cdot)}}^{(0)}\eta\frac{1}{B}\sum_{n\in\mathcal{B}_b}(+z^n)  a_{i}\mathbbm{1}[\bfW_{O_{(i,\cdot)}}\sum_{s=1}^{l+1}(\bfW_V\bfp_s^n)\text{softmax}({\bfp_s^n}^\top\bfW_K^\top\bfW_Q\bfp_{query}^n)\geq0]\\
    &\cdot\sum_{s=1}^{l+1} (\bfW_V^{(t-1)}\bfp_s^n)\text{softmax}({\bfp_s^n}^\top{\bfW_K^{(t-1)}}^\top\bfW_Q^{(t-1)}\bfp_{query}^n)\\
    >&0.
\end{aligned}
\end{equation}
Hence,
\begin{equation}
    \bfW_{O_{(i,\cdot)}}^{(0)}{\bfW_{O_{(i,\cdot)}}^{(t)}}^\top=\sum_{b=0}^{t-1}\bfW_{O_{(i,\cdot)}}^{(b)}{\bfW_{O_{i,\cdot}}^{(t)}}^\top>0,
\end{equation}
and
\begin{equation}
    \sum_{b=0}^{t-1}\eta \sum_{i\in\mathcal{W}_n}V_i(b)\bfW_{O_{(i,\cdot)}}^{(b)}{\bfW_{O_{i,\cdot}}^{(t)}}^\top>0.
\end{equation}
By the gradient update when $t=0$, we know that the largest component in the feature embedding is the IDR pattern for $\bfp_{query}^n$, and the label embedding is close to being in the direction of the label embedding of $\bfp_{query}^n$. Hence,
\begin{equation}
    \eta\frac{1}{B}\sum_{n\in\mathcal{B}_b}\frac{1}{a}\sum_{s=1}^{l+1}\text{softmax}({\bfp_s^n}^\top{\bfW_K^{(t)}}^\top\bfW_Q^{(t)}\bfp_{query}^n)\delta{\bfp_s^n}^\top{\bfW_{O_{(i,\cdot)}}^{(t)}}^\top>0.\label{O_P}
\end{equation}
Denote $\theta_n^i$ as the angle between the feature embeddings of $\bfV^n(0)$ and $\bfW_{O_{(i,\cdot)}}^{(0)}$. Since that the feature embedding of $\bfW_{O_{(i,\cdot)}}^{(0)}$ is initialized uniformed on the $d_\mathcal{X}-1$-sphere, we have $\mathbb{E}[\theta_n^i]=0$. By Hoeffding's inequality (\ref{hoeffding}), we have
\begin{equation}
    \Big\|\frac{1}{|\mathcal{W}_n|}\sum_{i\in\mathcal{W}_n}\theta_n^i-\mathbb{E}[\theta_n^i]\Big\|=\Big\|\frac{1}{|\mathcal{W}_n|}\sum_{i\in\mathcal{W}_n}\theta_n^i\Big\|\leq \sqrt{\frac{\log M_1}{m}},\label{concentration_theta}
\end{equation}
with probability of at least $1-M_1^{-10}$. When $m\gtrsim M^2\log M_1$, we can obtain that 
\begin{equation}
    \Big\|\frac{1}{|\mathcal{W}_n|}\sum_{i\in\mathcal{W}_n}\theta_n^i-\mathbb{E}[\theta_n^i]\Big\|\leq \Theta(\frac{1}{M_1}).
\end{equation}
Therefore, for $i\in\mathcal{W}_n$, as long as $m\gtrsim M_1^2\log M_1$, we have
\begin{equation}
    \bfW_{O_{(i,\cdot)}}^{(0)}\sum_{b=0}^{t-1}\sum_{i\in\mathcal{W}_n}{\bfW_{O_{(i,\cdot)}}^{(b)}}^\top>0.\label{OO1}
\end{equation}
Given $B\gtrsim M_1\log M_1$, by Lemma \ref{lemma: V}, and combining (\ref{OO1}), we have that
\begin{equation}
\begin{aligned}
    &\eta\frac{1}{B}\sum_{n\in\mathcal{B}_b}\frac{1}{a}\sum_{s=1}^{l+1}\text{softmax}({\bfp_s^n}^\top{\bfW_K^{(t)}}^\top\bfW_Q^{(t)}\bfp_{query}^n)\sum_{b=0}^{t-1}\eta(\sum_{i\in\mathcal{W}_n}V_i(b)\bfW_{O_{(i,\cdot)}}^{(b)}\\
        &+\sum_{i\in\mathcal{U}_n}V_i(b)\bfW_{O_{(i,\cdot)}}^{(b)}+\sum_{i\notin\mathcal{W}_n\cup\mathcal{U}_n}V_i(b)\bfW_{O_{(i,\cdot)}}^{(b)})^\top{\bfW_{O_{(i,\cdot)}}^{(t)}}^\top\\
        >&0.
\end{aligned}
\end{equation}
Therefore, the conclusion holds when $t=1$. \\
\noindent Suppose that the conclusion holds when $t\leq t_0$. When $t=t_0+1$, 
by (\ref{WU_update_to_prove}), we can check that
\begin{equation}
    \begin{aligned}
        &\bfW_{O_{(i,\cdot)}}^{(0)}{\bfW_{O_{i,\cdot}}^{(t)}}^\top\\
        =& \bfW_{O_{(i,\cdot)}}^{(0)}({\bfW_{O_{i,\cdot}}^{(0)}}^\top+\eta\frac{1}{B}\sum_{n\in\mathcal{B}_b}(+z^n)  a_{i}\mathbbm{1}[\bfW_{O_{(i,\cdot)}}\sum_{s=1}^{l+1}(\bfW_V\bfp_s^n)\text{softmax}({\bfp_s^n}^\top{\bfW_K^{(t-1)}}^\top\bfW_Q^{(t-1)}\bfp_{query}^n)\geq0]\\
    &\cdot\sum_{s=1}^{l+1} (\bfW_V^{(t-1)}\bfp_s^n)\text{softmax}({\bfp_s^n}^\top{\bfW_K^{(t-1)}}^\top\bfW_Q^{(t-1)}\bfp_{query}^n)\\
    >&0+0=0,
    \end{aligned}\label{WOWO}
\end{equation}
where the second $0$ comes from (\ref{O_P}) and the conditions that such conclusion in (\ref{WOWO}) holds when $t\leq t_0$. Combining (\ref{WO_p}) and the fact that the weighted summation of $\bfp_s^n$ is close to be in the direction of $\bfmu_j$ and $\bfq$ in the feature label embeddings, respectively, where $\bfmu_j$ is the IDR pattern of the $\bfp_{query}$, we have
\begin{equation}
    \eta\frac{1}{B}\sum_{n\in\mathcal{B}_b}\frac{1}{a}\sum_{s=1}^{l+1}\text{softmax}({\bfp_s^n}^\top{\bfW_K^{(t)}}^\top\bfW_Q^{(t)}\bfp_{query}^n)\delta{\bfp_s^n}^\top{\bfW_{O_{(i,\cdot)}}^{(t)}}^\top>0,\label{O_P_2}
\end{equation}
\begin{equation}
    \begin{aligned}
        &\sum_{b=0}^{t-1}\eta \sum_{i\in\mathcal{W}_n}V_i(b)\bfW_{O_{(i,\cdot)}}^{(b)}{\bfW_{O_{i,\cdot}}^{(t)}}^\top\\
        =& (\sum_{b=0}^{t_0-1}\eta \sum_{i\in\mathcal{W}_n}V_i(b)\bfW_{O_{(i,\cdot)}}^{(b)}+\eta \sum_{i\in\mathcal{W}_n}V_i(t_0)\bfW_{O_{(i,\cdot)}}^{(t_0)})({\bfW_{O_{(i,\cdot)}}^{(t_0)}}+\quad\eta\frac{1}{B}\sum_{n\in\mathcal{B}_b}\frac{\partial \ell(\tilde{\bfP}^n,z^n;\Psi)}{\partial \bfW_{O_{(i,\cdot)}}^{(t)}}\Big|t=t_0)^\top\\
        >& 0+(\sum_{b=0}^{t_0-1}\eta \sum_{i\in\mathcal{W}_n}V_i(b)\bfW_{O_{(i,\cdot)}}^{(b)}+\eta \sum_{i\in\mathcal{W}_n}V_i(t_0)\bfW_{O_{(i,\cdot)}}^{(t_0)}){\quad\eta\frac{1}{B}\sum_{n\in\mathcal{B}_b}\frac{\partial \ell(\tilde{\bfP}^n,z^n;\Psi)}{\partial \bfW_{O_{(i,\cdot)}}^{(t)}}\Big|t=t_0}^\top\\
        =& (\sum_{b=0}^{t_0-1}\eta \sum_{i\in\mathcal{W}_n}V_i(b){\bfW_{O_{(i,\cdot)}}^{(t_0)}}^{(b)}+\eta \sum_{i\in\mathcal{W}_n}V_i(t_0)\bfW_{O_{(i,\cdot)}}^{(t_0)})\eta\frac{1}{B}\sum_{n\in\mathcal{B}_b}(-z^n)  a_{i}\mathbbm{1}[\bfW_{O_{(i,\cdot)}}^{(t_0)}\sum_{s=1}^{l+1}(\bfW_V^{(t_0)}\bfp_s^n)\\
        &\cdot\text{softmax}({\bfp_s^n}^\top{\bfW_K^{(t_0)}}^\top\bfW_Q^{(t_0)}\bfp_{query}^n)\geq0]
    \cdot\sum_{s=1}^{l+1} \text{softmax}({\bfp_s^n}^\top{\bfW_K^{(t_0)}}^\top\bfW_Q^{(t_0)}\bfp_{query}^n)(\delta({\bfp_s^n}^\top, \boldsymbol{0}^\top)^\top\\
    &+\sum_{b=0}^{t_0-1}\eta(\sum_{i\in\mathcal{W}_n}V_i(b)\bfW_{O_{(i,\cdot)}}^{(b)}+\sum_{i\in\mathcal{U}_n}V_i(b)\bfW_{O_{(i,\cdot)}}^{(b)}+\sum_{i\notin\mathcal{W}_n\cup\mathcal{U}_n}V_i(b)\bfW_{O_{(i,\cdot)}}^{(b)})^\top),\\
    >&0,
    \end{aligned}
\end{equation}
where the first step is by the formula of the gradient descent, and the second step is by $(\sum_{b=0}^{t_0-1}\eta \sum_{i\in\mathcal{W}_n}V_i(b)\bfW_{O_{(i,\cdot)}}^{(b)}+\eta \sum_{i\in\mathcal{W}_n}V_i(t_0)\bfW_{O_{(i,\cdot)}}^{(t_0)}){\bfW_{O_{(i,\cdot)}}^{(t_0)}}>0$ from the induction steps. The last step comes from the fact that $\|\sum_{b=0}^{t_0-1}\eta\sum_{i\in\mathcal{W}_n}V_i(b)\bfW_{O_{(i,\cdot)}}^{(b)}\|^2>0$ and $\sum_{b=0}^{t_0-1}\eta\sum_{i\in\mathcal{W}_n}V_i(b)\bfW_{O_{(i,\cdot)}}^{(b)}\cdot {\eta \sum_{i\in\mathcal{W}_n}V_i(t_0)\bfW_{O_{(i,\cdot)}}^{(t_0)}}^\top>0$ by the induction, and that $V_i(t)$ for $i\notin\mathcal{W}_n$ is much smaller than that in $\mathcal{W}_n$ given $B\gtrsim M_1$. Then, 
\begin{equation}
    \sum_{b=0}^{t-1}\eta(\sum_{i\in\mathcal{W}_n}V_i(b)\bfW_{O_{(i,\cdot)}}^{(b)}+\sum_{i\in\mathcal{U}_n}V_i(b)\bfW_{O_{(i,\cdot)}}^{(b)}+\sum_{i\notin\mathcal{W}_n\cup\mathcal{U}_n}V_i(b)\bfW_{O_{(i,\cdot)}}^{(b)})^\top{\bfW_{O_{(i,\cdot)}}^{(t)}}^\top>0,\label{O_O_2}
\end{equation}
since the norm $\bfW_{O_{(i,\cdot)}}^{(t)}$ for $i\notin\mathcal{W}_n$ is no larger than that for $i\in\mathcal{W}_n$. 
Combining (\ref{WOWO}), (\ref{O_P_2}), and (\ref{O_O_2}), we have
\begin{equation}
    \bfW_{O_{(i,\cdot)}}^{(t)}\bfV^n(t)>0.
\end{equation}
Hence, we finish the induction. \\
\noindent (2) When $t\gtrsim \Theta(\beta)$, for $i\in\mathcal{W}$, by checking (\ref{WU_update_to_prove}), we can deduce that
\begin{equation}
    \bfW_{O_{(i,\cdot)}}^{(0)}{\bfW_{O_{(i,\cdot)}}^{(t)}}^\top+\eta\frac{1}{B}\sum_{n\in\mathcal{B}_b}\frac{1}{a}\sum_{s=1}^{l+1}\text{softmax}({\bfp_s^n}^\top{\bfW_K^{(t)}}^\top\bfW_Q^{(t)}\bfp_{query}^n)\delta{\bfp_s^n}^\top{\bfW_{O_{(i,\cdot)}}^{(t)}}^\top>0,\label{W_general_1}
\end{equation}
since that the accumulated label embedding term of $\bfW_{O_{(i,\cdot)}}^{(t)}$ contributed positively to ${\bfp_s^n}^\top{\bfW_{O_{(i,\cdot)}}^{(t)}}^\top$ and is larger than that of the feature embedding contribution by (\ref{hoef_xi1}) and (\ref{hoef_xi2}) (the gradient updates is close in the direction of the IDR pattern of $\bfp_{query}^n$ when $m\gtrsim M_1^2$). Since that $\|\bfW_{O_{(i,\cdot)}}^{(0)}
(\bfmu_j^\top,\boldsymbol{0}^\top)^\top\|\leq \beta\xi$ for any $j\in[M_1]$, the effect of $\bfW_{O_{(i,\cdot)}}^{(0)}{\bfW_{O_{(i,\cdot)}}^{(t)}}^\top$ to the sign is much smaller than the remaining terms in (\ref{W_general_1}). Hence, we show (\ref{W_general_1}). \\
\noindent Then, since that the label embedding of $\bfW_{O_{(i,\cdot)}}$, $\bfW_{O_{(j,\cdot)}}$ are both close to $\bfq$ for $i,j\in\mathcal{W}$, and that the feature embedding of $\bfW_{O_{(i,\cdot)}}, i\in\mathcal{W}_n$ is close to the IDR pattern of $\bfp_{query}^n$, which is not the negative direction of the feature embedding of $\bfW_{O_{(j,\cdot)}}, j\in\mathcal{W}_{n'}$, we have for $j\in\mathcal{W}\backslash\mathcal{W}_n$,
\begin{equation}
    \sum_{b=0}^{t-1}\eta\sum_{i\in\mathcal{W}_n}V_i(b){\bfW_{O_{(i,\cdot)}}^{(b)}}^\top{\bfW_{O_{(j,\cdot)}}^{(t)}}^\top>0.
\end{equation}
Given that $V_i(t)$ for $i\notin\mathcal{W}_n$ is much smaller than that in $\mathcal{W}_n$ given $B\gtrsim M_1$ and the norm $\bfW_{O_{(i,\cdot)}}^{(t)}$ for $i\notin\mathcal{W}_n$ is no larger than that for $i\in\mathcal{W}_n$, we have
\begin{equation}
    \sum_{b=0}^{t-1}\eta(\sum_{i\in\mathcal{W}_n}V_i(b)\bfW_{O_{(i,\cdot)}}^{(b)}+\sum_{i\in\mathcal{U}_n}V_i(b)\bfW_{O_{(i,\cdot)}}^{(b)}+\sum_{i\notin\mathcal{W}_n\cup\mathcal{U}_n}V_i(b)\bfW_{O_{(i,\cdot)}}^{(b)})^\top{\bfW_{O_{(i,\cdot)}}^{(t)}}^\top>0.
\end{equation}
Therefore, we can derive that for $i\in\mathcal{W}$,
\begin{equation}
    \bfW_{O_{(i,\cdot)}}^{(t)}\bfV^n(t)>0.
\end{equation}

\end{proof}

\end{document}